\documentclass[journal]{IEEEtran}

\usepackage{cite}

\usepackage{epsfig}
\usepackage{epstopdf}

\usepackage{tabularx}
\usepackage{multirow}      
\usepackage{multicol}   
\usepackage{booktabs}
\usepackage{times}
\usepackage{float}

\usepackage{algorithmic}
\usepackage[ruled]{algorithm2e}

\usepackage{amsmath}
 \usepackage{amssymb}
\usepackage{mathptmx}

\usepackage{enumitem}
\usepackage{placeins}

\usepackage{color}
\usepackage{subfig}

\usepackage{url}

\usepackage{balance}
\usepackage{breqn}

\DeclareMathAlphabet{\mathcal}{OMS}{cmsy}{m}{n}

\begin{document}

\title{Cluster Pruning: An Efficient Filter Pruning Method for Edge AI Vision Applications}

\author{
	\IEEEauthorblockN{
		Chinthaka Gamanayake\IEEEauthorrefmark{1},
		Lahiru Jayasinghe\IEEEauthorrefmark{2}, 
		Benny Ng\IEEEauthorrefmark{3},
		Chau Yuen\IEEEauthorrefmark{4}}\\
	
	\IEEEauthorblockA{
		Singapore University of Technology and Design\\
		Email: \{\IEEEauthorrefmark{1}chinthaka\_madhushan,
		\IEEEauthorrefmark{2}aruna\_jayasinghe,
		\IEEEauthorrefmark{3}benny\_ng,
		\IEEEauthorrefmark{4}yuenchau\}@sutd.edu.sg}
	}


\maketitle

\begin{abstract}
	
Even though the Convolutional Neural Networks (CNN) has shown superior results in the field of computer vision, it is still a challenging task to implement computer vision algorithms in real-time at the edge, especially using a low-cost IoT device due to high memory consumption and computation complexities in a CNN. Network compression methodologies such as weight pruning, filter pruning, and quantization are used to overcome the above mentioned problem. Even though filter pruning methodology has shown better performances compared to other techniques, irregularity of the number of filters pruned across different layers of a CNN might not comply with majority of the neural computing hardware architectures. In this paper, a novel greedy approach called cluster pruning has been proposed, which provides a structured way of removing filters in a CNN by considering the importance of filters and the underlying hardware architecture. The proposed methodology is compared with the conventional filter pruning algorithm on Pascal-VOC open dataset, and Head-Counting dataset, which is our own dataset developed to detect and count people entering a room. We benchmark our proposed method on three hardware architectures, namely CPU, GPU, and Intel Movidius Neural Computer Stick (NCS) using the popular SSD-MobileNet and SSD-SqueezeNet neural network architectures used for edge-AI vision applications. Results demonstrate that our method outperforms the conventional filter pruning methodology, using both datasets on above mentioned hardware architectures. Furthermore, a low cost IoT hardware setup consisting of an Intel Movidius-NCS is proposed to deploy an edge-AI application using our proposed pruning methodology.

\end{abstract}


%
\IEEEpeerreviewmaketitle

\section{Introduction}

\begin{figure}[h]
	\centering
	\includegraphics[scale=0.7]{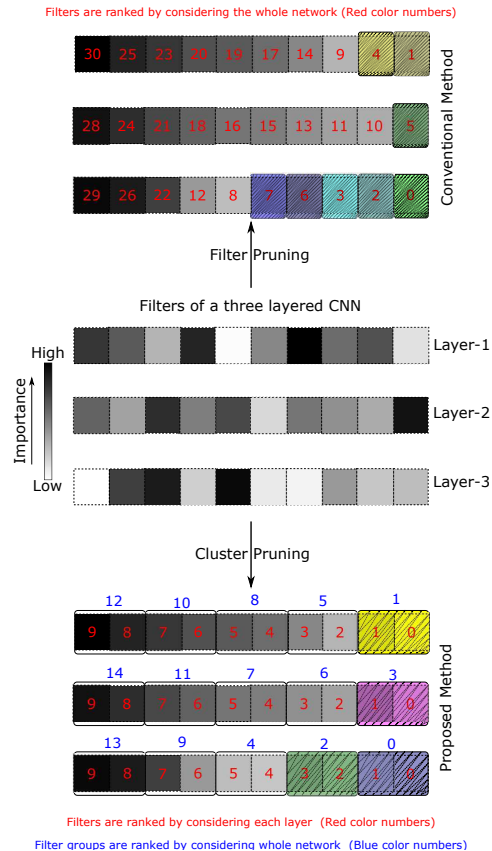}
	\caption{Filter Pruning vs Cluster Pruning. For the demonstration purpose we have selected only three layers of a CNN, where each layer consists of 9 filters.}
	\label{system_overview}
\end{figure}

In recent years, computer vision applications achieved significant improvement in accuracy over image classification and object detection applications. Such progress is made mainly due to the growth of underlying Convolution Neural Networks (CNNs), deeper and wider. Then, Deep Neural Networks (DNNs) \cite{krizhevsky2012imagenet, szegedy2015going, simonyan2014very, he2016deep} became the general trend after the introduction of AlexNet \cite{russakovsky2015imagenet} in ImageNet Challenge in 2012. Most of these CNNs usually have hundreds of layers and thousands of channels, thus requiring computation at billions of floating point operations (FLOPS) with a memory footprint at hundreds of megabytes. Since the improvement of the accuracy does not necessarily make networks more efficient with respect to size and speed, directly hand-craft more efficient mobile architectures were introduced. Lower-cost 1x1 convolutions inside the fire-modules reduces the number of parameters in SqueezeNet \cite{iandola2016squeezenet}. Xception \cite{chollet2017xception}, MobileNets \cite{howard2017mobilenets}, \cite{sandler2018mobilenetv2} and Network-decoupling \cite{guo2018network} employ depthwise separable convolution to minimize computation density replacing the conventional convolutional layers. ShuffleNets \cite{zhang2018shufflenet, ma2018shufflenet} utilize low-cost group convolution and channel shuffle. Learning of the group convolution is used across layers in CondenseNet \cite{huang2018condensenet}. On the other hand, faster object detections has been achieved in YOLO \cite{redmon2016you} by introducing a single-stage detection pipeline, where region proposition and classification is performed by one single network simultaneously. SSD \cite{liu2016ssd} has outperformed YOLO by eliminating region proposals and pooling in the neural network architecture. Inspired by YOLO, SqueezeDet \cite{wu2017squeezedet} further reduces parameters by the design of \textit{ConvDet} layer. Based on the deeply supervised object detection(DSOD) \cite{shen2017dsod} framework, Tiny-DSOD \cite{li2018tiny} introduces two innovative and ultra-efficient architecture blocks namely depthwise dense block (DDB) and depthwise feature-pyramid-network (D-FPN) for resource-restricted usages. These novel convolution operations are not supported by most of the current hardware and software libraries. That leaves difficulties in implementations and also these models take significant human efforts at the design phase.

Implementing real-time edge-AI applications such as face-detection, pedestrian detection, and object classification on resource-constrained devices, especially low-cost Internet-of-Things (IoT) devices, require models with less memory and fewer number of FLOPS. Pioneered from the work done in Optimal Brain Damage \cite{lecun1990optimal} and Optimal Brain Surgeon \cite{hassibi1993second}, network compression has become a reasonable solution to simplify high capacity networks. Network magnitude based weight pruning methodologies suggested in \cite{lecun1990optimal, hassibi1993second, yu2012exploiting, han2015learning, han2015deep, hu2016network, molchanov2016pruning, xu2018deep} can dramatically decrease CNN model sizes and the number of multiply–accumulate operations (MAC). However the regular structure of dense matrices is distorted by weight pruning. This introduces sparse weigh matrices, which require additional computations and special hardware designs to evaluate the network.

In line with our work, several pruning methods have been proposed in \cite{molchanov2016pruning, li2016pruning, anwar2016compact, he2017channel, luo2018thinet}, where entire convolutional filters are removed. When aforementioned methods prune filters after an initial training phase of the network, the network slimming method \cite{liu2017learning} learns to remove filters in the training phase in-cooperating a scaling factor. Since these filter pruning methods do not introduce sparsity to the original network structure, it requires no special software or hardware implementations to gain the peak performance. However, most of the edge-AI hardware architectures provide optimum performance when the workload size and memory required is aligned to hardware dependant numbers, which is in most cases exist as numbers in power of two \cite{corp, han2016eie, han2017ese}. This is due to the schedulers load balancing problem over the processing element and memory alignment requirement. Thus pruning filters across layers might introduce a performance degradation in some hardware architectures due to the irregularity of number of filters pruned across layers. 

To develop a hardware aware DNN pruning methodology, it is important to explore different hardware architectures used for DNN processing. For instance, the x86 Family is not meant for DNN, but there are some attempts to use clusters of CPUs for Deep Learning (DL) (BigDL from Intel \cite{dai2018bigdl}) and optimizing DL libraries for CPUs (Caffe con Troll \cite{hadjis2015caffe}). The Intel Xeon scalable processors features AVX instructions for deep learning. Then the Nvidia GPU's features massively parallel accelerations with its concurrent programming and hardware platform CUDA \cite{nvidia_cuda}. Many real world applications such as robotics, self-driving cars, augmented reality, video surveillance, mobile-apps and smart city application \cite{lau2019survey, marakkalage2018understanding, liu2020cooperative} require IoT devices capable of AI inference. Thus, DNN inference has also been demonstrated on various embedded System-onChips (SoC) such as Nvidia Tegra, Samsung Exynos, as well as application specific FPGA designs (ESE \cite{han2017ese}, SCNN \cite{parashar2017scnn}, \cite{piyasena2019reducing}, \cite{piyasena2019lowering}), and ASICs such as GoogleTPU and Movidius-NCS, which is used later in our experiment. Except FPGAs, most of these devices are generalized to work with majority of DNN architectures. Therefore, theoretical performance gain from conventional pruning methods might not be achieved directly using these hardware architectures.

Inspired by the work done related to Neural Architecture Search \cite{zoph2018learning, cai2017reinforcement, ashok2017n2n}, AutoML for Model Compression (ACM) \cite{he2018amc} has leveraged reinforcement learning for neural network compression to achieve state of the art results. On the other hand, NetAdapt \cite{yang2018netadapt} proposes an algorithm that automatically adapts a pre-trained deep neural network to a mobile platform given a resource budget using empirical measurements. The crucial difference between aforementioned methods and ours is that we do not propose a fully automated pruning methodology, which does not have a learning or an exhaustive network searching phase to find the optimal pruning ratio, but rather a rule-based, three steps method for faster implementation. Our method has better control over selecting layer to be pruned manually, and also we can learn the behaviour of different hardware devices susceptible to pruning of the network. Furthermore, we can choose the pruning complexity required for each layer manually based on the obtained observations. Nonetheless, we expect the automatic pruning be a promising future work, which potentially can obtain a better performance than manual pruning.

In this paper, we propose a novel pruning methodology, named \textit{cluster pruning}, which in-cooperate hardware dependent parameters and follows a rule based greedy algorithm to prune the entire network. We formulate an optimization problem to measure the hardware response towards the performance (accuracy and inference latency) of the network. Then, we solve this problem by three steps. First we analyse the performance by pruning one layer at a time. Then we identify the optimum cluster size that maximizes the performance. Finally, we apply cluster pruning for the entire network. Since we do not have a simulation model of the hardware, we carry out above three steps empirically using direct metric measurements while considering the hardware architecture as a black box. 

Fig. \ref{system_overview} demonstrates the cluster pruning and filter pruning methodologies using three layers of a CNN as an example. Normally, the importance of filters in the CNN is randomly distributed. Cluster pruning method ranks the filters considering each layer, while filter pruning method ranks them considering the whole network. Then, cluster pruning method goes one step ahead by ranking groups of filters considering the whole network. As shown in the figure by faded colours, cluster pruning method prunes 4 groups of filters, while filter pruning method removes 8 filters one by one. We utilize the open dataset Pascal-VOC and own-created Head-Counting dataset to demonstrate the practical applicability of our method along with a real-time application. The results show that the proposed method can successfully mitigate the performance degradation and outperform the filter pruning method.

This paper is organized as follows. Neural computing hardware architectures used are described in the Section II. The cluster pruning methodology is proposed in Section III, and the experimental results are shown in Section IV. Section V makes the concluding remarks.

\section{Parallel Computer Paradigms}

\begin{figure}
	\centering
	\includegraphics[scale = 0.8]{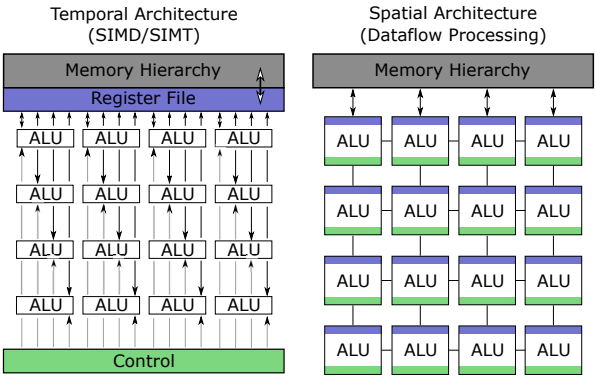}
	\caption{Parallel Compute Paradigms}
	\label{compute_paradigms}
\end{figure}

Understanding of how the processing is being performed on AI computing architectures is crucial to describe the performance response after pruning a single layer of a CNN. The fundamental component of both the convolution and fully connected layers are the multiply-and-accumulate (MAC) operations, which can be easily parallelized. The processing elements inside computer architectures mainly can be separated into two compute paradigms as shown in Fig. \ref{compute_paradigms}, as mentioned in \cite{sze2017efficient}. 

Mostly CPUs and GPUs exploit the temporal architectures, and mainly employ parallelization techniques such as vectors (SIMD) or parallel threads (SIMT). These temporal architectures use centralized controllers such as schedulers to manage a large number of ALUs. Normally these ALUs can't communicate directly with each other and only fetch data from the memory hierarchy. Due to high computational capability, and also memory and scheduling efficiency in temporal architecture, we do not experience the effect on performance significantly after pruning filters irregularly across layers of a CNN.

\begin{figure}
	\centering
	\includegraphics[scale = 0.85]{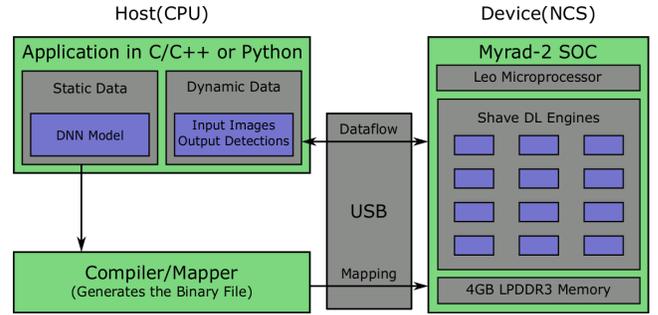}
	\caption{Movidius-NCS Architecture}
	\label{ncs_arch}
\end{figure}

In contrast, spatial architectures use data-flow processing, where the ALUs form a processing chain, so that they can pass data from one to another directly. This architecture is commonly used for DNNs in ASIC and FPGA based designs. The design principles for Movidius-NCS, which is used for our experiment is also based on a spatial architecture. It is made of Myriad-VPUs follows from a careful balance of programmable vector-processors, dedicated hardware accelerators, and memory architecture for optimized data flow \cite{movidius}. There is a fixed data-flow designed that adapt to certain DNN shapes and sizes. Therefore, irregular number of filters remain in a single layer after pruning might introduce a performance degradation. Fig. \ref{ncs_arch} shows how the Movidius-NCS architecture is organized. The pre-trained DNN model used in the application is compiled and mapped into the Movidius-NCS before the real-time application is started. Workload in the DNN is distributed over the DL engines. This mapping might reduce the network size and might introduce accuracy drop generally. If the mapping is fixed for certain network shapes, pruning might effect the performance of the application adversely.

\section{Proposed Approach}

Consider a set of training examples $\mathcal{D}=\{(x_n,y_n)|n=1,...,N\}$, where $N$ represents the number of training examples, $x_n$ and $y_n$ represent the $n^{th}$ input and its target output, respectively. Consider a CNN has the convolution filters $\mathcal{F} = \{F_l^{(k)}|l=1,...,L;k=1,...,K_l\}$, where $l$ represents the convolution layer index, $k$ represents the filter index in the $l^{th}$ layer and $K_l$ is the total number of filters inside $l^{th}$ layer. All the filters of the network are trained to minimize a cost function $C(\mathcal{D}|\mathcal{F})$, which in turn maximize the accuracy of detections. 

During pruning, we refine a subset of filters $\mathcal{F}'$, which preserves the accuracy of the adapted network such that $C(\mathcal{D}|\mathcal{F}')\approx C(\mathcal{D}|\mathcal{F})$. Then the problem can be formulated as
\begin{multline}\label{cost_eqn}
\min\limits_{\mathcal{F}'} \mid C(\mathcal{D}|\mathcal{F}') - C(\mathcal{D}|\mathcal{F}) \mid  \\ s.t \ Cons_{m}(\mathcal{F}') < B_{m} \ and \ Cons_{t}(\mathcal{F}') < B_{t}, 
\end{multline}
where $Cons_{m}(.)$ and $Cons_{t}(.)$ evaluate the memory and inference time consumption for a selected sub set of filters in the network. $B_{m}$ and $B_{t}$ represent the memory bound and latency bound at our hand. Intuitively, if $\mathcal{F} = \mathcal{F}'$, we reach the global minimum of Eq. \ref{cost_eqn}.

While pruning a CNN, some filters along with the corresponding feature maps are removed, resulting in a structural change in the network. Therefore, pruning could lead to a potential problem of unbalanced workload over processing elements and might not fit well on parallel computer architectures, specially for edge-AI devices with limited resource. Hence, workload imbalance may cause a gap between the expected performance and peak performance \cite{han2017ese}.

In order to address this issue, we consider the effect from hardware architecture on performance of the network, which is accuracy and throughput. The motivation behind our approach is to identify and maximize the hardware architecture dependent accuracy and throughput response while pruning the network. For selected $\mathcal{F}'$, the accuracy response is given by $H_{acc}(\mathcal{F}')$, and the throughput response is given by $H_{speed}(\mathcal{F}')$. Since these two responses are connected with each other, solving the Eq. \ref{hardware_eqn} would provided the filter subset, which is required to gain the optimum hardware-aware performances, where $\alpha_{acc}$ and $\alpha_{speed}$ represent the scaling factors for the $H_{acc}(\mathcal{F}')$ and $H_{speed}(\mathcal{F}')$, respectively.

\begin{multline}\label{hardware_eqn}
\max\limits_{\mathcal{F}'} \{ \alpha_{acc}H_{acc}(\mathcal{F}') + \alpha_{speed}H_{speed}(\mathcal{F}') \} \\ s.t. \ Cons_{m}(\mathcal{F}') < B_{m} \ and \ Cons_{t}(\mathcal{F}') < B_{t}, 
\end{multline}

However, solving Eq. \ref{cost_eqn} and Eq. \ref{hardware_eqn} is a combinatorial optimization problem \cite{molchanov2016pruning}. There are $2^{\mid \mathcal{F} \mid}$ number of evaluations for both of the equations to select the optimum subset of filters. Moreover, there are thousands of convolutional filters in modern CNN architectures. Hence, it is difficult to solve this optimization exactly using an exhaustive search. 

Therefore, we implement a greedy methodology to solve this problem empirically by considering the underlying hardware architecture. Greedy methodologies consist of a least important filter selection criteria and  then remove those filters iteratively until the expected memory and latency bounds are reached.

\subsection{Analyzing Single Layer Performance Response}

As the first step, we prune the less significant filters in a layer, then profile the accuracy and latency response for a given hardware architecture. In the literature, there are some heuristic criteria have been proposed to evaluate the importance of each filter in a neural network. Some of the important criteria include Minimum Weight \cite{li2016pruning}, Average Percentage of Zeros \cite{hu2016network}, Talor Criteria \cite{molchanov2016pruning}, and Thinet greedy algorithm \cite{luo2018thinet}. We adapt the minimum weight criteria to rank the convolutional filters to determine their significance toward the performances \cite{li2016pruning}. Minimum weight criteria for an individual filter can be represented as $\theta_{MW}: \mathbb{R}^{\mid F_l^k \mid \times p \times p}\longrightarrow \mathbb{R}$, which can be formulated as \begin{equation}\label{mw} \theta_{MW}(F_l^k) = \frac{1}{\mid F_l^k \mid \times p \times p} \sum_j^{\mid F_l^k \mid} \sum_i^{p \times p} w_{i,j}^2, \end{equation} where $p \times p$ represent the kernel size, $w$ denotes a individual kernel weight, $\mid F_l^k \mid$ represent the cardinality, which is the number of kernels in the $k^{th}$ filter of the $l^{th}$ layer. 

Using the Eq. \ref{mw}, we ranked the filters according to their increasing order of significant.  Then we start to prune them in ascending order of the rank and profiled the accuracy and latency of the network for each pruning instance as shown in Fig. \ref{fig:single_layer_ncs}, Fig. \ref{fig:single_layer_all}, and Fig. \ref{fig:accuracy_single_layer}.

\subsection{Identifying The Optimum Cluster Size}

If the hardware architecture is susceptible to workload imbalance, the influence will be reflected in the performance graphs when we analyse the single layer pruning results. If there exist a particular pattern of inference time drops in latency graphs with respect to the number of filters left in a layer after pruning, networks forward inference time is influenced. On the other hand, there might be particular patterns of rises in accuracy graphs. As an example, in Fig. \ref{fig:single_layer_all} and Fig. \ref{fig:accuracy_single_layer}, we can identify periodic bottoms with significant drops and periodic peaks with significant rises in the latency and accuracy graphs, respectively. Consider the $p_l^{acc}$ and $p_l^{lat}$ as the identified periodic lengths from the accuracy and latency graphs of the $l^{th}$ layer. The optimum cluster size $P_l$ can be calculated as the $LCM(p_l^{acc},p_l^{lat})$, where $LCM(.,.)$ represent the calculation of least common multiple of the two given periodic lengths. Likewise, we calculate the optimum cluster size for every layer, which is denoted by $\mathcal{P}=\{P_l|l=1,...,L\}$. 

\begin{algorithm}[t]
	\DontPrintSemicolon
	\SetAlgoLined
	\SetKwInOut{Input}{Input}
	\KwIn{Pretrained Network with Filter Set: $F$\\ \ \ \ \ \ \ \ \ \ Optimum Cluster Sizes per layer: $\mathcal{P}$}
	\SetKwInOut{Output}{Output}
	\KwOut{Pruned Network with Filter Set: $F'$}	
	\BlankLine
	Set for filter clusters: $G=\{\}$\;
	Set for avg $\Theta_{MW}(.)$ values of filter clusters: $MW_G=\{\}$ \;
	\For{each layer in network $l=1,...,L$}{
		Set for the filters in current layer: $F_l = \{\}$\;
		Set for $\Theta_{MW}(.)$ values in current layer: $MW_{l}=\{\}$\;
		\For{each filter in current layer $k=1,...,K_l$}{
			$MW_{l} \cup \{\Theta_{MW}(F_l^k)\}$\;
			$F_l \cup \{F_l^k\}$
		}
		$F_R$ = Rank $F_l$ according to values in $MW_{l}$\;
		$i=1$;
		
		\While{all filters groups are processed: $i+P_l > K_l$}{
			Select a cluster of filters:  $F_R^{i:i+p_l}$\;
			Add the cluster to the set: $G \cup \{F_R^{i:i+p_l}\}$\;
			Add avg $\Theta_{MW}(.)$ value of the cluster to the set:\;
			$MW_G \cup \{\frac{\sum_{j=0}^{p_l-1}\Theta_{MW}(F_R^{i+j})}{P_l}\}$\;
			Increment to the next cluster: $i = i + P_l$
		}
	}
	$G_R$ = Rank $G$ according to $MW_G$\;
	
	Until the pruning objective is reached, prune filter groups in $G_R$ consecutively.\;
	\caption{Cluster Pruning Algorithm}
	\label{cluster_pruning}
\end{algorithm}

\subsection{Applying Cluster Pruning to the Whole Network}

The methodology of utilizing the optimum cluster size that is just described in Section III B is shown in Algorithm \ref{cluster_pruning}. First, we iterate through each layer of the network and identify the importance of individual filter in corresponding layer according to the minimum weight criteria using Eq. \ref{mw}. Then we rank them inside the layer according to the calculated importance, which is denoted as $F_R$. For each layer, filter clusters are formed according to the optimum cluster size and those clusters are inserted into the global set denoted as $G$. The importance of the filter groups are calculated by taking the average of $\Theta_{MW}$ values of the filters inside the corresponding group. After that, all the groups in the network are ranked and pruned according to their increasing order of significance.  Finally, iterative pruning can be stopped after reaching the target trade-off between accuracy and pruning objective, which can be the FLOPS, inference latency or memory utilization of the model.

\section{Experimental Results}

In this section, the proposed cluster pruning methodology is empirically evaluated using the popular \textit{SSD-MobileNet} and \textit{SSD-SqueezeNet} neural network architectures for object detection. The popular \textit{Pascal-VOC} dataset and our application specific dataset named as \textit{Head-Counting} is used to pre-train the SSD-MobileNet and \textit{SSD-SqueezeNet} before pruning. Iterative fine-tuning step has been carried out to retain the accuracy of the networks according to the baseline described in \cite{liu2018rethinking}.

We use latency and accuracy as two performance measurements for evaluation. The average network forward inference time across a single layer or the whole network is measured in milliseconds as the latency, and the mean average precision (mAP) value is calculated as the accuracy for test datasets. We use Caffe framework implementation of SSD-MobileNet \cite{chuanqi305_2018} and \textit{SSD-SqueezeNet} \cite{chuanqi305} to develop the pruning methodologies.

The experiment is divided into three parts. In the first part, we profile the performance of different hardware architectures by pruning the filters in a single layer to identify the optimum cluster sizes per layer (Section IV B). Then this optimum cluster size is used for the proposed cluster pruning methodology to prune the whole network (Section IV C). Filter pruning method is also carried out to compare the performance. Finally, we demonstrate the performance comparison of a edge-AI application in different hardware setups after applying cluster pruning and filter pruning methods (Section IV D).

\subsection{Data sets and models}

\subsubsection{Pascal-VOC Dataset}
Pascal-VOC \cite{everingham2010pascal} provides standardized image datasets for object class recognition and consist of 20 classes. The training and validation data has 11,530 images containing 27,450 ROI annotated objects and 6,929 segmentation's, while the testing dataset consist of 4952 images. We use this dataset to train and test our pruned models to get the accuracy an latency values.

\subsubsection{Head-Counting Dataset}
For our edge-AI vision application, we collect data from a live video feed from 5 different cameras mounted on top of the entrance of rooms under various lighting conditions. This dataset consists of 2622 images for training and validation, while 786 images are used for testing. These images are labelled with bounding boxes using only the \textit{person} object category. Since these images were captured in 304x304 resolution, images are re-sized into 300x300 at the beginning.

\subsubsection{SSD-MobileNet Detection Network}
Depthwise separable convolutions are used in MobileNets neural network architecture \cite{howard2017mobilenets} for faster inference. For detection of objects, we use the SSD variation \cite{liu2016ssd} of it. For our experiment, we use two models from this network, which are pre-trained on above mentioned two datasets. The first model is trained on Pascal-VOC dataset from scratch and the second model is fine-tuned on top of the first model using Head-Counting dataset. We prune both of these models using the filter pruning methodology and our proposed cluster pruning methodology to measure the performance response.

\subsubsection{SSD-SqueezeNet Detection Network}
SqueezeNet CNN architecture \cite{iandola2016squeezenet} comprises of blocks called \textit{fire modules}, where conventional convolution has been replaced by a \textit{squeeze} convolution layer feeding into an \textit{expand} layer that has a mix of 1x1 and 3x3 convolution filters. SqueezeNet achieves AlexNet-level accuracy on ImageNet with 50x fewer parameters. We use the SSD variation \cite{liu2016ssd} on top of the backbone SqueezeNet for the detections. For our experiment, we use two models from this network, which are pre trained on Pascal-VOC dataset, and then fine-tuned on Head-Counting dataset. We prune them using both cluster pruning and filter pruning methodologies.

\subsection{Optimum Cluster Size through Single Layer Pruning}

\begin{figure*}
	\centering
	\subfloat[][NCS inference time: Conv0 pruned]
	{\includegraphics[width=0.30\textwidth]{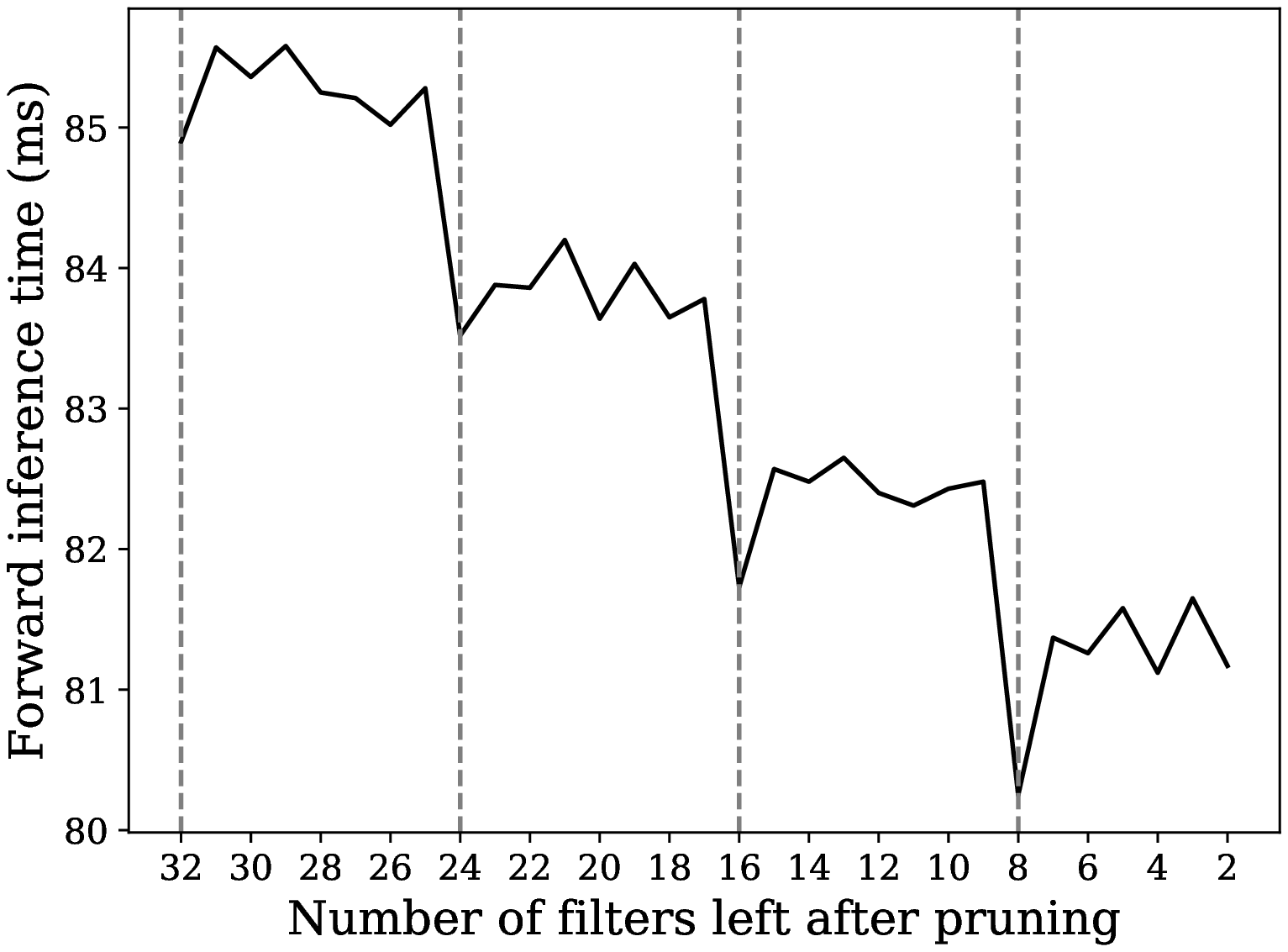}%
		\label{fig:latancy_conv0_ncs}}\hfill
	\subfloat[][CPU inference time: Conv0 pruned]
	{\includegraphics[width=0.30\textwidth]{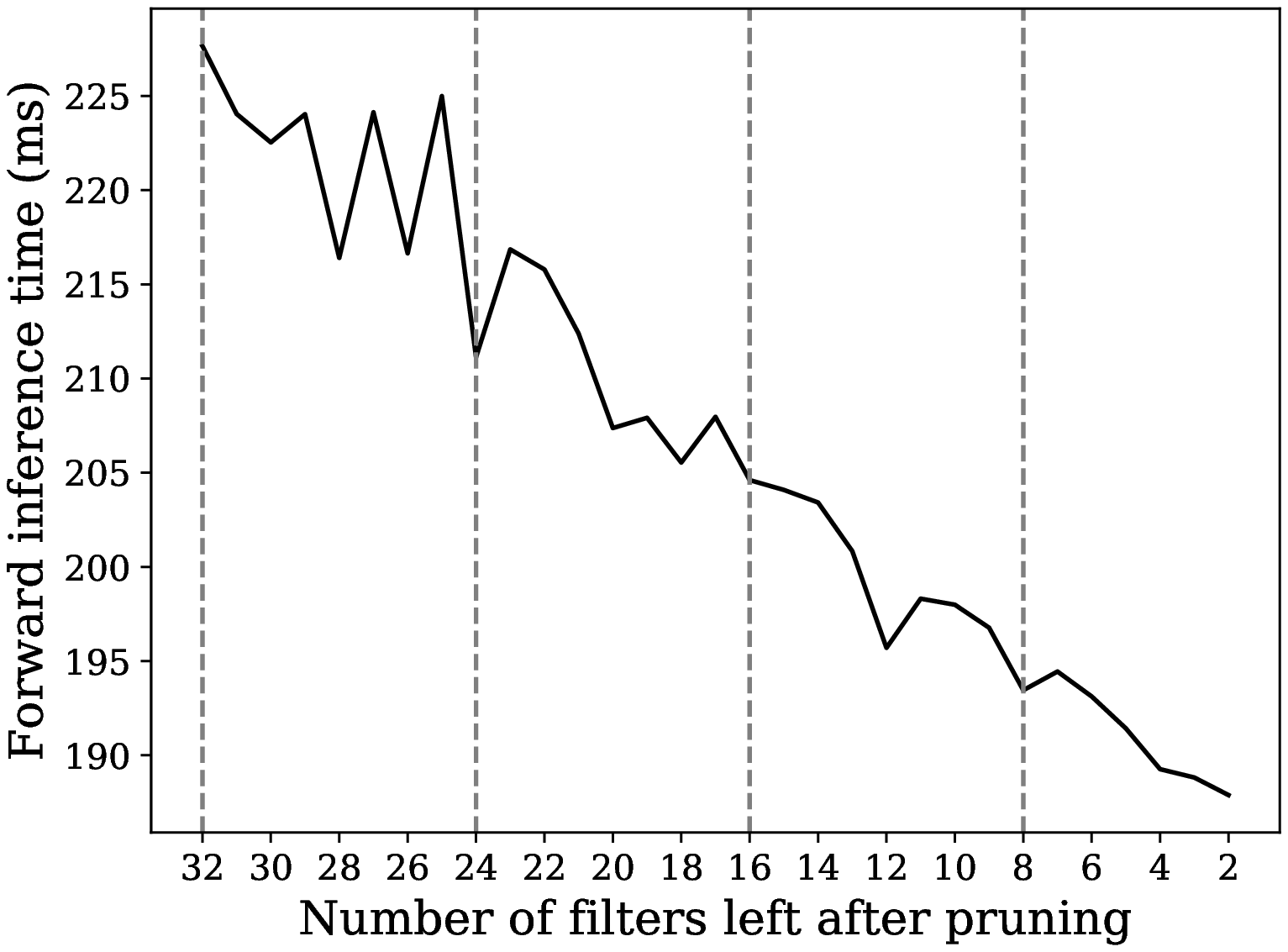}%
		\label{fig:latancy_conv0_cpu}}\hfill
	\subfloat[][GPU inference time: Conv0 pruned]
	{\includegraphics[width=0.30\textwidth]{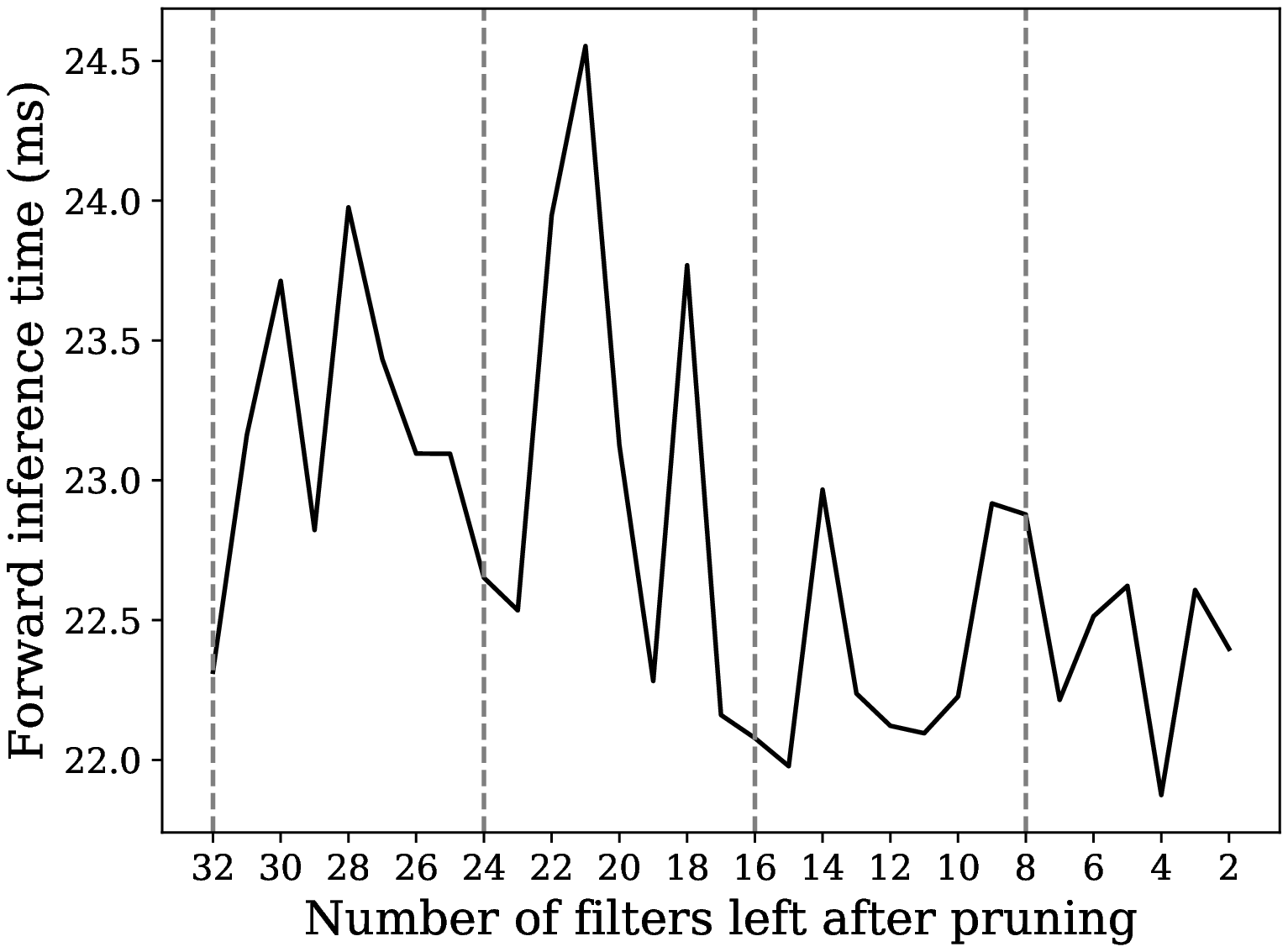}%
		\label{fig:latancy_conv0_gpu}}\hfill
	\subfloat[][NCS inference time: Conv1 pruned]
	{\includegraphics[width=0.30\textwidth]{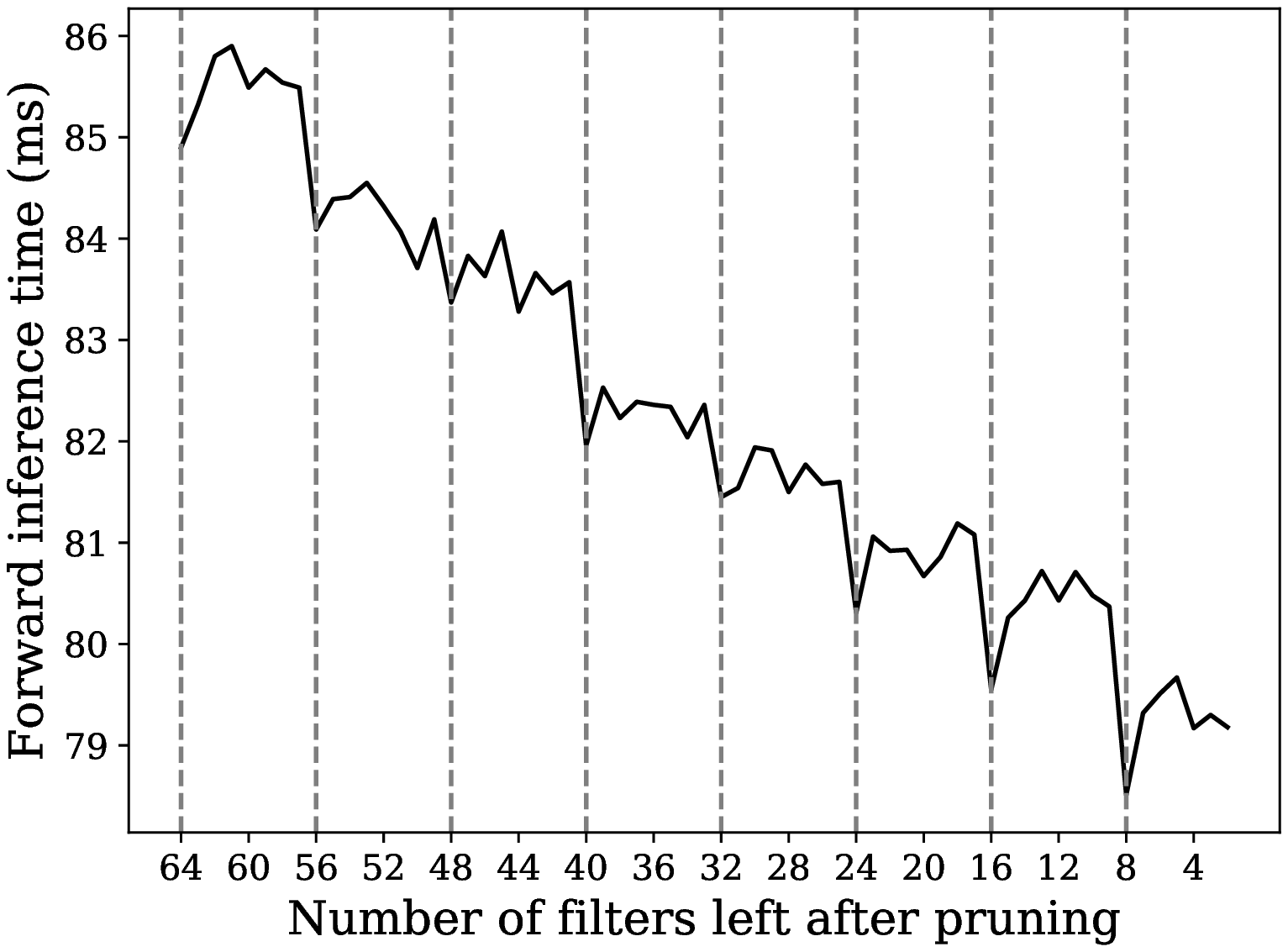}%
		\label{fig:latancy_conv1_ncs}}\hfill
	\subfloat[][CPU inference time: Conv1 pruned]
	{\includegraphics[width=0.30\textwidth]{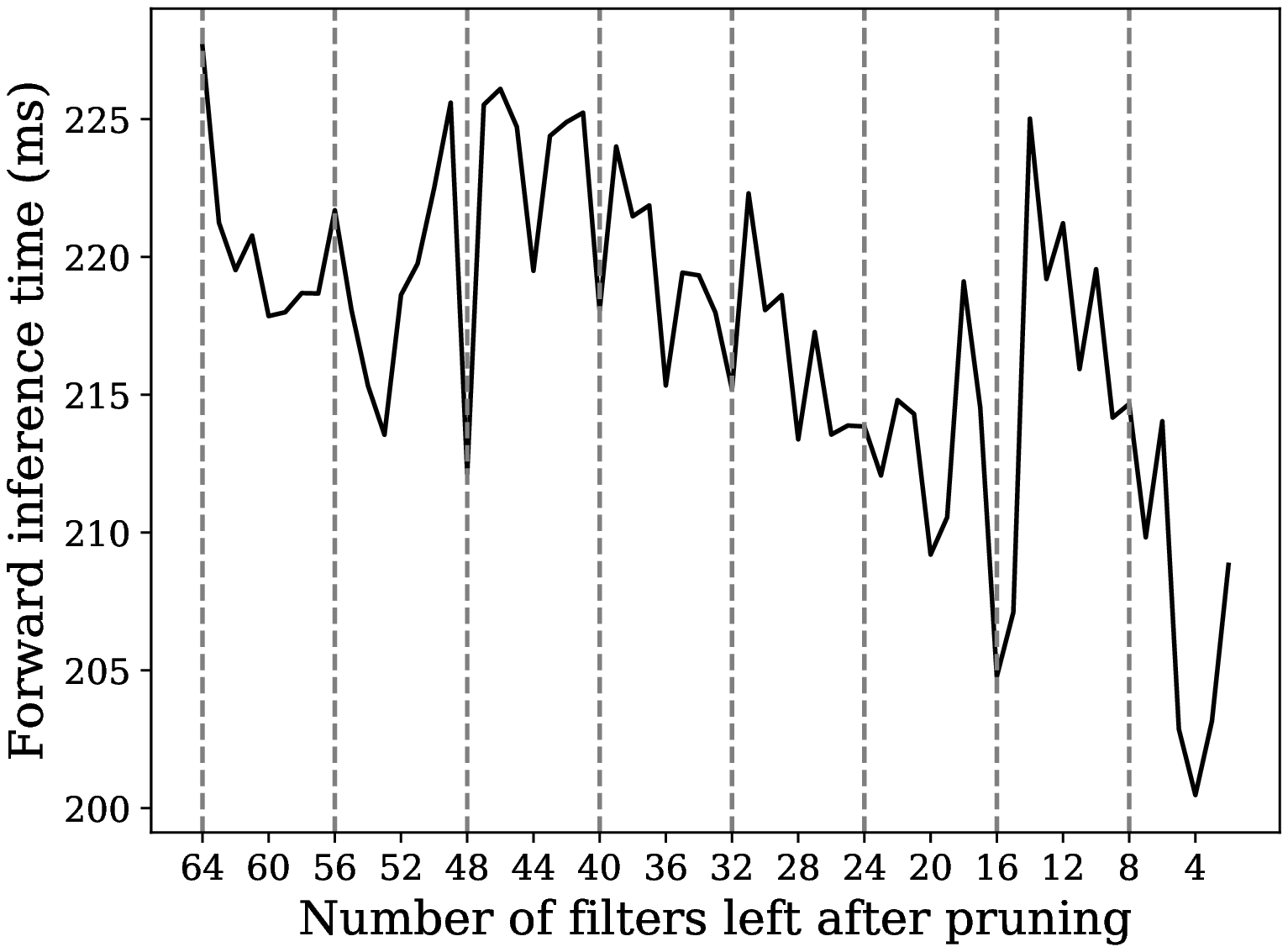}%
		\label{fig:latancy_conv1_cpu}}\hfill
	\subfloat[][GPU inference time: Conv1 pruned]
	{\includegraphics[width=0.30\textwidth]{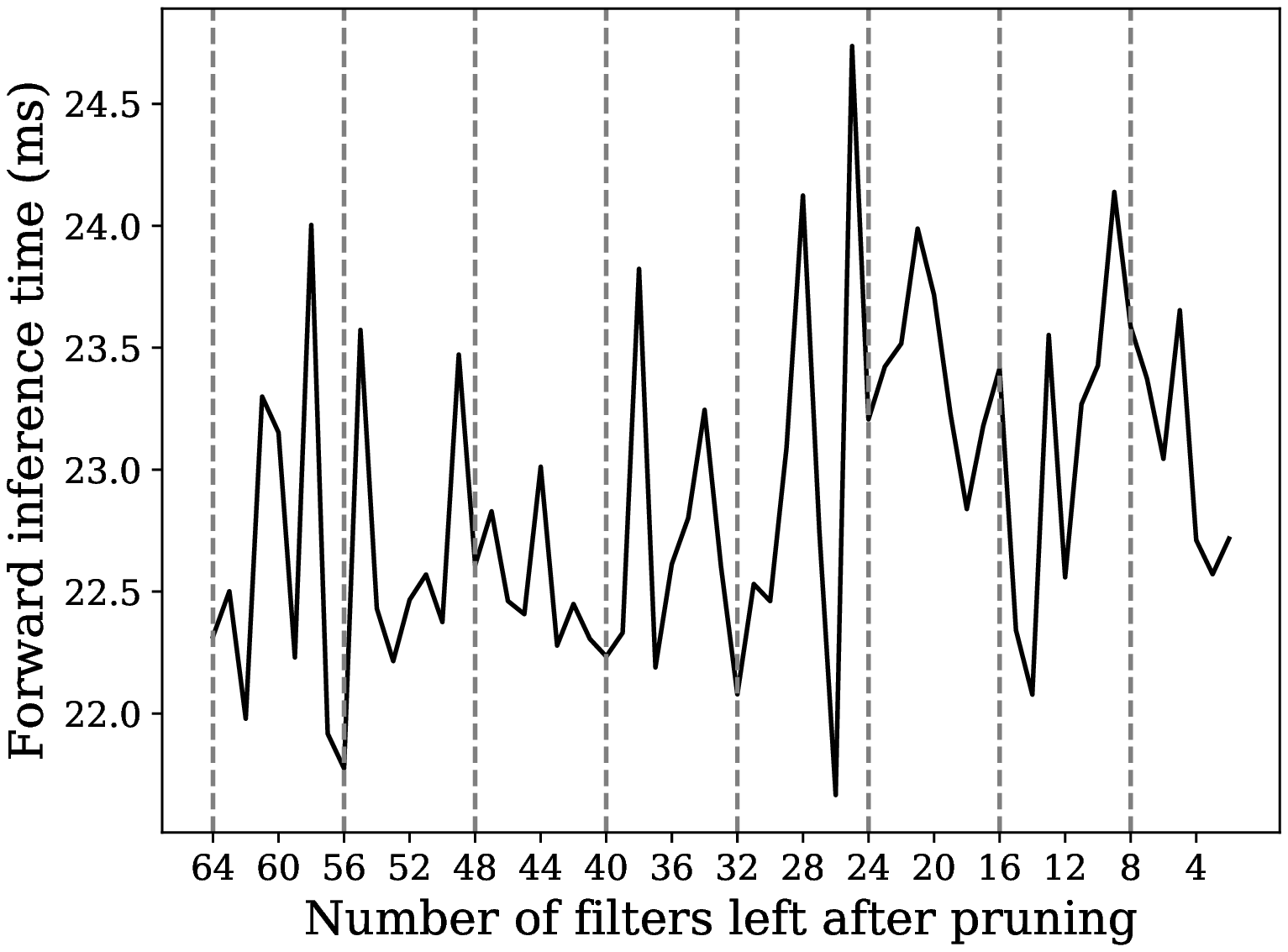}%
		\label{fig:latancy_conv1_gpu}}\hfill
	\subfloat[][NCS inference time: Conv2 pruned]
	{\includegraphics[width=0.30\textwidth]{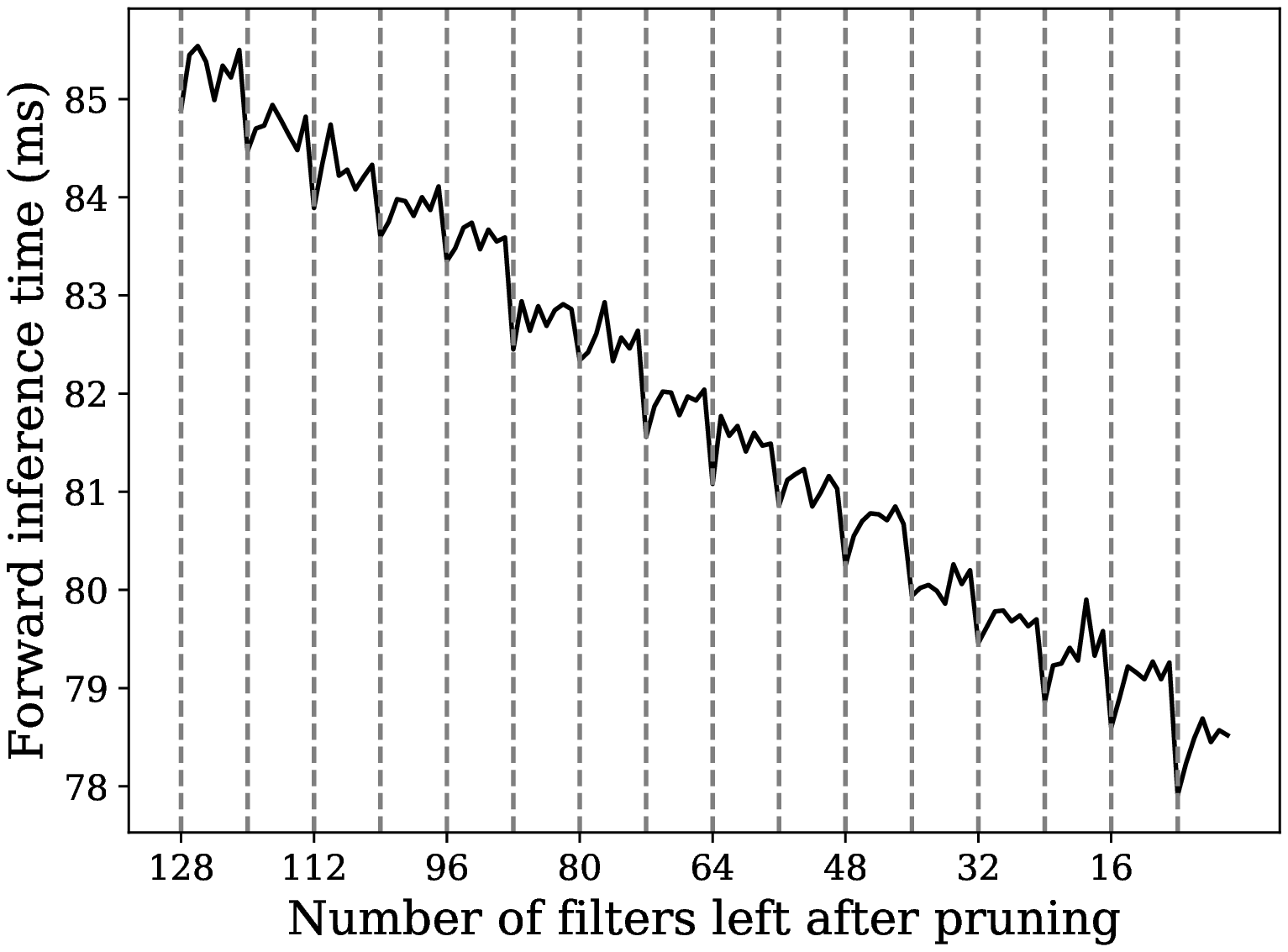}%
		\label{fig:latancy_conv2_ncs}}\hfill
	\subfloat[][CPU inference time: Conv2 pruned]
	{\includegraphics[width=0.30\textwidth]{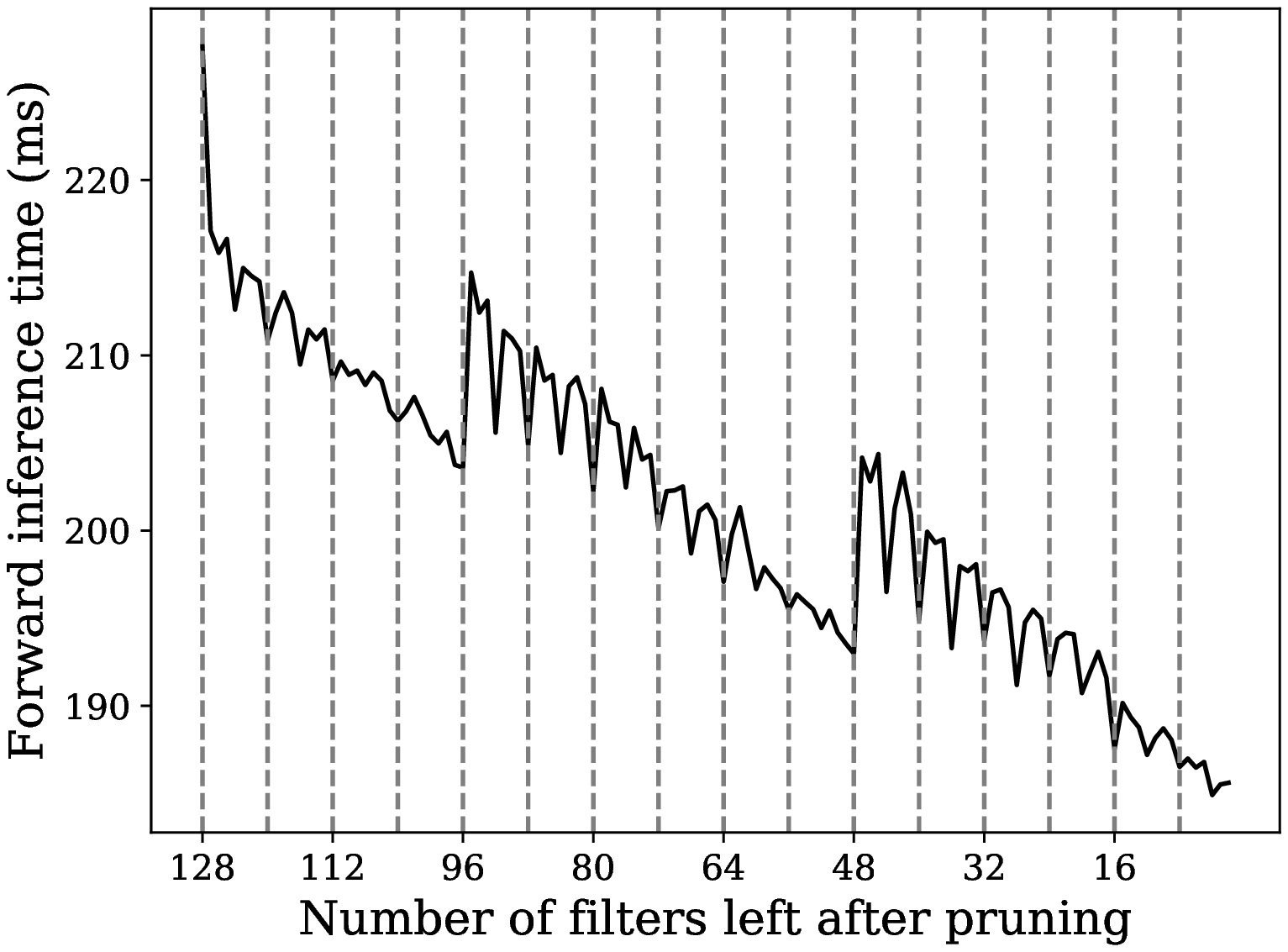}%
		\label{fig:latancy_conv2_cpu}}\hfill
	\subfloat[][GPU inference time: Conv2 pruned]
	{\includegraphics[width=0.30\textwidth]{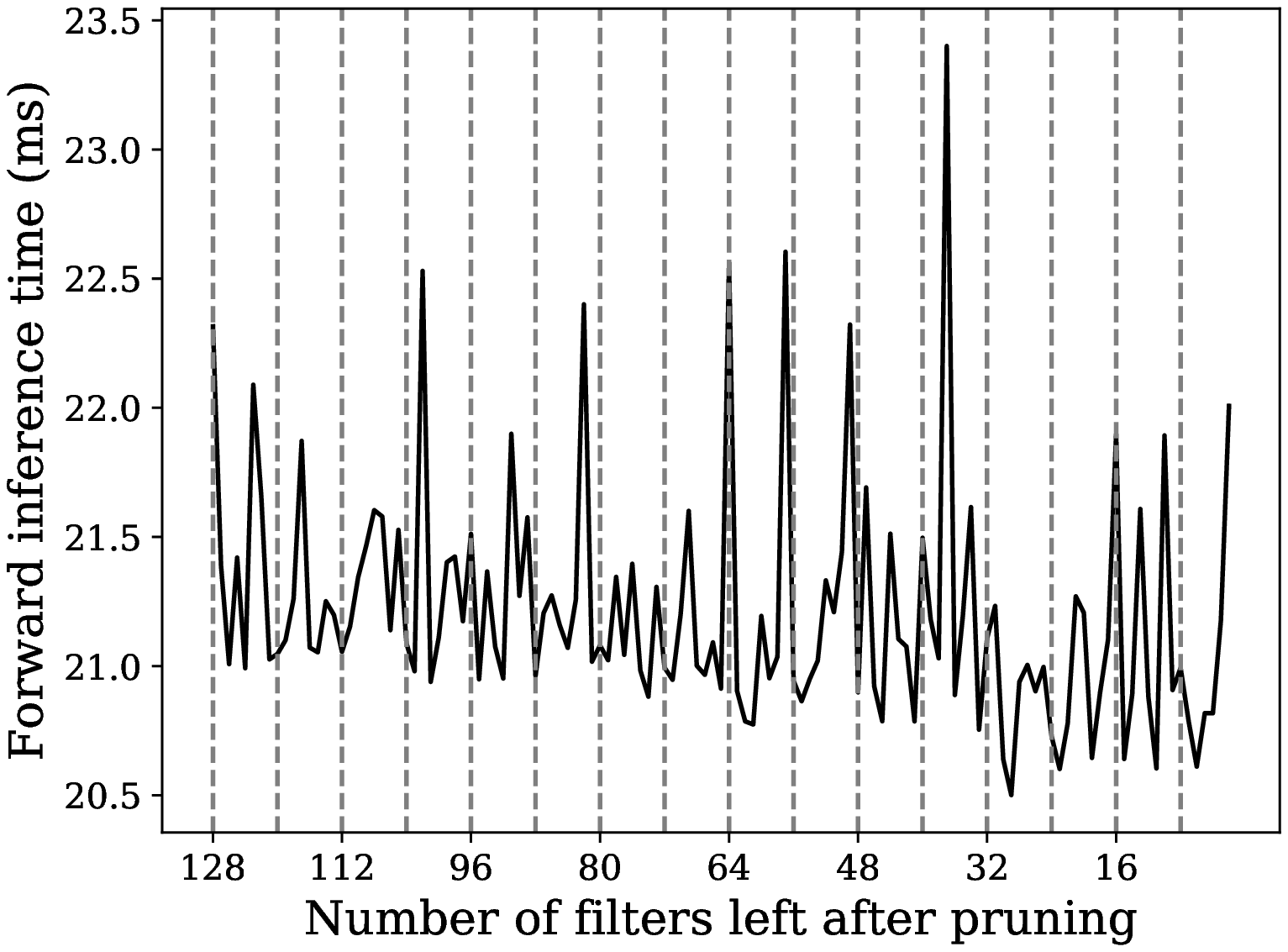}%
		\label{fig:latancy_conv2_gpu}}\hfil
	\caption{Latency through whole network (SSD-MobileNet): Single layer pruning using NCS, CPU, GPU}
	\label{fig:single_layer_all}
\end{figure*}

The main intention of this subsection is to determine the optimum cluster size used for cluster pruning as described in Section III Subsection A and B. We select first three convolution layers of SSD-MobileNet, which are named as \textit{Conv0}, \textit{Conv1}, \textit{Conv2}, and prune the filters inside them.  We prune one filter in a single layer at a time until two filters are left in that layer. The number of input channels of the next layer is reduced once we prune a filter in the current layer. Thus, corresponding kernels inside the filters in next layer is also pruned. Moreover, SSD-MobileNet is designed with depthwise convolution architecture, where convolutional layers are separated into two layers called pointwise and depthwise convolutions. Therefore, when we prune layer \textit{Conv0}, corresponding kernels inside filters of the depthwise convolutional layer \textit{Conv1/dw} and pointwise convolutional layer \textit{Conv1} are also pruned. As a result of that, three adjacent layers of the network is pruned at given filter pruning iteration. Then we measure the forward inference time of the network and accuracy for the test datasets at each iteration. Fig. \ref{fig:latancy_conv0_ncs}, \ref{fig:latancy_conv0_cpu}, \ref{fig:latancy_conv0_gpu} show the latency results after pruning the layer \textit{Conv0} of the SSD-MobileNet using three hardware architectures NCS, CPU, and GPU, respectively. Fig. \ref{fig:latancy_conv1_ncs}, \ref{fig:latancy_conv1_cpu}, \ref{fig:latancy_conv1_gpu} and Fig. \ref{fig:latancy_conv2_ncs}, \ref{fig:latancy_conv2_cpu}, \ref{fig:latancy_conv2_gpu} indicate pruning of the layers \textit{Conv1} and \textit{Conv2} of the SSD-MobileNet, respectively.

\begin{figure*}
	\centering
	\subfloat[][NCS inference time: Conv1 pruned]
	{\includegraphics[width=0.30\textwidth]{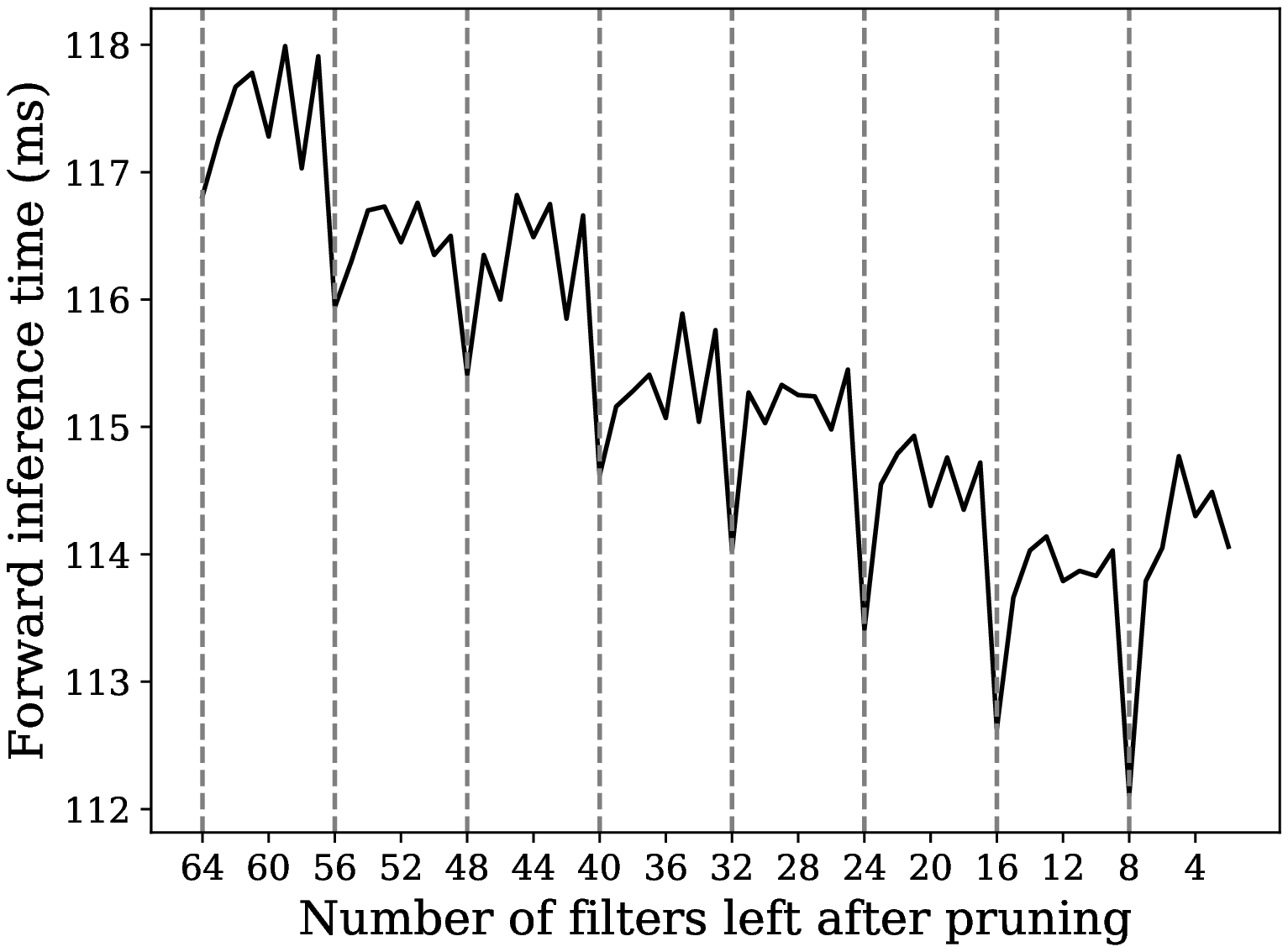}%
		\label{fig:latancy_sqconv1_ncs}}\hfill
	\subfloat[][CPU inference time: Conv1 pruned]
	{\includegraphics[width=0.30\textwidth]{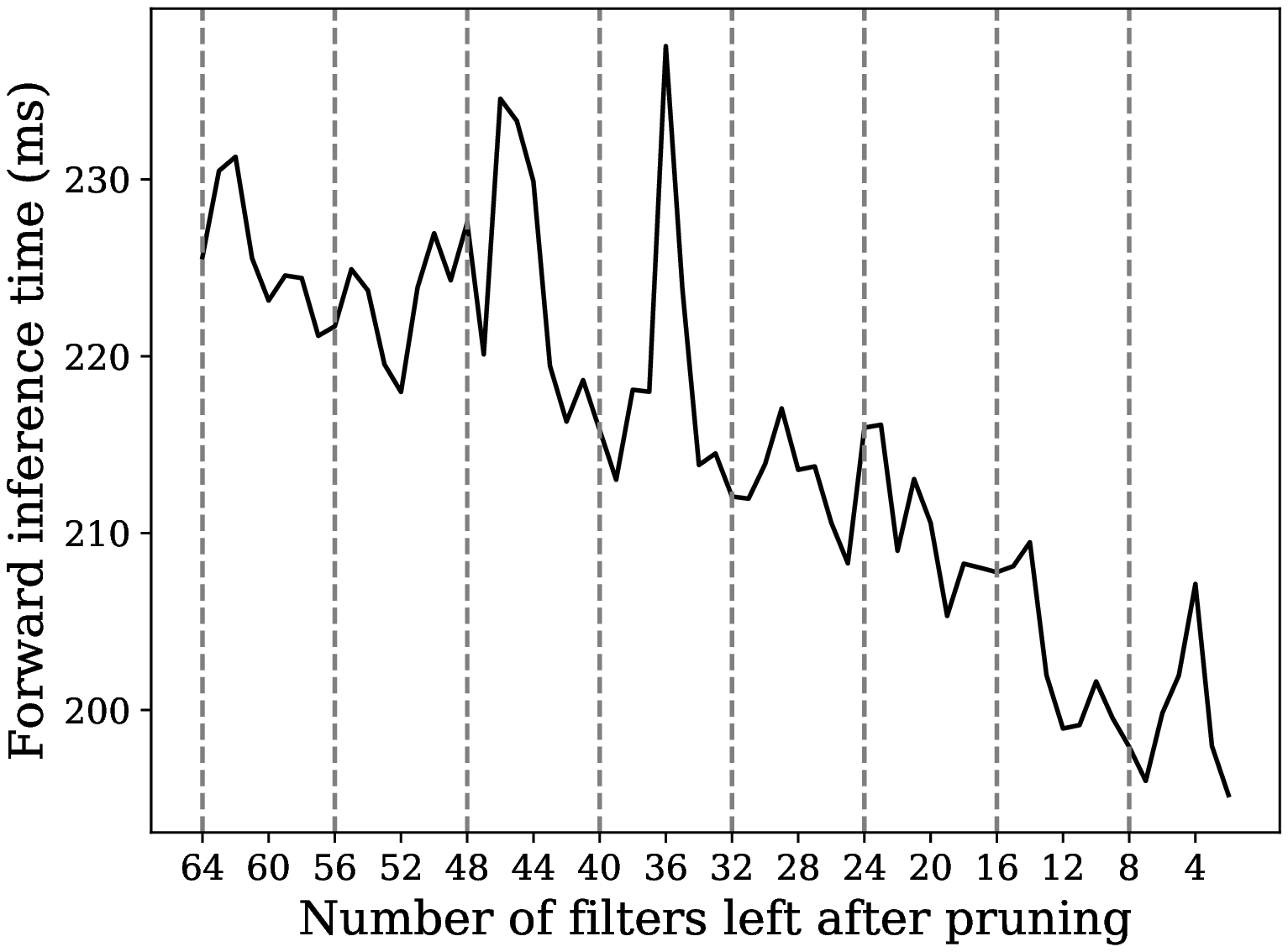}%
		\label{fig:latancy_sqconv1_cpu}}\hfill
	\subfloat[][GPU inference time: Conv1 pruned]
	{\includegraphics[width=0.30\textwidth]{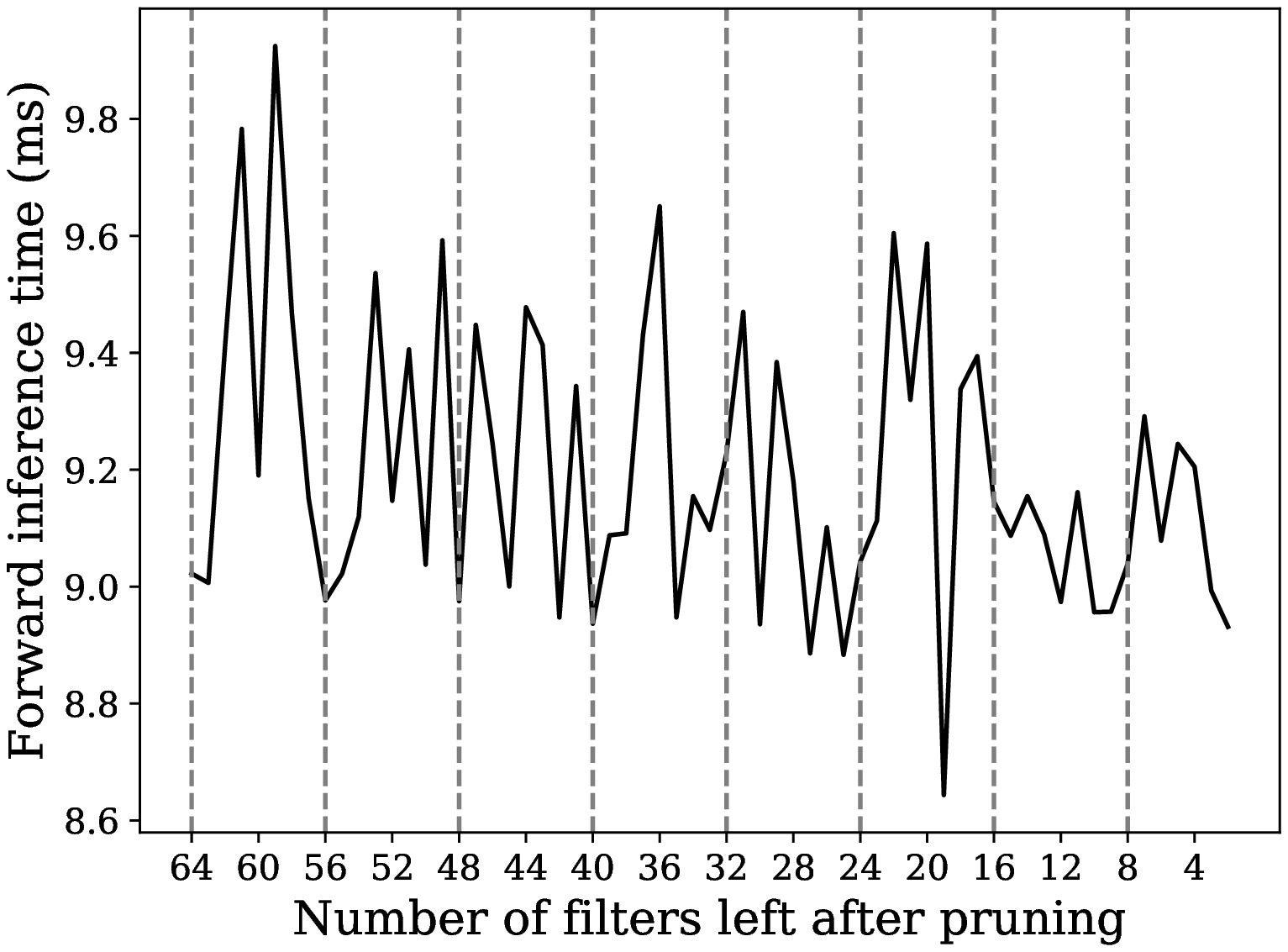}%
		\label{fig:latancy_sqconv1_gpu}}\hfill
	\subfloat[][NCS inference time: Fire2/Expand1x1 pruned]
	{\includegraphics[width=0.30\textwidth]{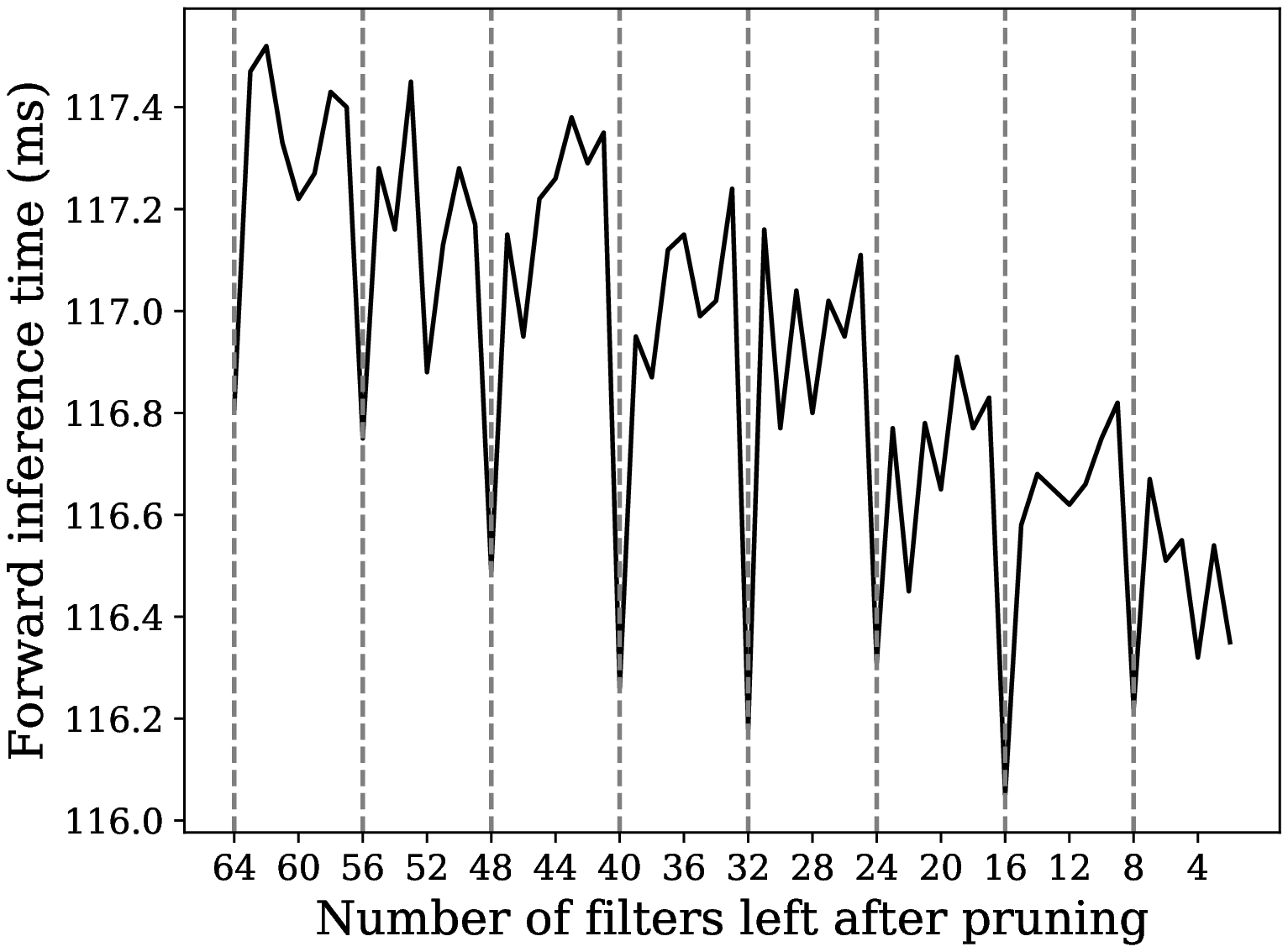}%
		\label{fig:latancy_sqfire2exp1_ncs}}\hfill
	\subfloat[][CPU inference time: Fire2/Expand1x1 pruned]
	{\includegraphics[width=0.30\textwidth]{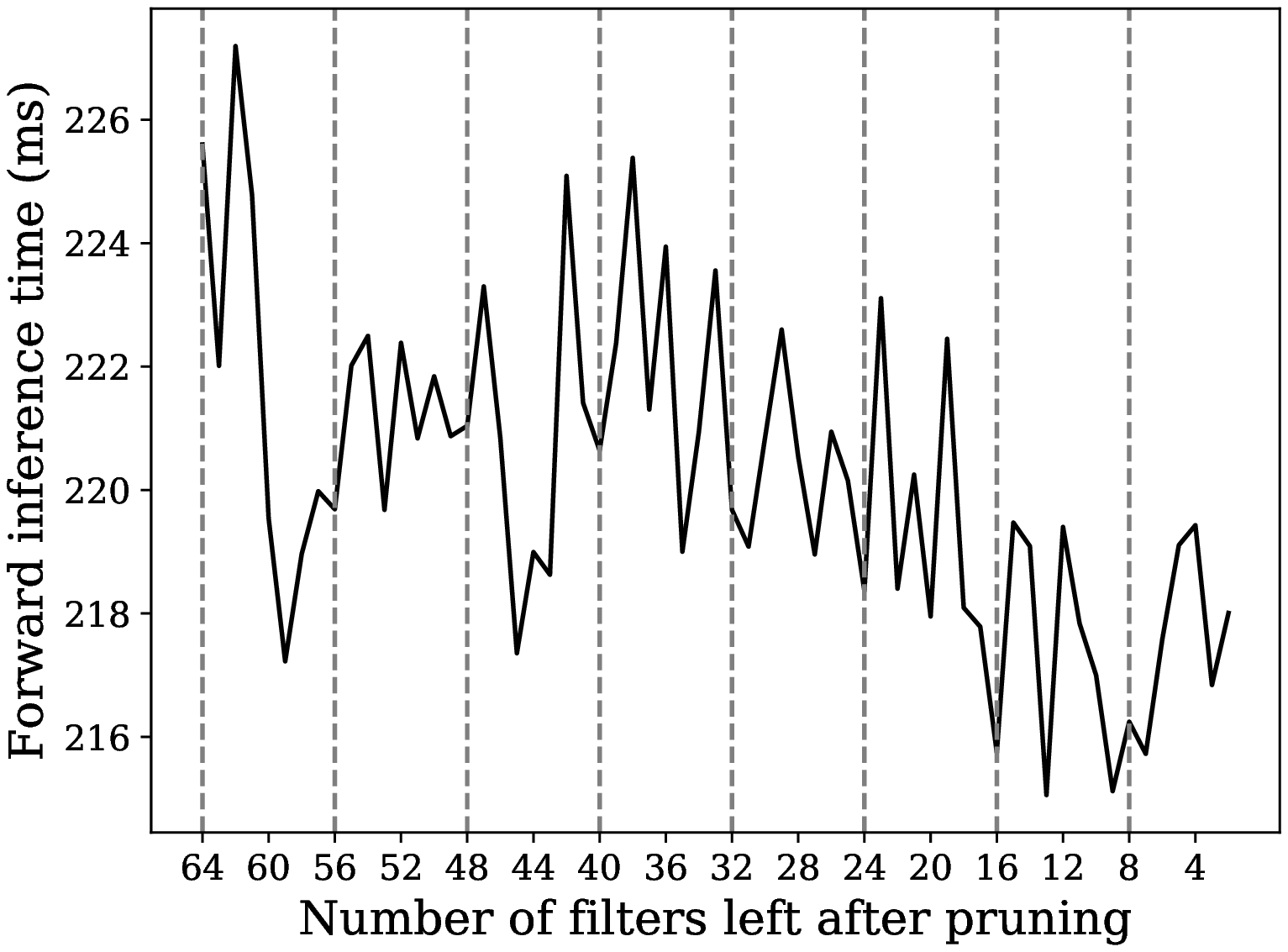}%
		\label{fig:latancy_sqfire2exp1_cpu}}\hfill
	\subfloat[][GPU inference time: Fire2/Expand1x1 pruned]
	{\includegraphics[width=0.30\textwidth]{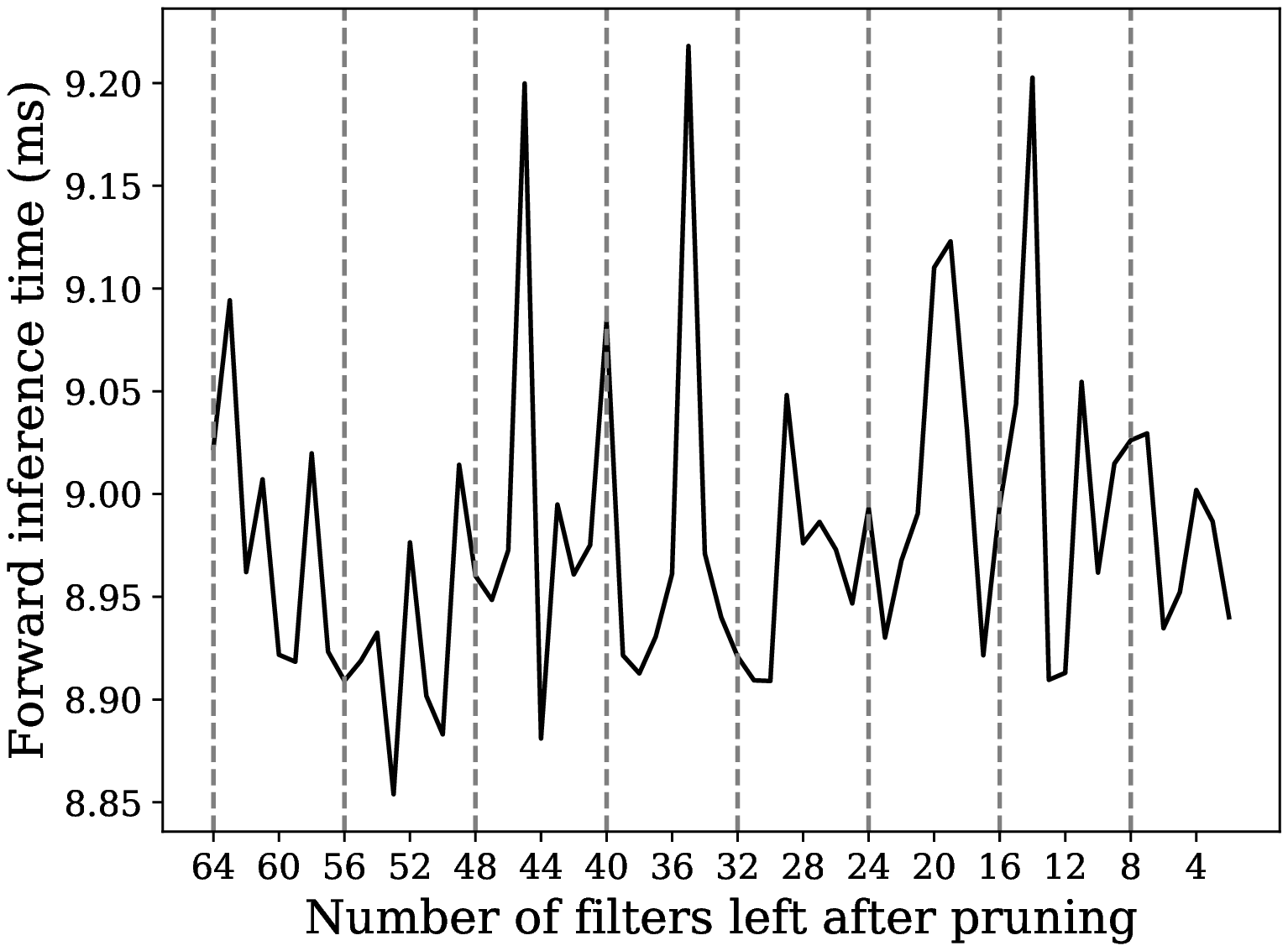}%
		\label{fig:latancy_sqfire2exp1_gpu}}\hfill
	\subfloat[][NCS inference time: Fire2/Expand3x3 pruned]
	{\includegraphics[width=0.30\textwidth]{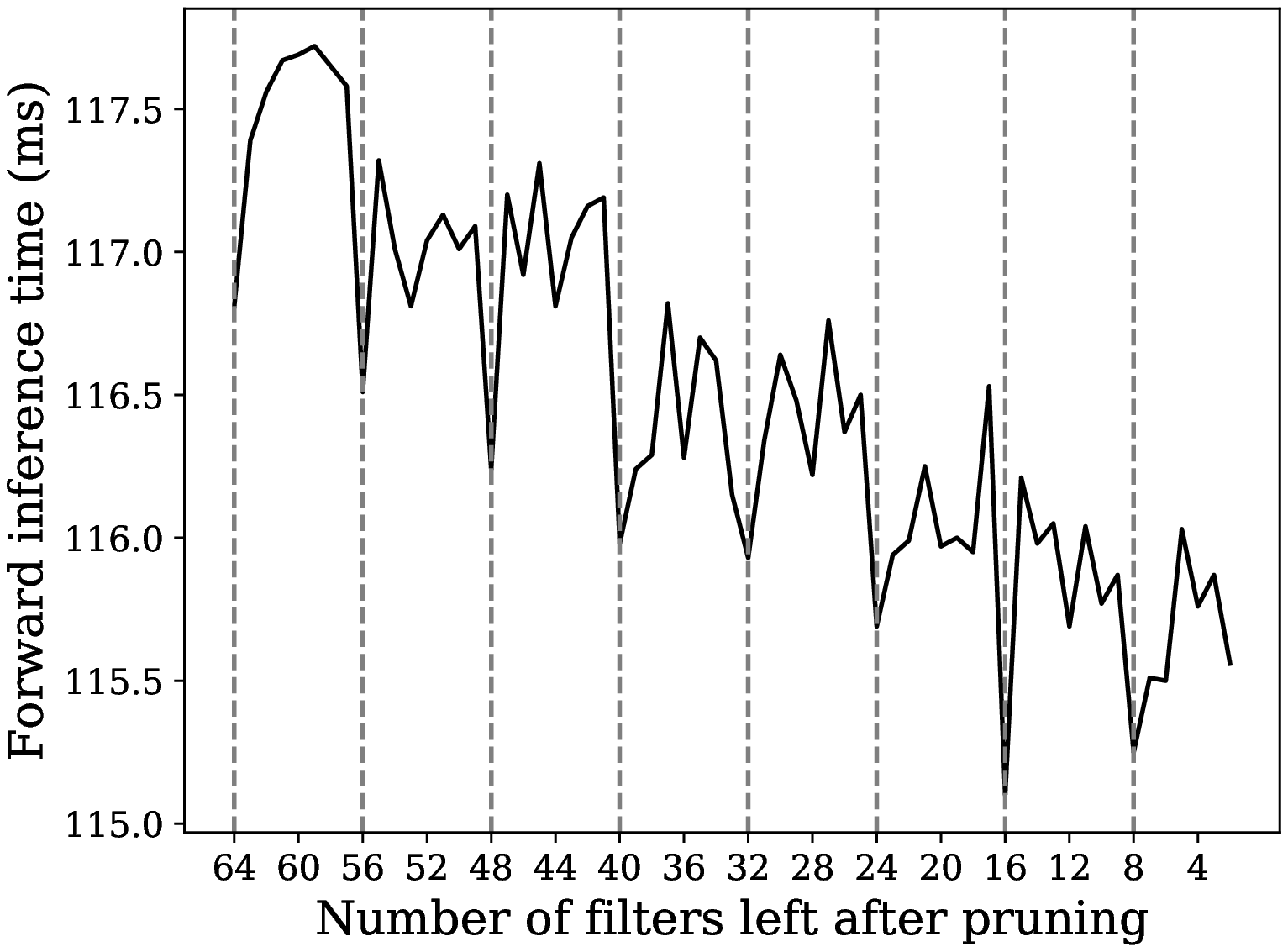}%
		\label{fig:latancy_sqfire2exp3_ncs}}\hfill
	\subfloat[][CPU inference time: Fire2/Expand3x3 pruned]
	{\includegraphics[width=0.30\textwidth]{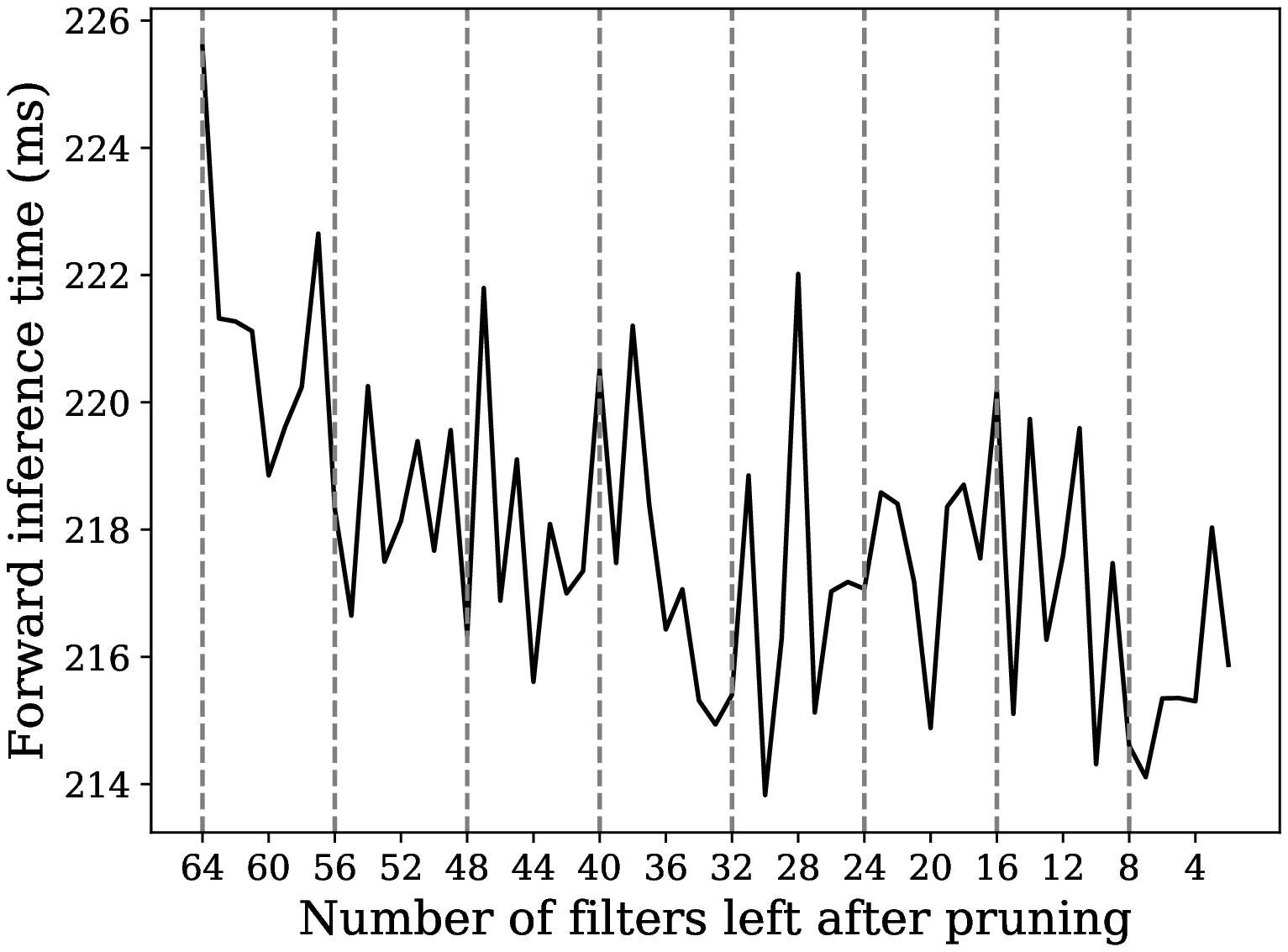}%
		\label{fig:latancy_sqfire2exp3_cpu}}\hfill
	\subfloat[][GPU inference time: Fire2/Expand3x3 pruned]
	{\includegraphics[width=0.30\textwidth]{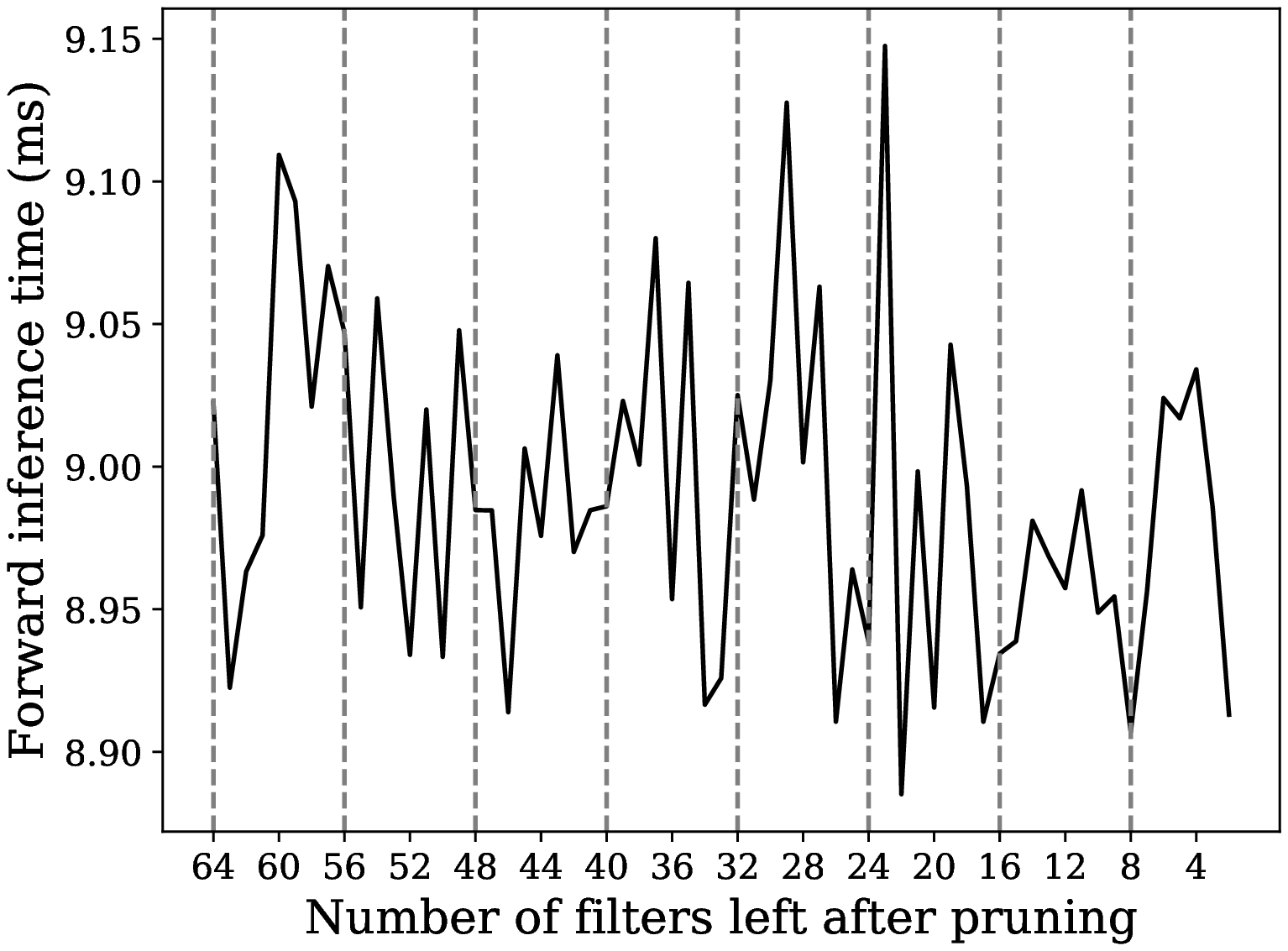}%
		\label{fig:latancy_sqfire2exp3_gpu}}\hfil
	\caption{Latency through whole network (SSD-SqueezeNet): Single layer pruning using NCS, CPU, GPU}
	\label{fig:single_layer_all_sq}
\end{figure*}

We select three convolution layers named as \textit{Conv1}, \textit{Fire2/Expand1x1}, and \textit{Fire2/Expand3x3} for the pruning of SSD-SqueezeNet. Once we prune a filter in SSD-SqueezeNet, only the corresponding channel of the next layers is pruned at a given filter pruning iteration. Fig. \ref{fig:latancy_sqconv1_ncs}, \ref{fig:latancy_sqconv1_cpu}, \ref{fig:latancy_sqconv1_gpu} show the inference latency results after pruning the layer \textit{Conv1} of the SSD-SqueezeNet using three hardware architectures NCS, CPU, and GPU, respectively. Fig. \ref{fig:latancy_sqfire2exp1_ncs}, \ref{fig:latancy_sqfire2exp1_cpu}, \ref{fig:latancy_sqfire2exp1_gpu} and Fig. \ref{fig:latancy_sqfire2exp3_ncs}, \ref{fig:latancy_sqfire2exp3_cpu}, \ref{fig:latancy_sqfire2exp3_gpu} indicate pruning of the layers \textit{Fire2/Expand1x1} and \textit{Fire2/Expand3x3}, respectively.

\begin{figure*}
	\centering
	\subfloat[][Latency through Conv0: Conv0 pruned ]
	{\includegraphics[width=0.30\textwidth]{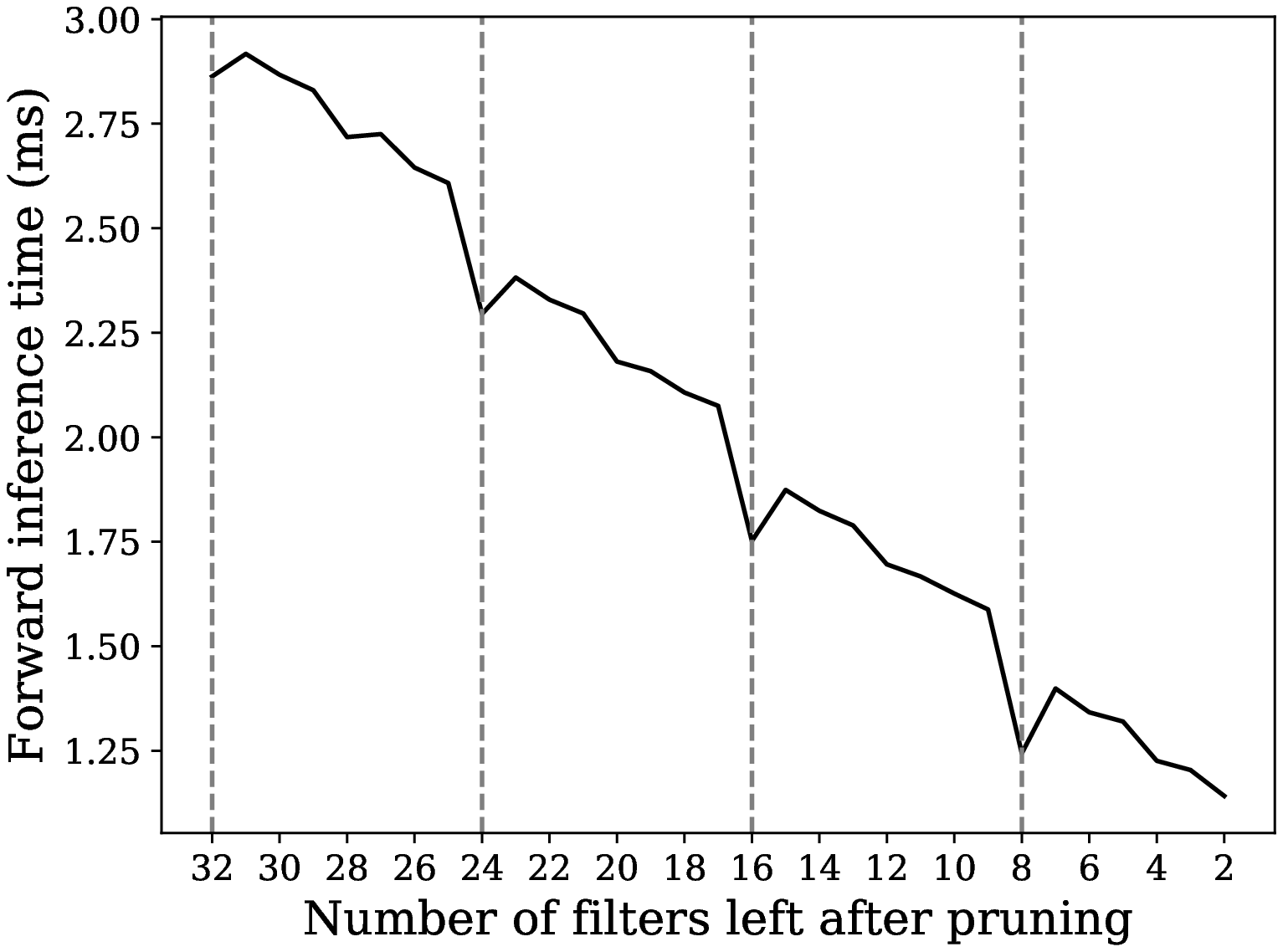}%
		\label{fig:latancy_conv0_conv0}}\hfill
	\subfloat[][Latency through Conv1/dw: Conv0 pruned]
	{\includegraphics[width=0.30\textwidth]{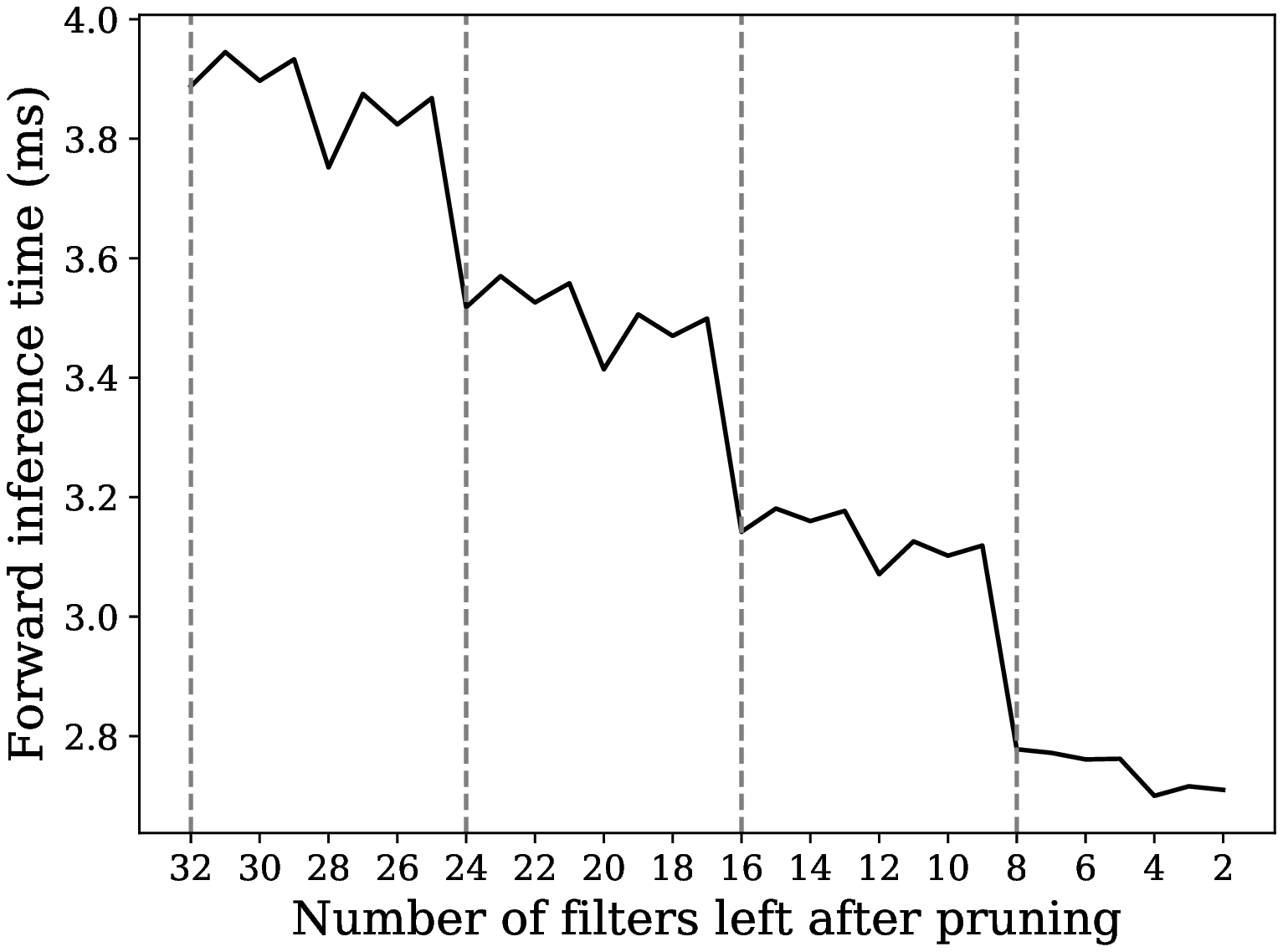}%
		\label{fig:latancy_conv0_conv1dw}}\hfill
	\subfloat[][Latency through Conv1: Conv0 pruned]
	{\includegraphics[width=0.30\textwidth]{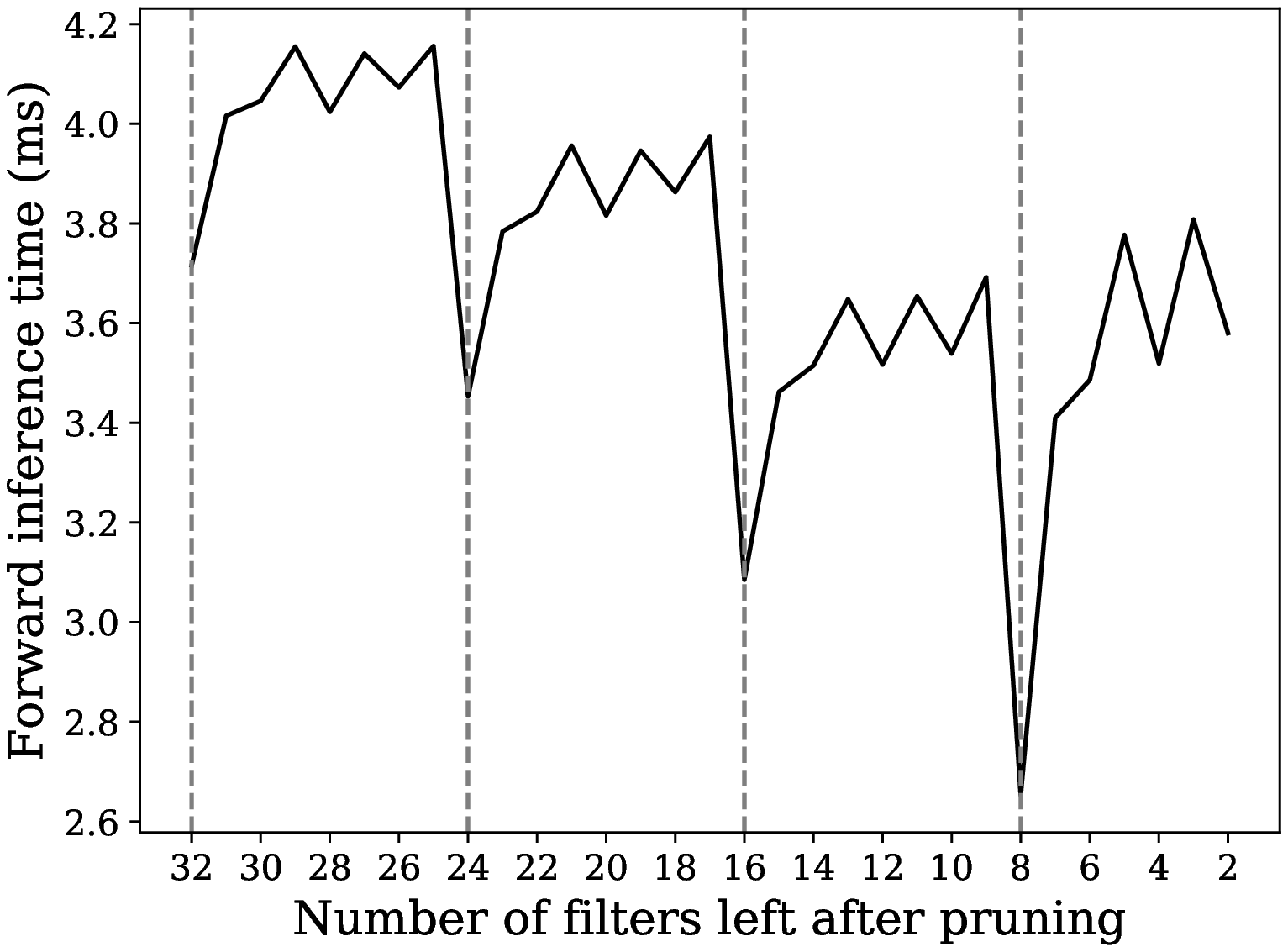}%
		\label{fig:latancy_conv0_conv1}}\hfill
	\subfloat[][Latency through Conv1: Conv1 pruned]
	{\includegraphics[width=0.30\textwidth]{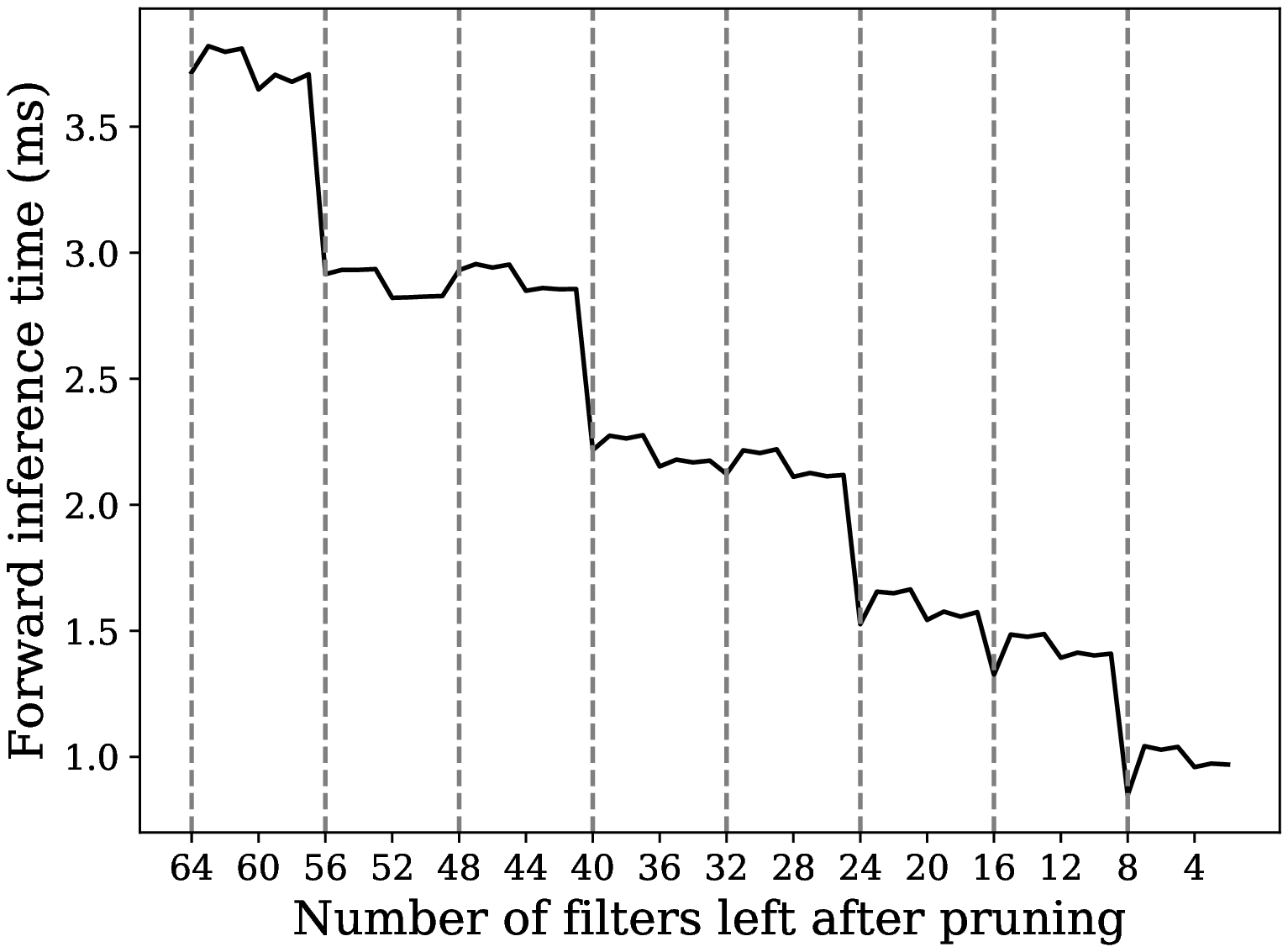}%
		\label{fig:latancy_conv1_conv1}}\hfill
	\subfloat[][Latency through Conv2/dw: Conv1 pruned]
	{\includegraphics[width=0.30\textwidth]{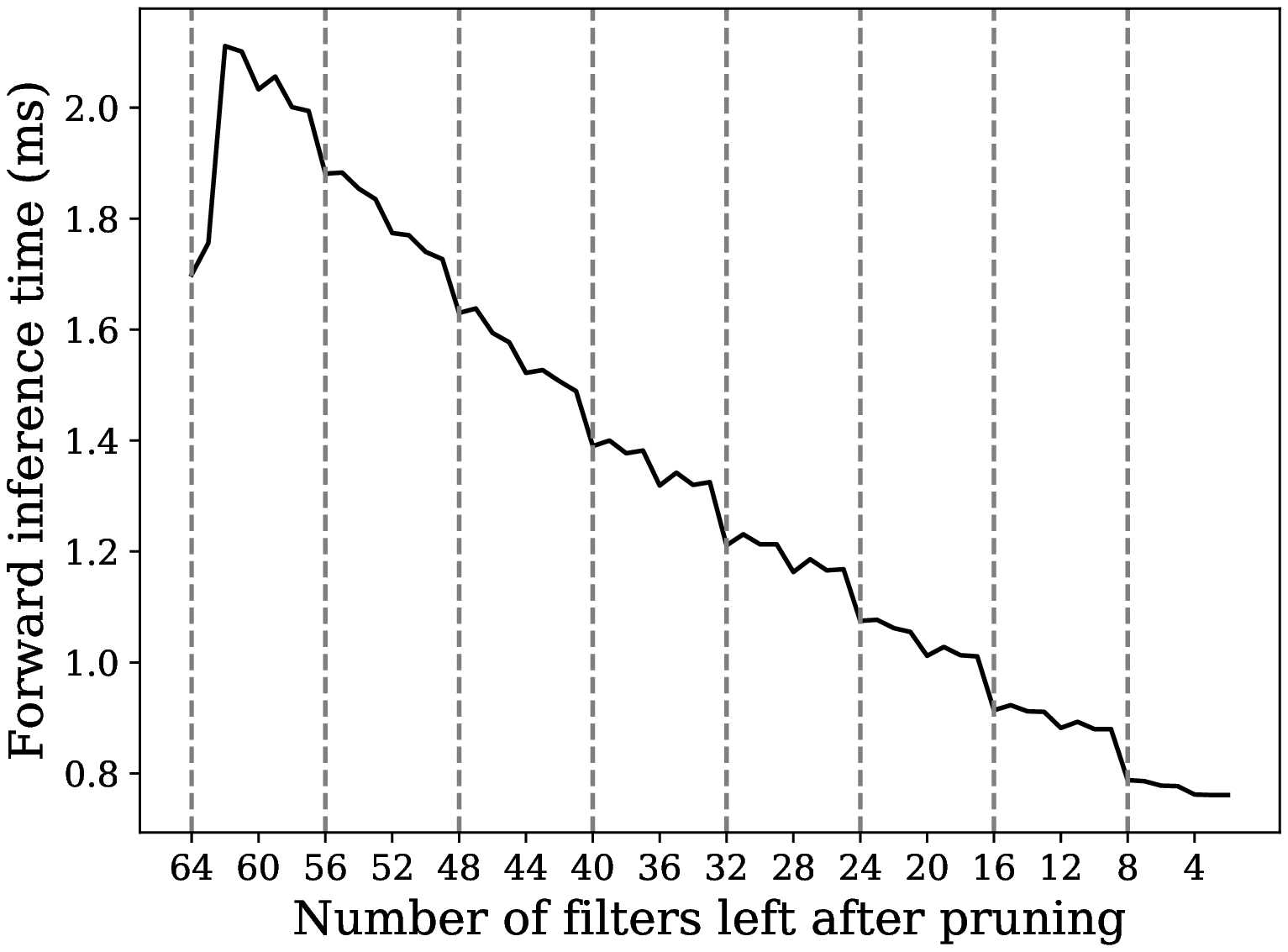}%
		\label{fig:latancy_conv1_conv2dw}}\hfill
	\subfloat[][Latency through Conv2: Conv1 pruned]
	{\includegraphics[width=0.30\textwidth]{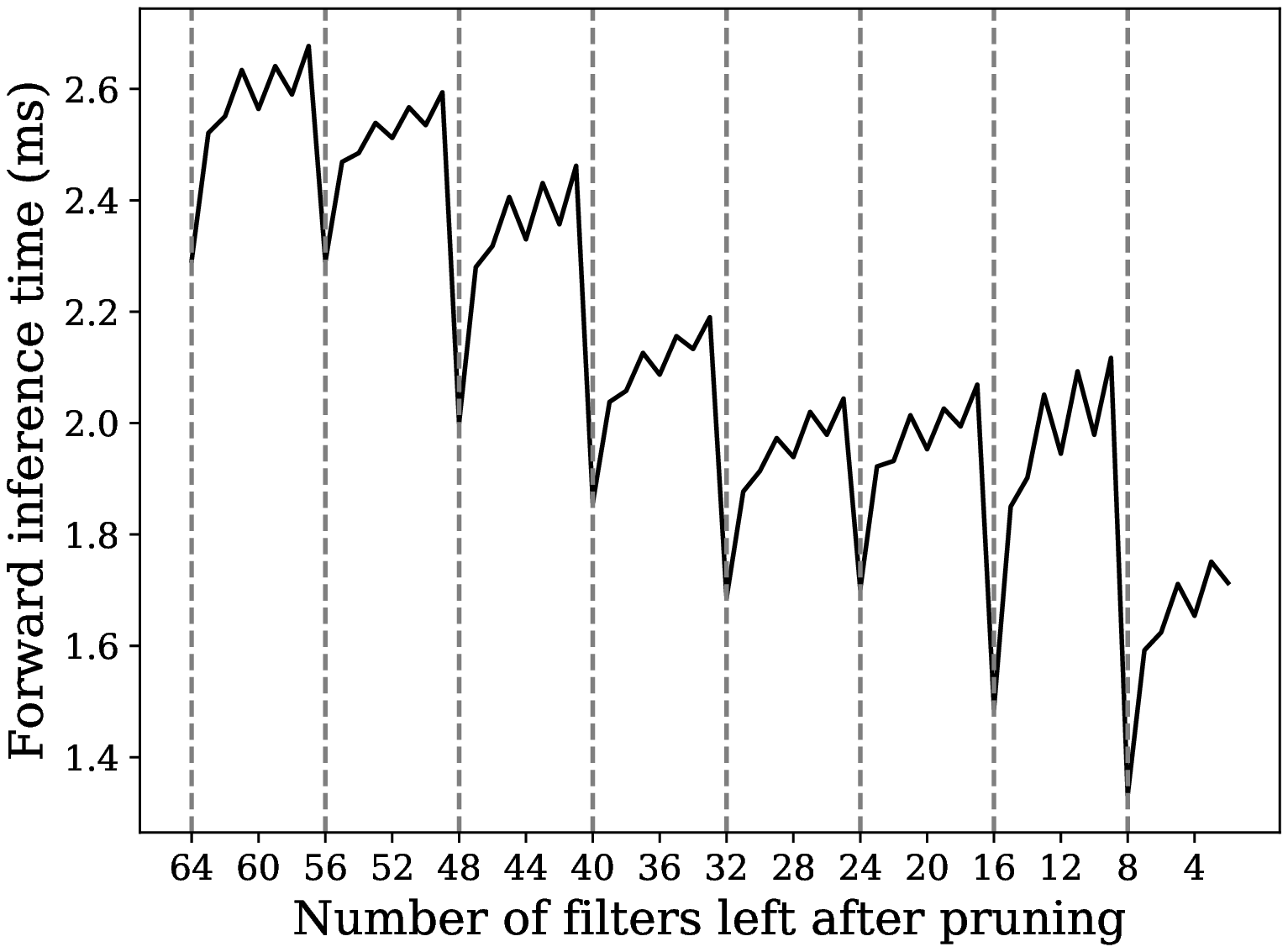}%
		\label{fig:latancy_conv1_conv2}}\hfill
	\subfloat[][Latency through Conv2: Conv2 pruned]
	{\includegraphics[width=0.30\textwidth]{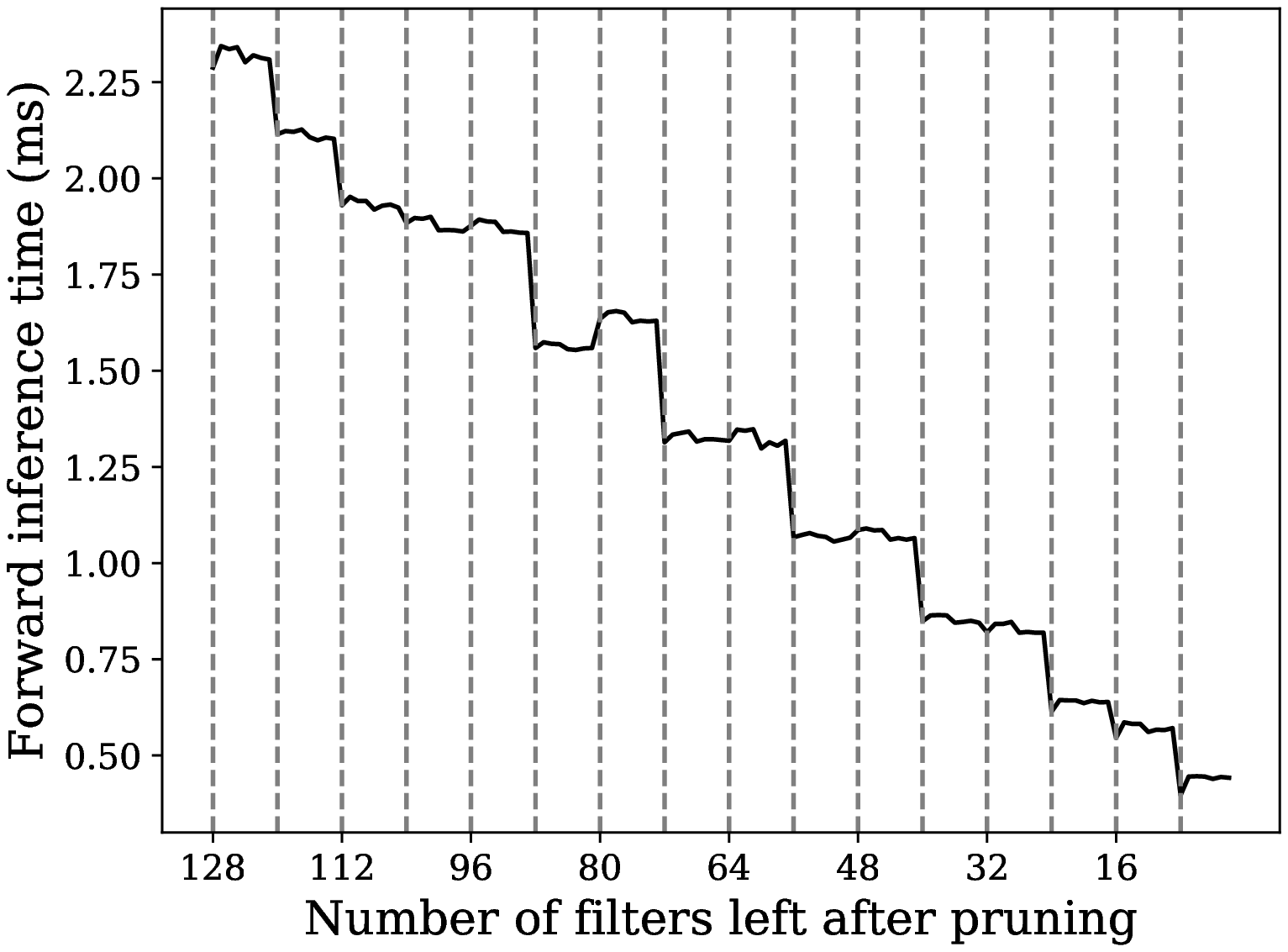}%
		\label{fig:latancy_conv2_conv2}}\hfill
	\subfloat[][Latency through Conv2/dw: Conv2 pruned]
	{\includegraphics[width=0.30\textwidth]{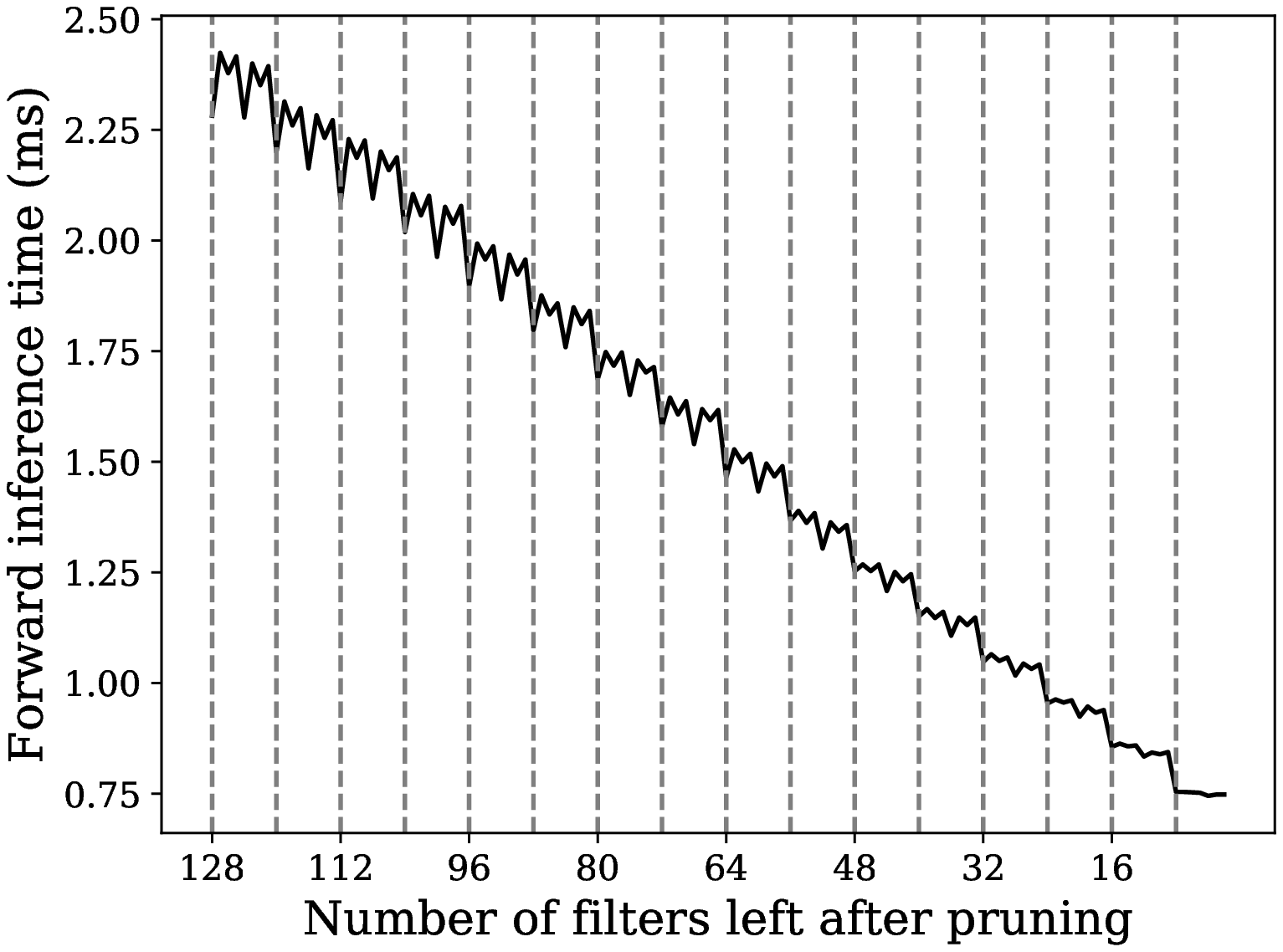}%
		\label{fig:latancy_conv2_conv3dw}}\hfill
	\subfloat[][Latency through Conv3: Conv2 pruned]
	{\includegraphics[width=0.30\textwidth]{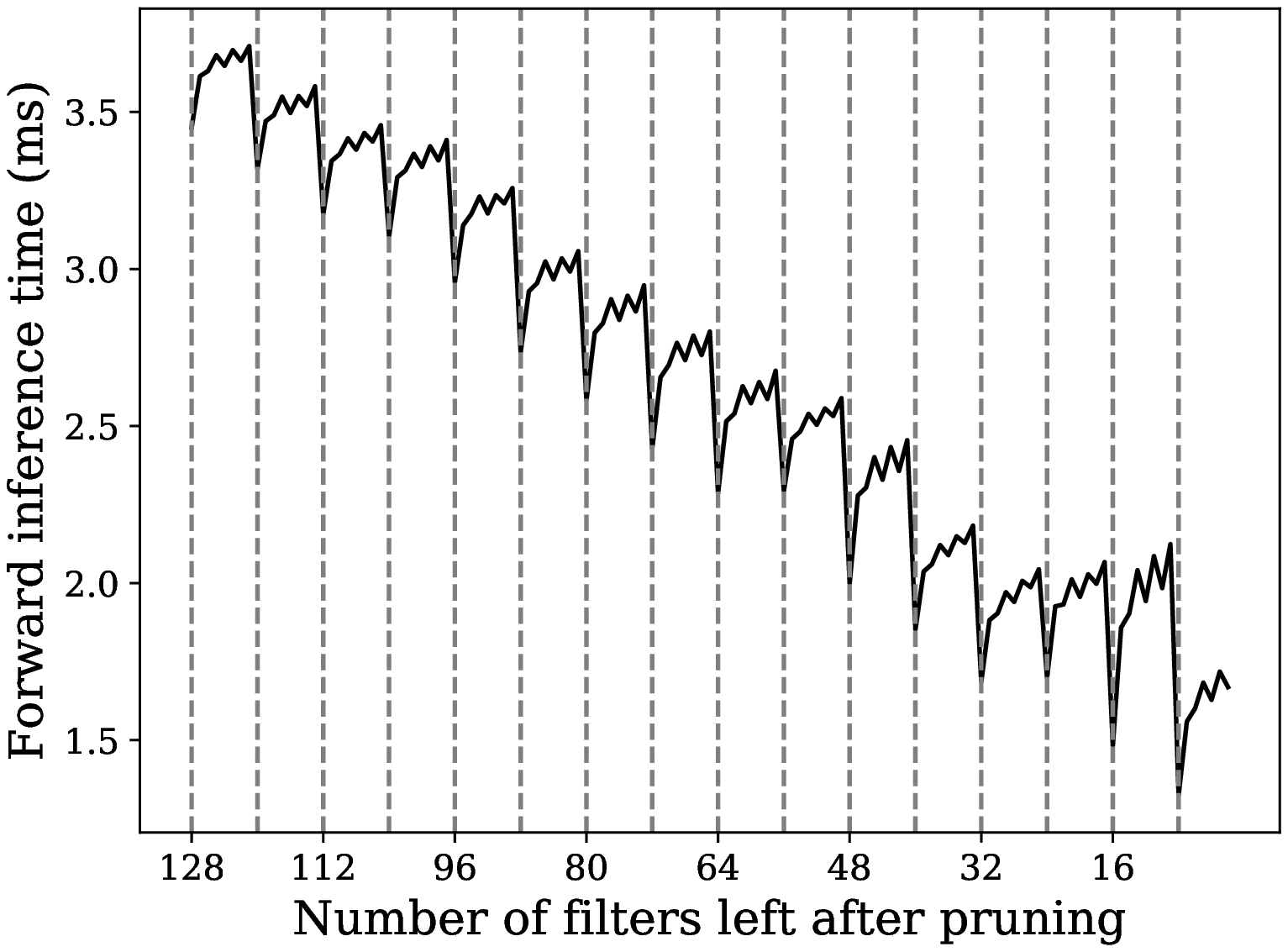}%
		\label{fig:latancy_conv2_conv3}}\hfill
	\caption{Latency through individual layers (SSD-MobileNet): Single layer pruning using NCS}
	\label{fig:single_layer_ncs}
\end{figure*}

\begin{figure*}
	\centering
	\subfloat[][Accuracy after pruning layer: Conv0]{\includegraphics[width=0.39\textwidth]{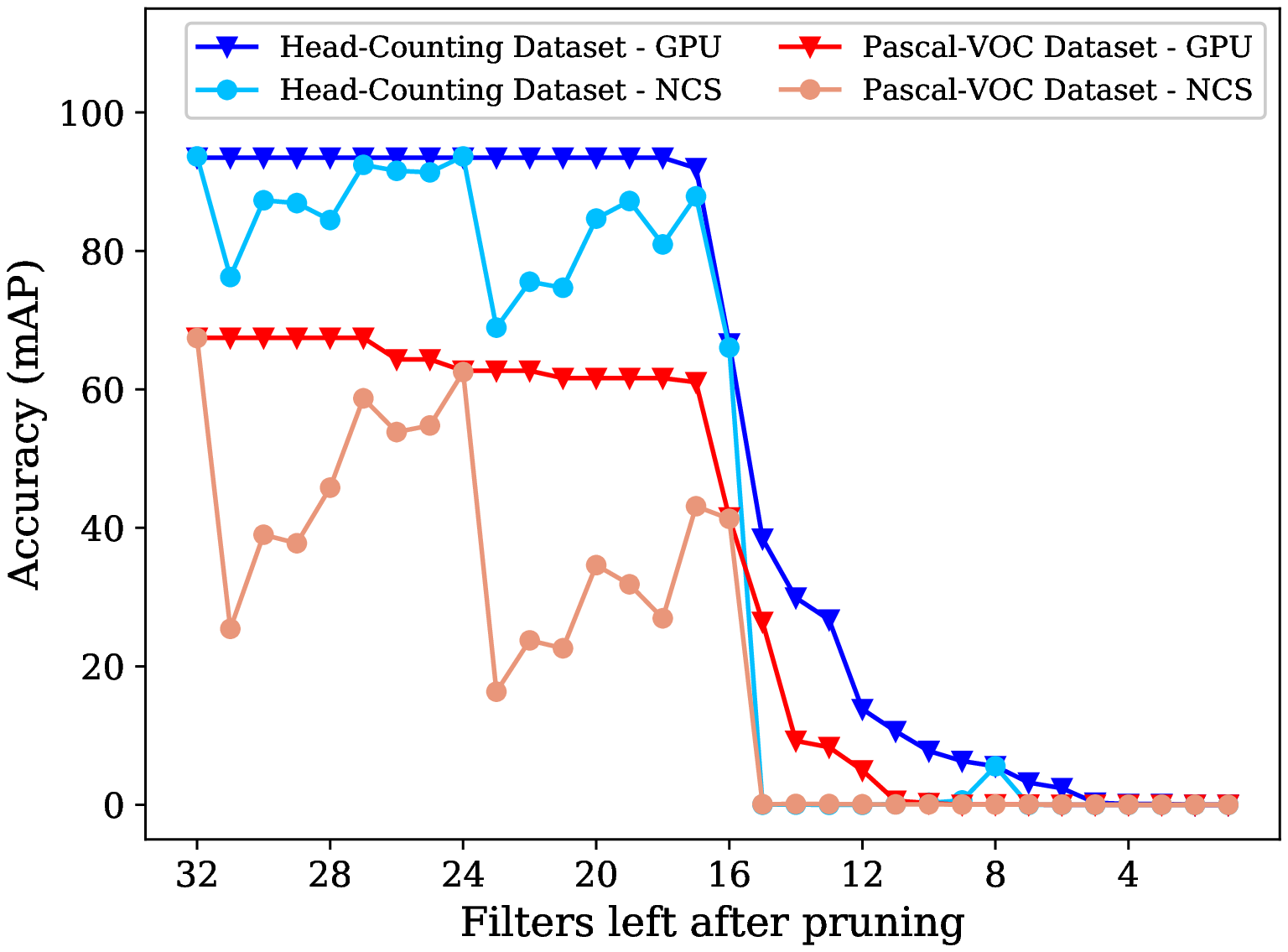}%
		\label{fig:accuracy_conv0}}
	\subfloat[][Accuracy after pruning layer: Conv1]{\includegraphics[width=0.39\textwidth]{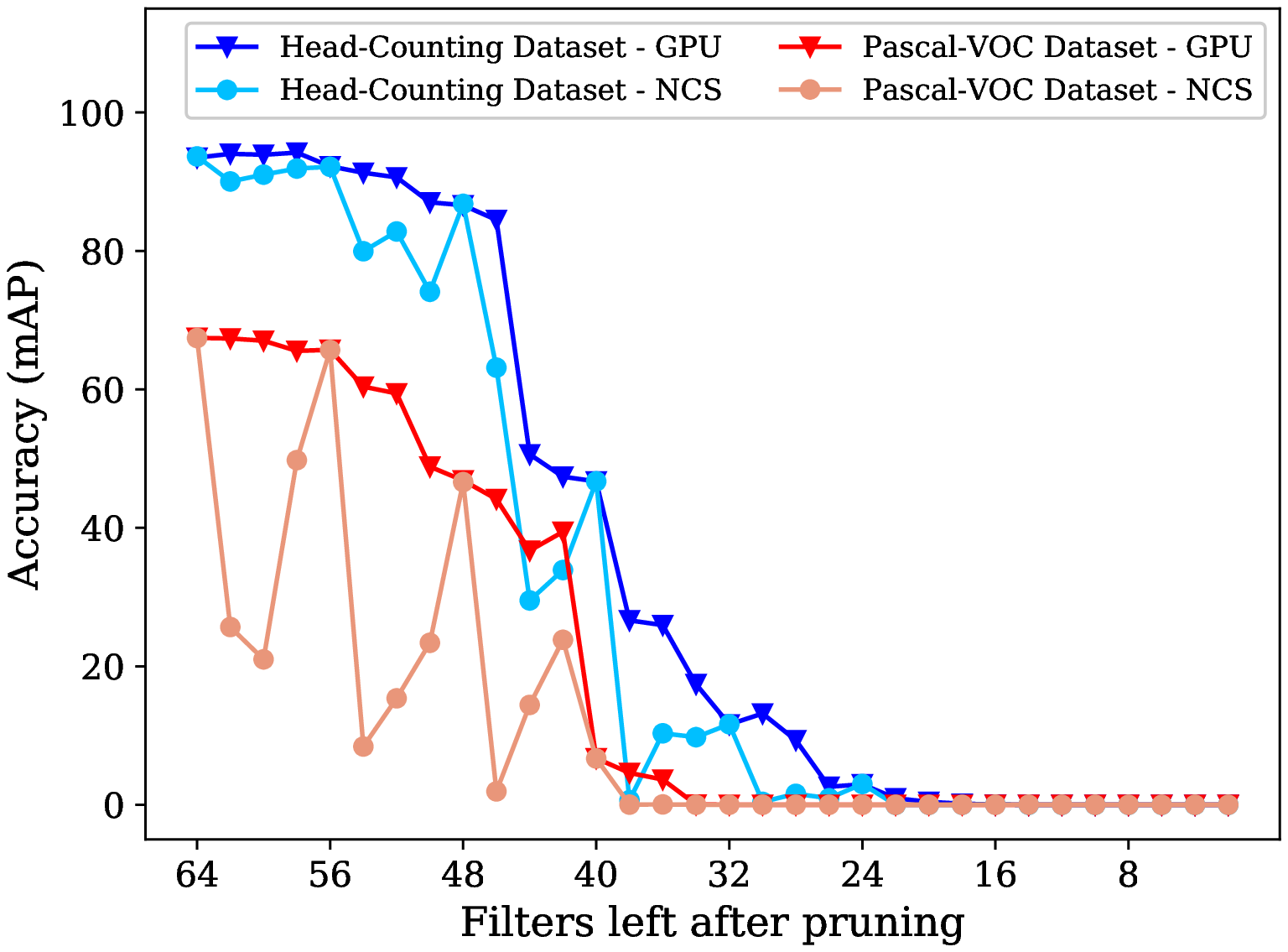}%
		\label{fig:accuracy_conv1}}\hfill
	\subfloat[][Accuracy after pruning layer: Conv2]{\includegraphics[width=0.39\textwidth]{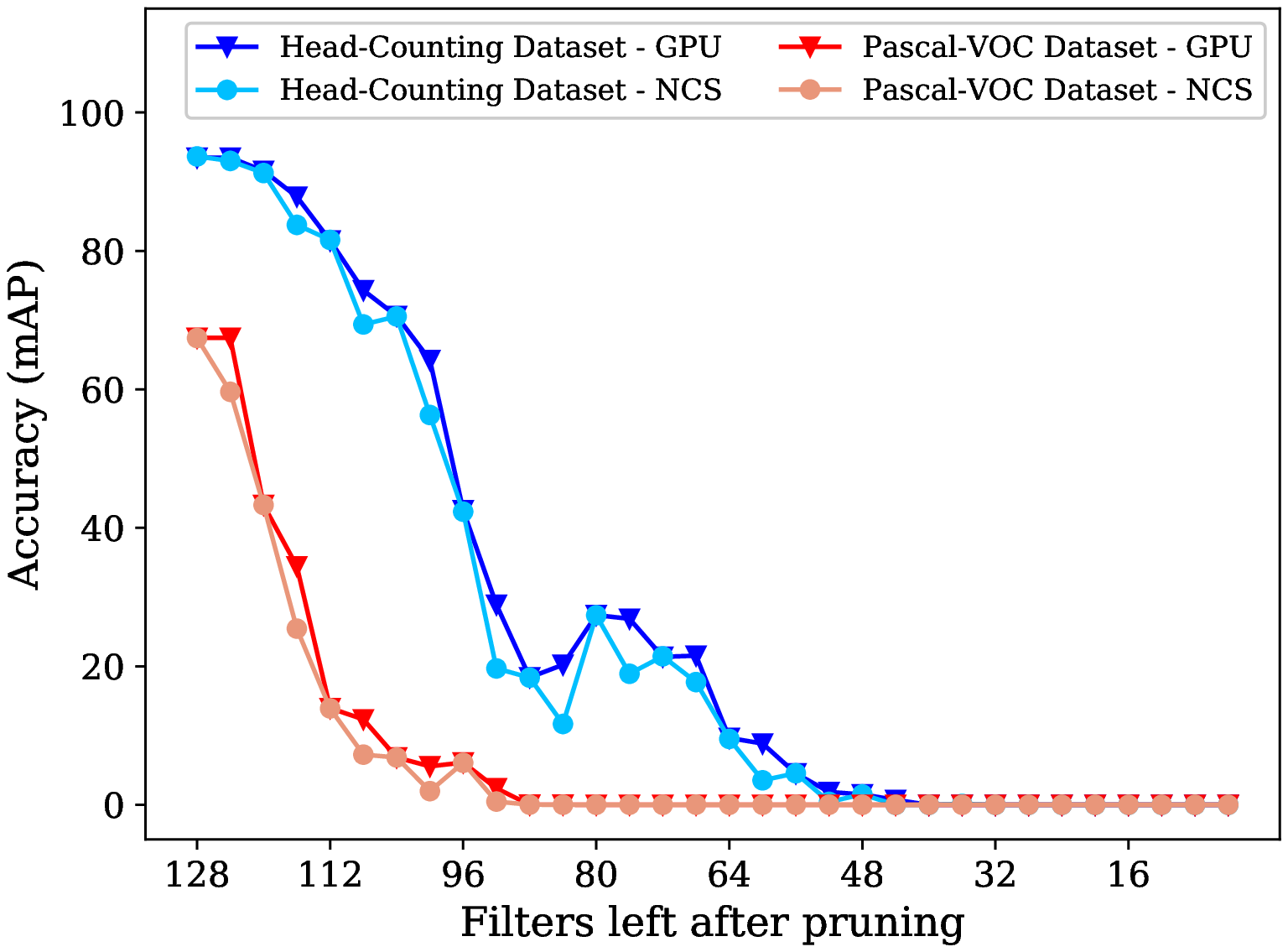}%
		\label{fig:accuracy_conv2}}
	\subfloat[][Accuracy after pruning layer: Conv6]{\includegraphics[width=0.39\textwidth]{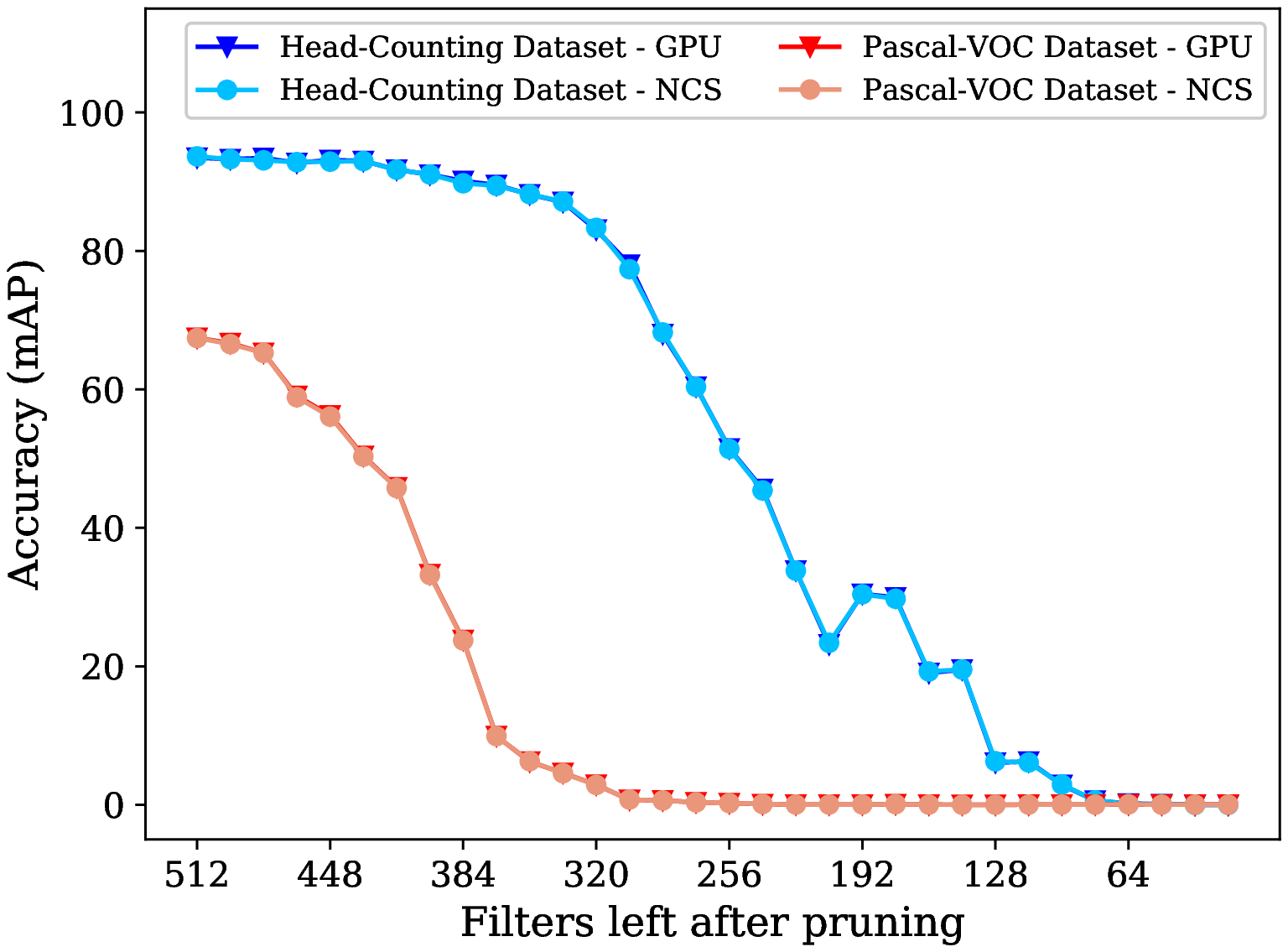}%
		\label{fig:accuracy_conv6}}
	\caption{Accuracy for Single layer pruning - SSD-MobileNet}
	\label{fig:accuracy_single_layer}
\end{figure*}

\begin{figure*}
	\centering
	\subfloat[][Accuracy after pruning layer: Conv1]{\includegraphics[width=0.39\textwidth]{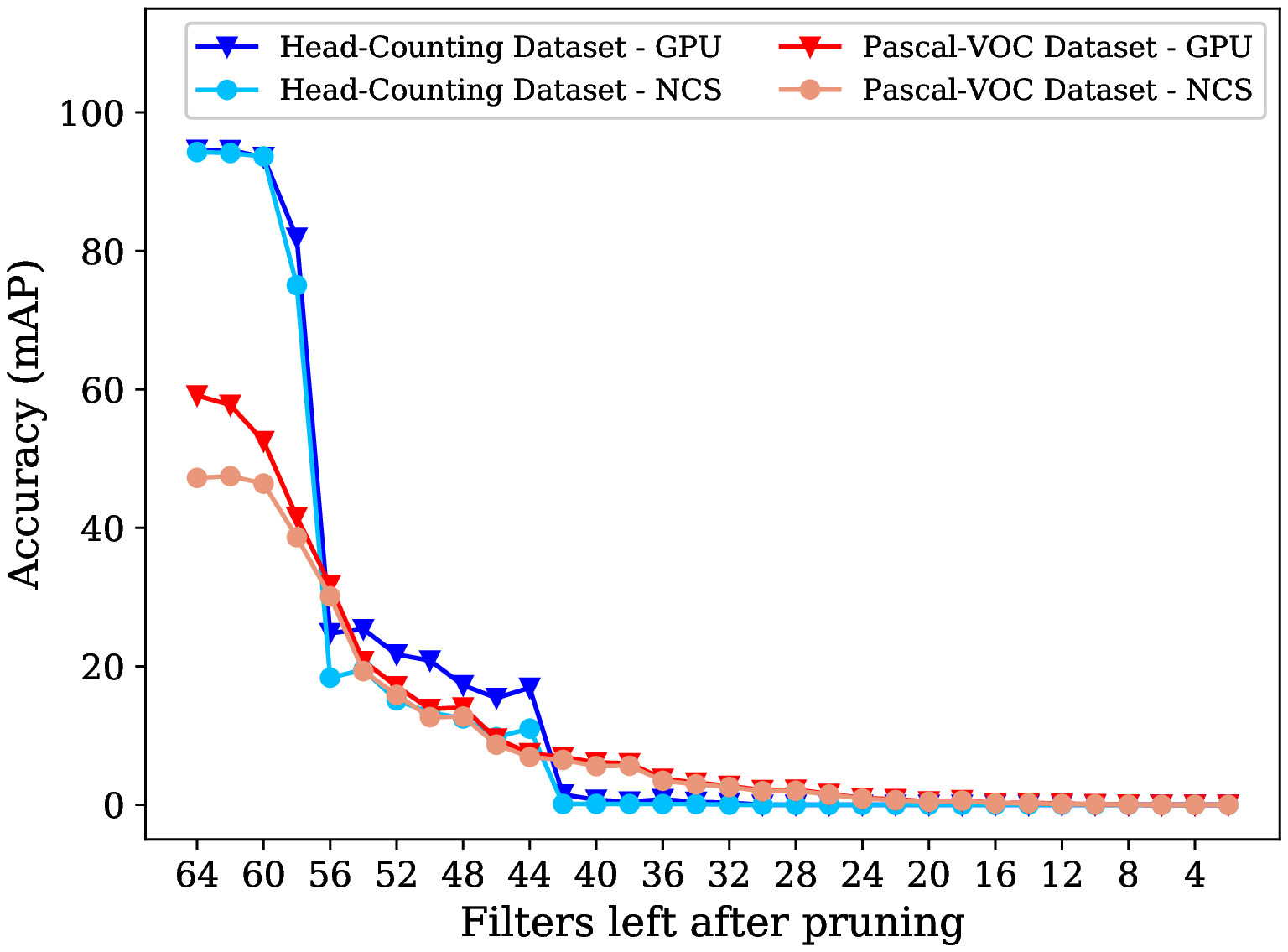}%
		\label{fig:accuracy_sq_conv1}}
	\subfloat[][Accuracy after pruning layer: Fire2/Squeeze1x1]{\includegraphics[width=0.39\textwidth]{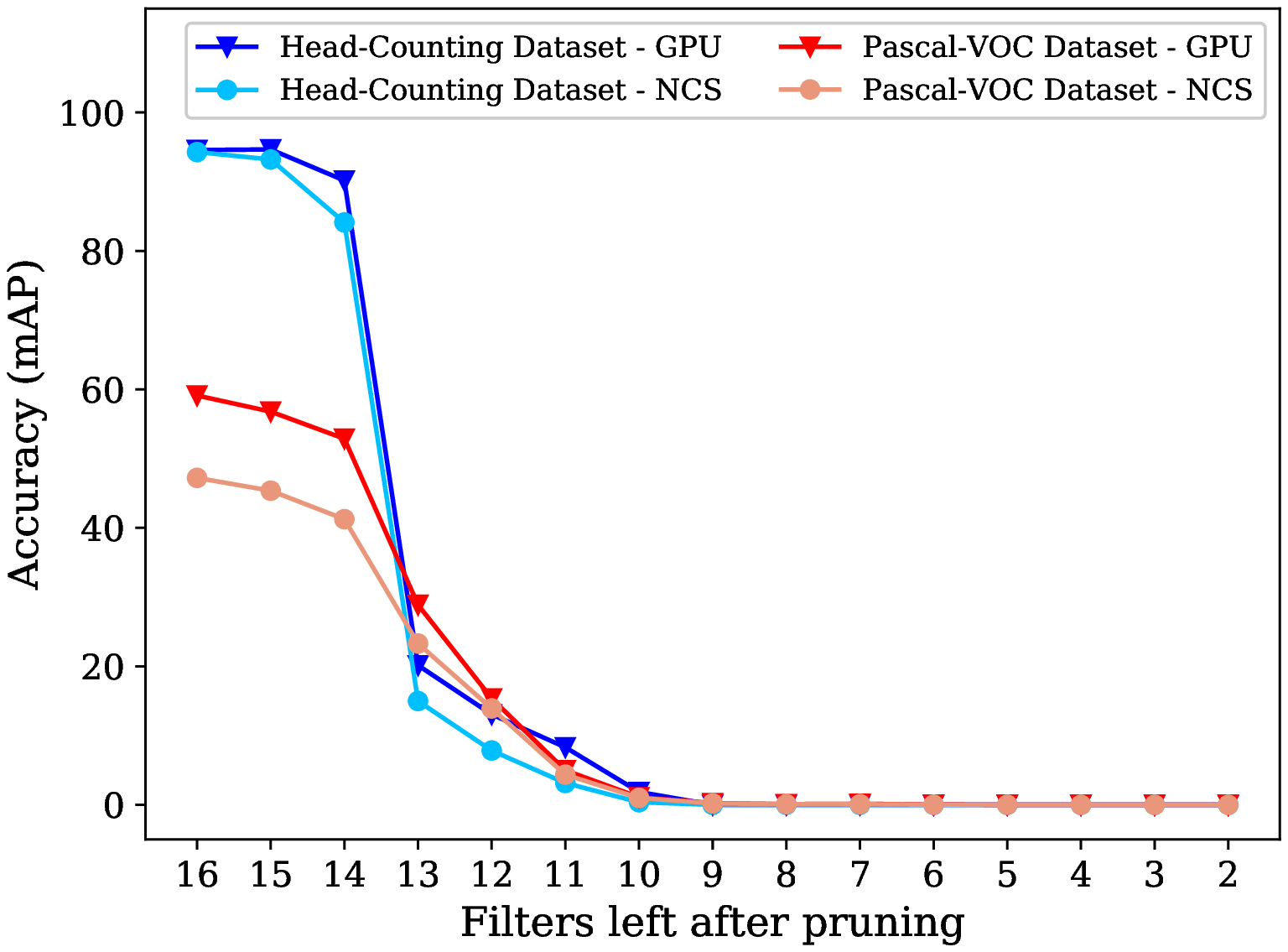}%
		\label{fig:accuracy_sq_fire2sq1}}\hfill
	\subfloat[][Accuracy after pruning layer: Fire2/Expand1x1]{\includegraphics[width=0.39\textwidth]{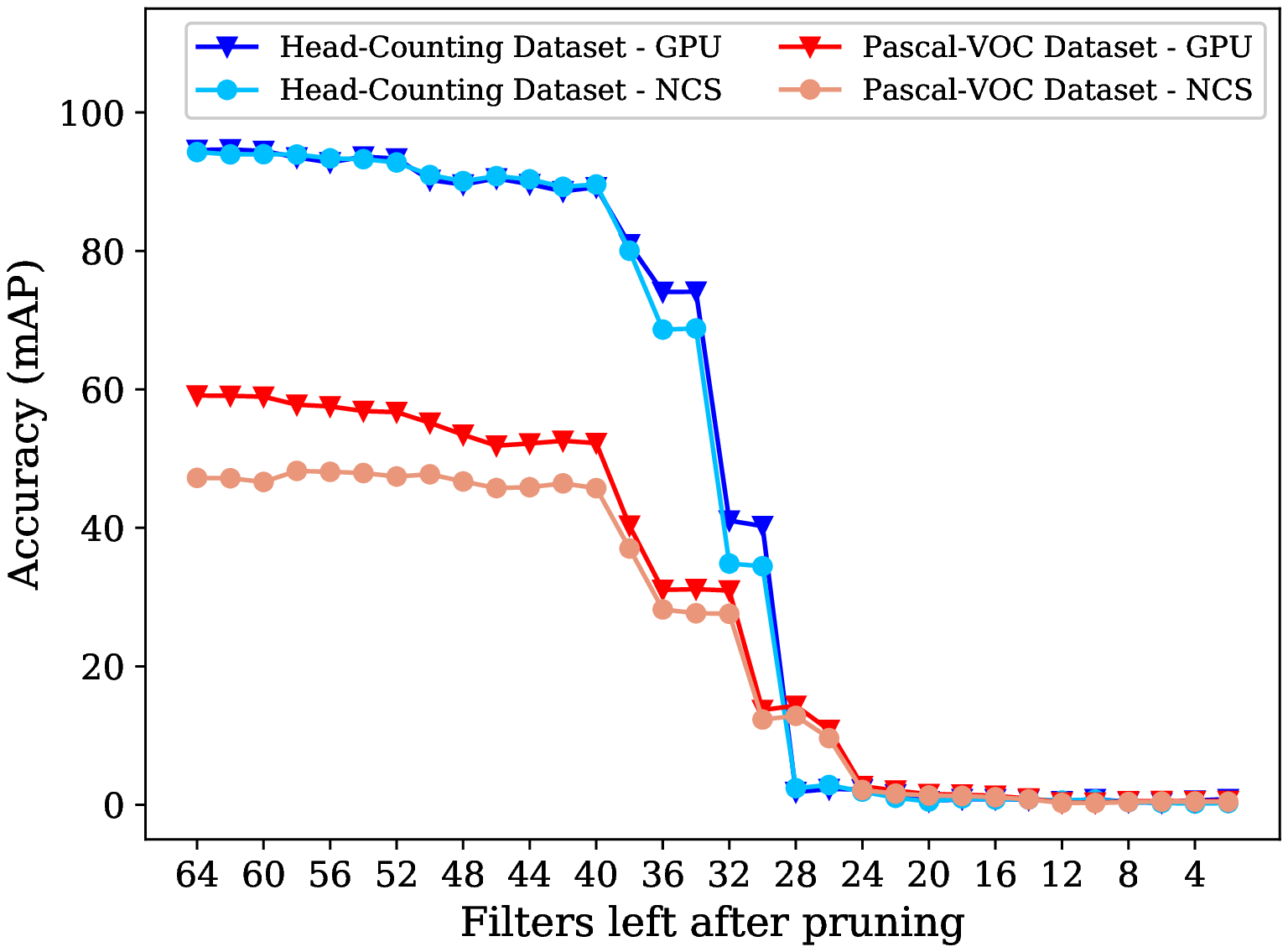}%
		\label{fig:accuracy_sq_fire2exp1}}
	\subfloat[][Accuracy after pruning layer: Fire2/Expand3x3]{\includegraphics[width=0.39\textwidth]{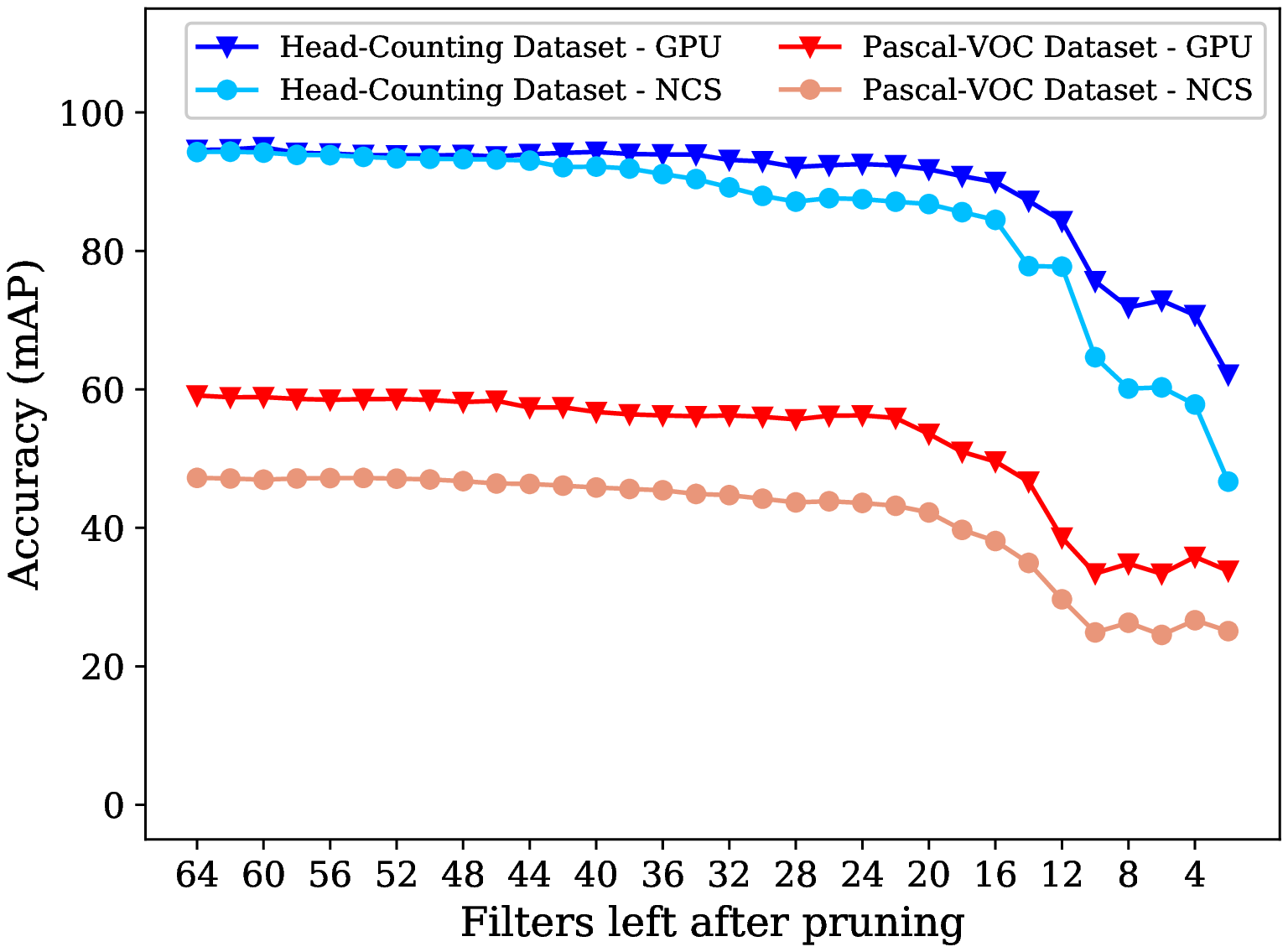}%
		\label{fig:accuracy_sq_fire2exp3}}
	\caption{Accuracy for Single layer pruning - SSD-SqueezeNet}
	\label{fig:accuracy_single_layer_sq}
\end{figure*}

SSD-MobileNet and SSD-SqueezeNet are trained on Pascal-VOC and Head-Counting datasets to measure the accuracy drop at each iteration of pruning. Fig. \ref{fig:accuracy_single_layer} and Fig. \ref{fig:accuracy_single_layer_sq} show the accuracy over the two datasets after pruning filters without any fine-tuning step. The pruned layers \textit{Conv0}, \textit{Conv1}, \textit{Conv2}, \textit{Conv6} of the SSD-MobileNet and \textit{Conv1}, \textit{Fire2/Squeeze1x1}, \textit{Fire2/Expand1x1}, \textit{Fire2/Expand3x3} of the SSD-SqueezeNet are illustrated by Fig. \ref{fig:accuracy_conv0}, \ref{fig:accuracy_conv1}, \ref{fig:accuracy_conv2}, \ref{fig:accuracy_conv6} and Fig. \ref{fig:accuracy_sq_conv1}, \ref{fig:accuracy_sq_fire2sq1}, \ref{fig:accuracy_sq_fire2exp1}, \ref{fig:accuracy_sq_fire2exp3}, respectively. To test the inference accuracy of the test datasets using GPU and CPU, we use the same network model based on Caffe framework at each pruning iteration. Therefore, same accuracy values are observed for both CPU and GPU evaluations. On the other hand, we get different accuracy results when we use Movidius-NCS, since we convert the Caffe based network model to a Movidius-NCS compatible network model called a graph file using the Movidius compiler. Thus, there are two plots of accuracy drops for each dataset as shown in the Fig. \ref{fig:accuracy_single_layer} and Fig. \ref{fig:accuracy_single_layer_sq}. We can summaries the single layer pruning results according to the three hardware architectures as follows.

\begin{figure*}
	\centering
	\subfloat[][Latency through Conv1: Conv1 pruned ]
	{\includegraphics[width=0.30\textwidth]{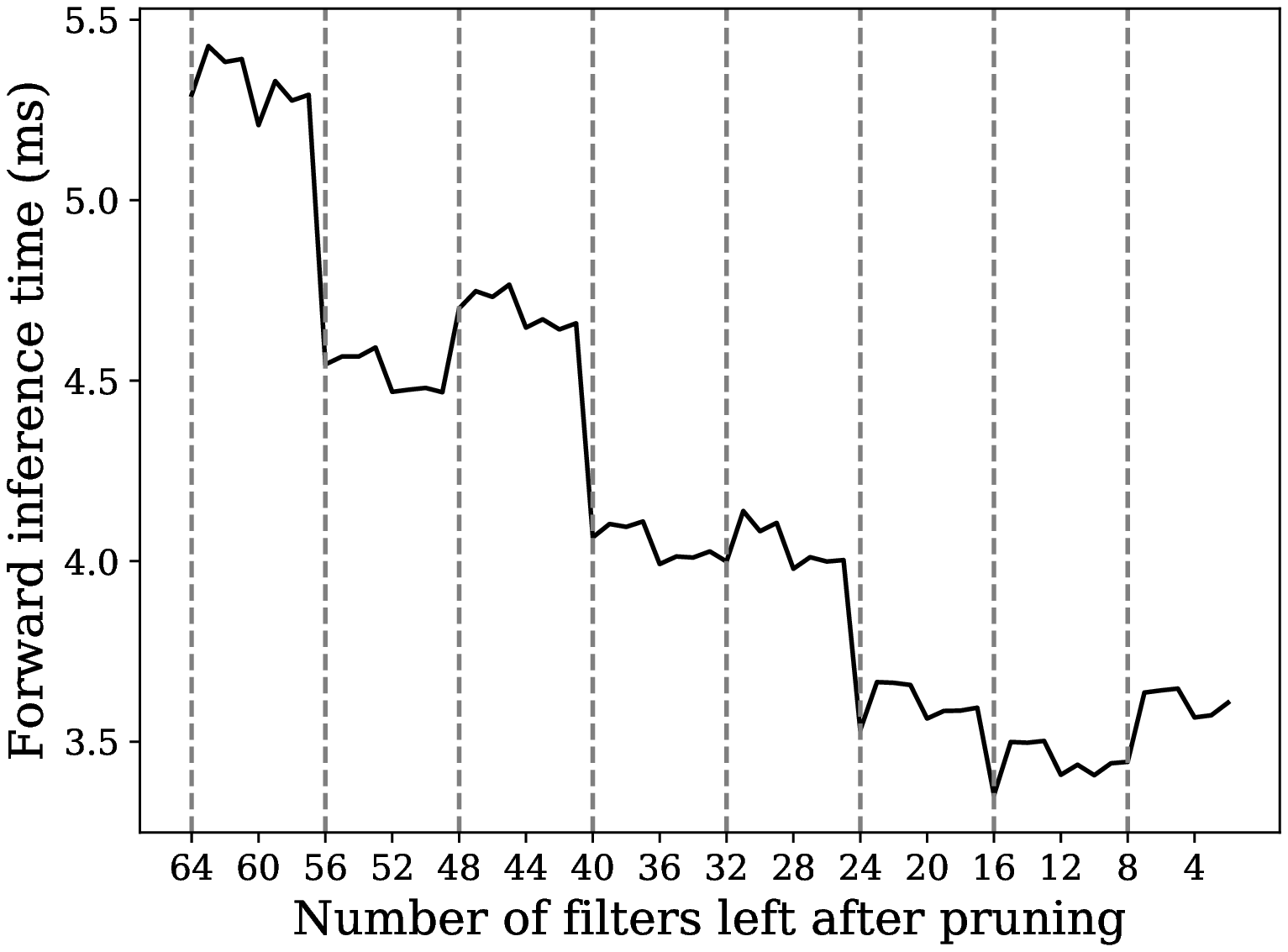}%
		\label{fig:latancy_sq_conv1_conv1}}\hfill
	\subfloat[][Latency through Fire2/Squeeze1x1: Conv1 pruned]
	{\includegraphics[width=0.30\textwidth]{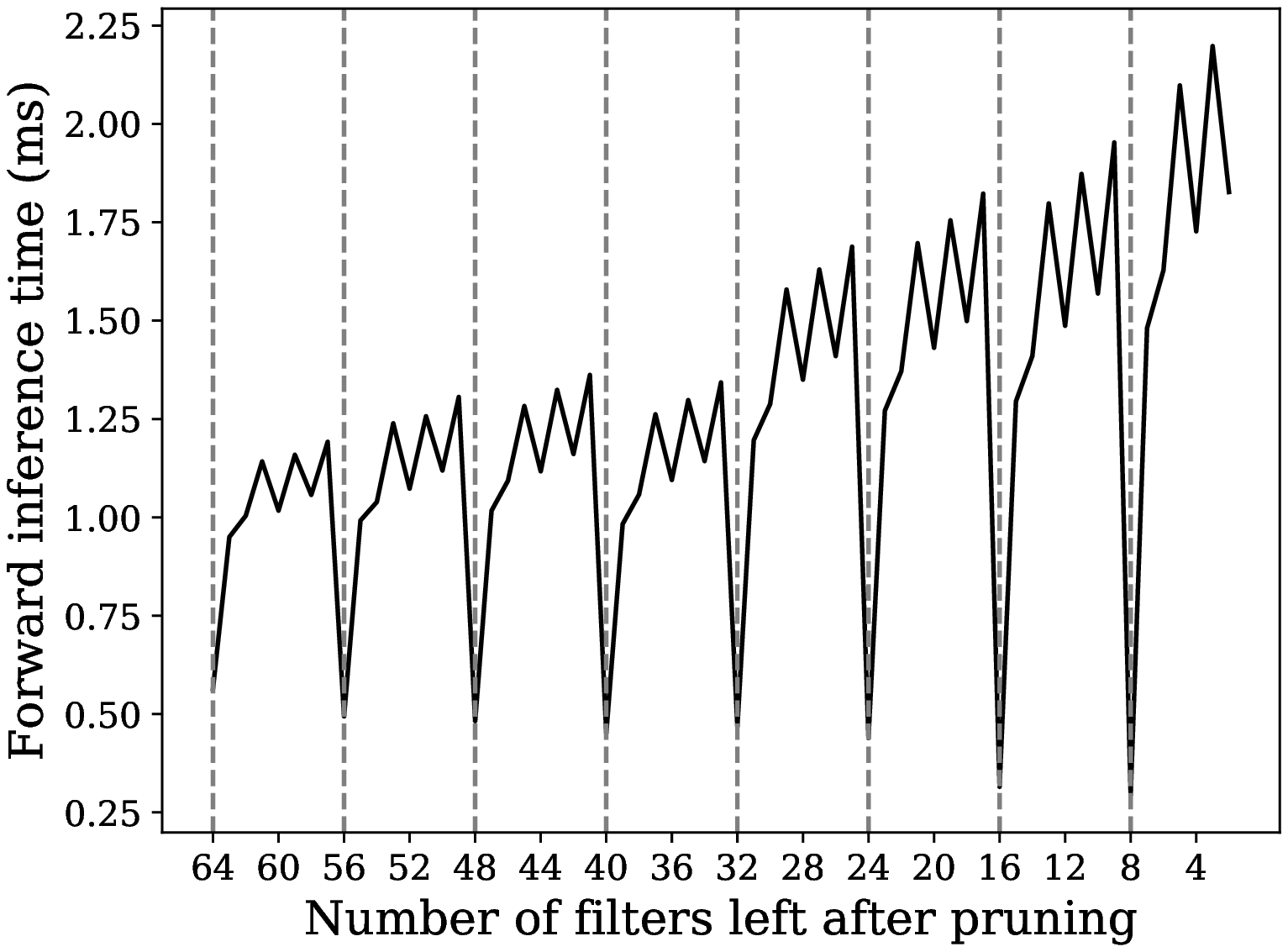}%
		\label{fig:latancy_sq_conv1_fire2sq1}}\hfill
	\subfloat[][Latency through Fire2/Expand1x1: Fire2/Expand1x1 pruned]
	{\includegraphics[width=0.30\textwidth]{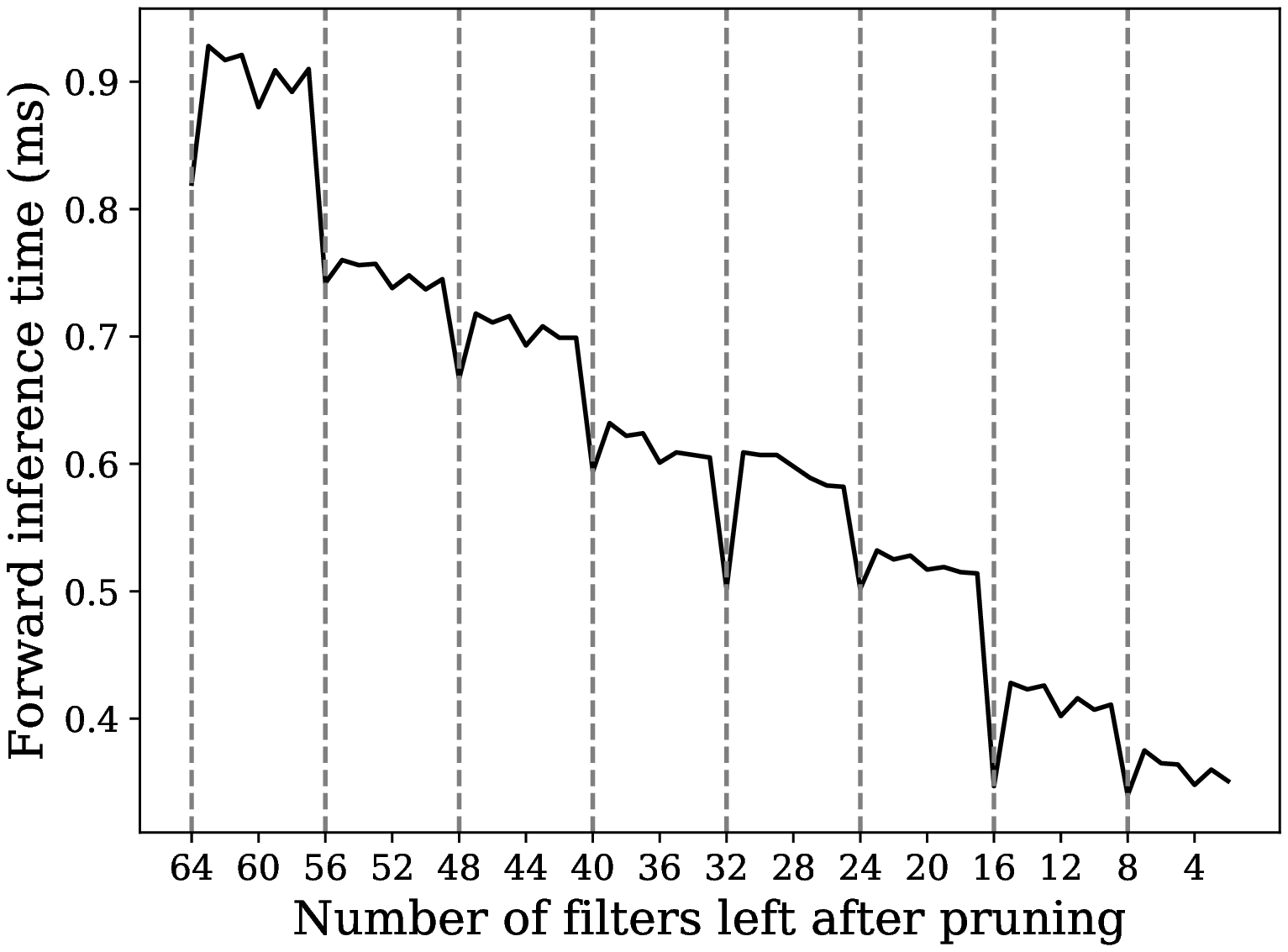}%
		\label{fig:latancy_sq_fire2exp1_fire2exp1}}\hfill
	\subfloat[][Latency through Fire3/Squeeze1x1: Fire2/Expand1x1 pruned]
	{\includegraphics[width=0.30\textwidth]{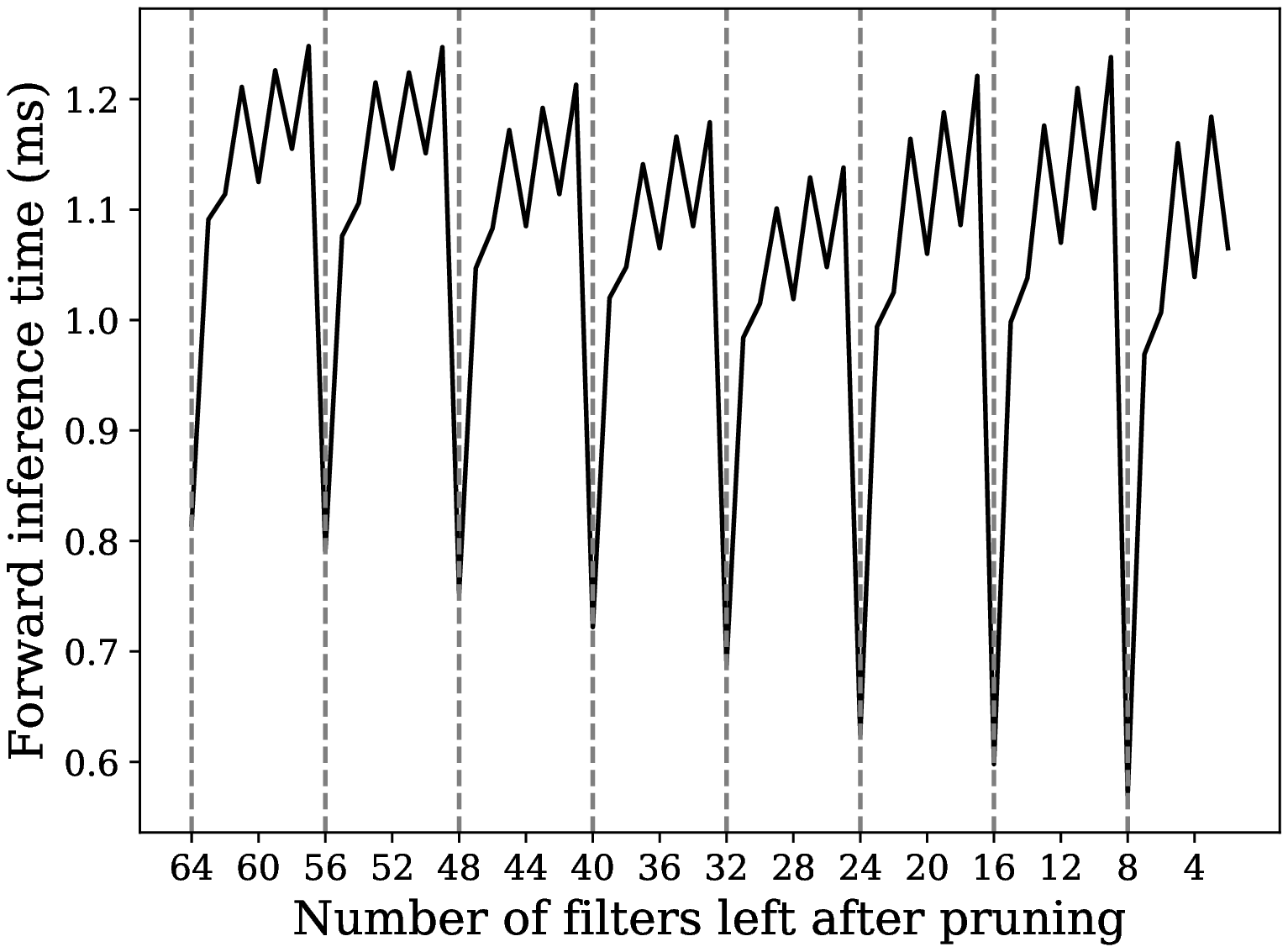}%
		\label{fig:latancy_sq_fire2exp1_fire3sq1}}\hfill
	\subfloat[][Latency through Fire2/Expand3x3: Fire2/Expand3x3 pruned]
	{\includegraphics[width=0.30\textwidth]{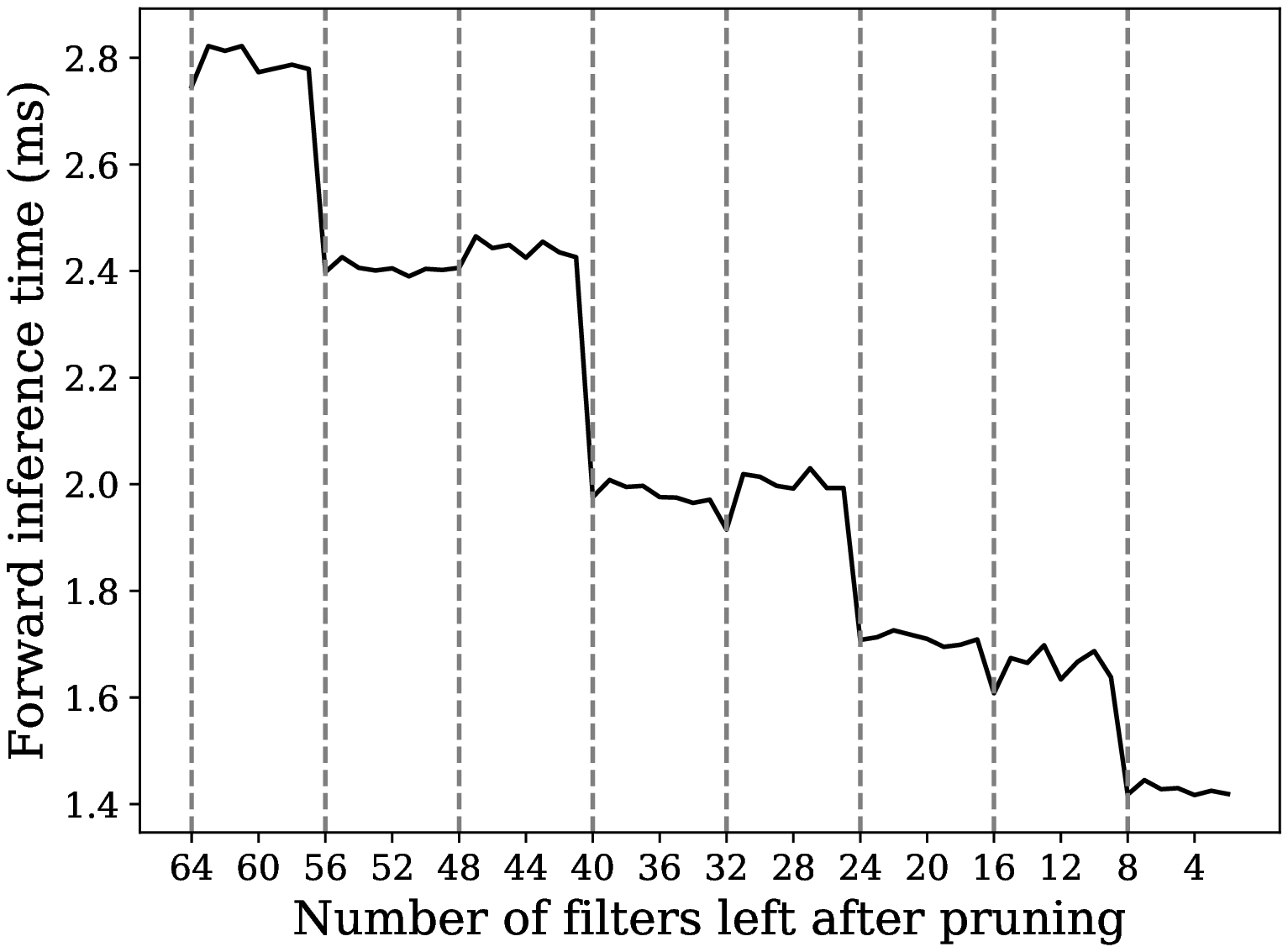}%
		\label{fig:latancy_sq_fire2exp3_fire2exp3}}\hfill
	\subfloat[][Latency through Fire3/Squeeze1x1: Fire2/Expand3x3 pruned]
	{\includegraphics[width=0.30\textwidth]{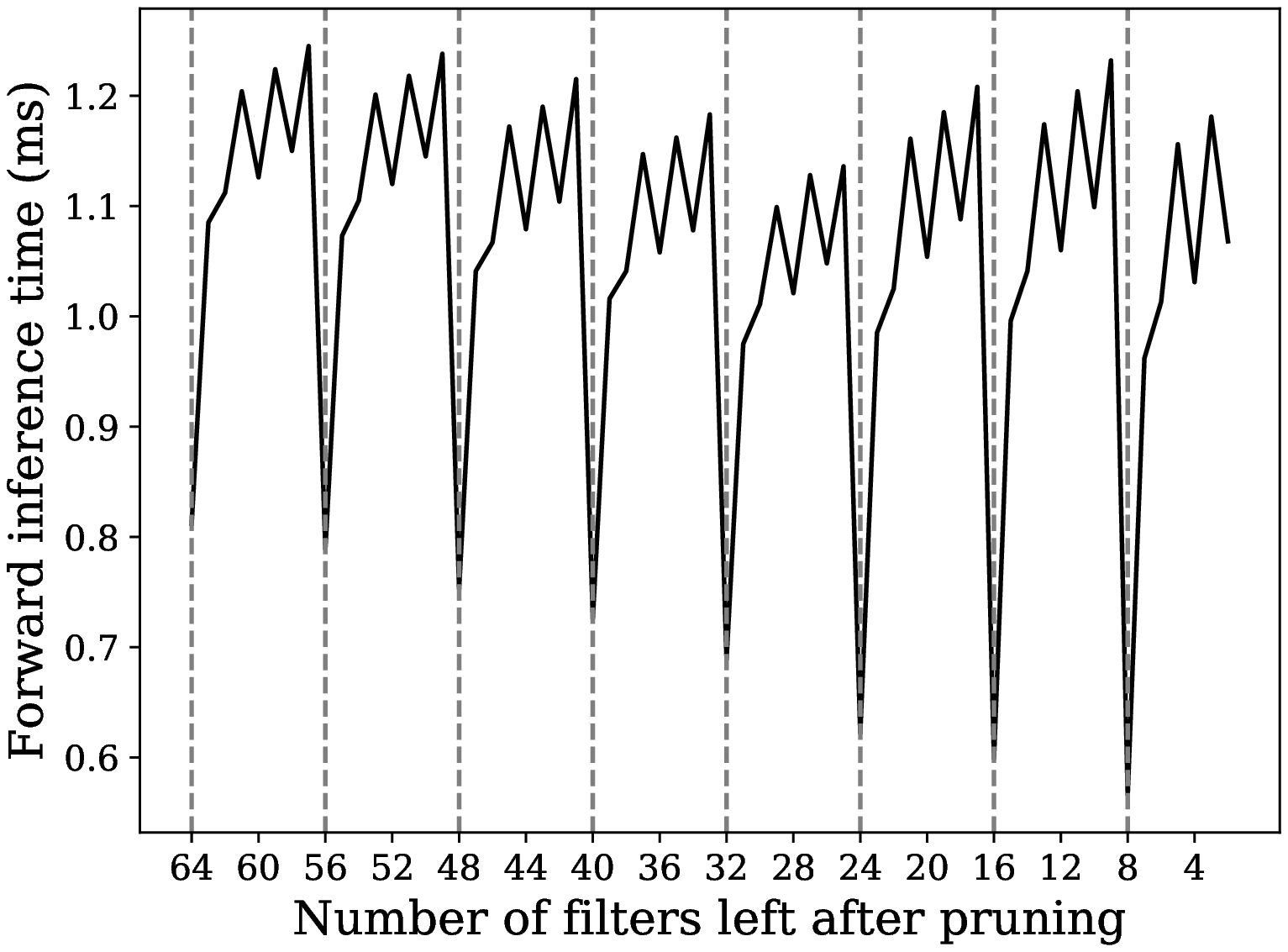}%
		\label{fig:latancy_sq_fire2exp3_fire2sq1}}\hfill
	\caption{Latency through individual layers (SSD-SqueezeNet): Single layer pruning using NCS}
	\label{fig:single_layer_ncs_sq}
\end{figure*}

\subsubsection{Movidius-NCS}

The Caffe implementation of both SSD-MobileNet and SSD-SqueezeNet are converted in to a binary file called a graph file capable of running in Movidius-NCS using the Movidius Neural Computing Software Development Kit (NCSDK) and Movidius compiler called \textit{mvNCCompiler}. When we are using the Movidius-NCS, forward inference time through each pruned layer in SSD-MobileNet and SSD-SqueezeNet are illustrated in the Fig. \ref{fig:single_layer_ncs} and Fig. \ref{fig:single_layer_ncs_sq}, respectively. At a given pruning iteration of layers \textit{Conv0}, \textit{Conv1}, and \textit{Conv2} in SSD-MobileNet, we evaluated the forward inference time through all three adjacent layers affected. Fig. \ref{fig:latancy_conv0_conv0}, \ref{fig:latancy_conv0_conv1dw}, \ref{fig:latancy_conv0_conv1} represent the pruning of layer \textit{Conv0}, while Fig. \ref{fig:latancy_conv1_conv1}, \ref{fig:latancy_conv1_conv2dw}, \ref{fig:latancy_conv1_conv2} and Fig. \ref{fig:latancy_conv2_conv2}, \ref{fig:latancy_conv2_conv3dw}, \ref{fig:latancy_conv2_conv3} represent pruning of the layers \textit{Conv1}, and \textit{Conv2}, respectively. Every graph in this figure shows periodic bottoms when remaining numbers of filters are equal to multiples of 8. The next pointwise convolution layer, which is effected by pruning of the filters in current pointwise convolution layer, has the most significant periodic bottoms as illustrated in Fig. \ref{fig:latancy_conv0_conv1}, \ref{fig:latancy_conv1_conv2}, and \ref{fig:latancy_conv2_conv3}. If the number of filters pruned are not in multiples of 8, forward inference time measured through that layer is increased. As a result of that, total network forward inference time shown in Fig.\ref{fig:latancy_conv0_ncs}, \ref{fig:latancy_conv1_ncs}, \ref{fig:latancy_conv2_ncs} follow the above mentioned periodic pattern of bottoms. Not only the SSD-MobileNet, but also the pruning of SSD-SqueezeNet shows the similar behaviour. Fig. \ref{fig:latancy_sq_conv1_conv1}, \ref{fig:latancy_sq_conv1_fire2sq1} represent the inference time through individual layers when pruning the layer \textit{Conv1} in SSD-SqueezeNet, while Fig. \ref{fig:latancy_sq_fire2exp1_fire2exp1}, \ref{fig:latancy_sq_fire2exp1_fire3sq1} and Fig. \ref{fig:latancy_sq_fire2exp3_fire2exp3}, \ref{fig:latancy_sq_fire2exp3_fire2sq1} represent pruning of the layers \textit{Fire2/Expand1x1}, and \textit{Fire2/Expand3x3}, respectively. According to these figures, forward inference time through the following layer pruned is greatly increased if the number of pruned filters are not in multiples of 8. Thus, the total inference time shown in Fig.\ref{fig:latancy_sqconv1_ncs}, \ref{fig:latancy_sqfire2exp1_ncs}, \ref{fig:latancy_sqfire2exp3_ncs} follows the periodic pattern of 8. This scenario is observed due to the specific data-flow design architecture that we observe in Movidius-NCS. Thus we can select the $p_l^{lat}$ value mentioned in Section III Subsection B as 8 for both of the networks which is used to calculate optimum cluster sizes per layer ($\mathcal{P}$) used in Algorithm \ref{cluster_pruning}.

Not only the latency graphs, but also the accuracy graphs of the SSD-MobileNet shown in Fig. \ref{fig:accuracy_single_layer} show periodic tops when the pruned number of filters are equal to multiples of 8. As it is shown in Fig. \ref{fig:accuracy_single_layer}, the Movidius compiler preserves the accuracy if the number of filters are pruned in multiples of 8 for both datasets. Thus we can select $p_l^{acc}$ value mentioned in Section III Subsection B as 8 for the SSD-Mobilenet. The accuracy graphs of the SSD-SqueezeNet shown in Fig. \ref{fig:accuracy_single_layer_sq} do not show any specific pattern except the degradation of accuracy than the CPU and GPU accuracy plot. That indicates the accuracy and optimum cluster size are independent of each other for the SSD-SqueezeNet when we use the Movidius-NCS. Therefore we can select $p_l^{acc}$ value to be 1 for the SSD-SqueezeNet. Consequently, we can come to the conclusion empirically that the optimum cluster size ($LCM(p_l^{acc},p_l^{lat})$) for each layer is 8 for both of the detection networks when we use Movidius-NCS. In the next subsection, we are going to use this optimum cluster size for the cluster pruning method we proposed.

\subsubsection{CPU}

For the experiment we use an Intel-Xeon-CPU with Caffe-CPU run-time framework to measure the network forward inference time and test accuracy. Fig. \ref{fig:latancy_conv0_cpu}, \ref{fig:latancy_conv1_cpu}, \ref{fig:latancy_conv2_cpu} show the forward inference time after pruning the layers \textit{Conv0}, \textit{Conv1}, \textit{Conv2} of SSD-MobileNet, while Fig. \ref{fig:latancy_sqconv1_cpu}, \ref{fig:latancy_sqfire2exp1_cpu}, \ref{fig:latancy_sqfire2exp3_cpu} show the forward inference time after pruning the layers \textit{Conv1}, \textit{Fire2/Expand1x1}, \textit{Fire2/Expand3x3} of SSD-SqueezeNet, respectively. There is a general trend of decreasing total inference time with random fluctuations when number of pruned filter are increasing. But there is no periodic pattern that we observe in the test using a CPU for both networks. Accuracy results of CPU test is identical to the GPU variant. When the remaining number of filters inside the layers \textit{Conv0}, \textit{Conv1}, \textit{Conv2}, \textit{Conv6} of SSD-MobileNet and \textit{Conv1}, \textit{Fire2/Squeeze1x1}, \textit{Fire2/Expand1x1}, \textit{Fire2/Expand3x3} of SSD-SqueezeNet are decreasing, accuracy for detection of objects in both test datasets are dropping. Sensitivity for the accuracy of the bottom layers are less than the top most layers of the networks, where Fig \ref{fig:accuracy_conv0}, and Fig. \ref{fig:accuracy_conv6} show the highest and least sensitivity of the SSD-MobileNet. We do not observe a remarkable patterns of accuracy associated with the number of filters removed in CPU and GPU experiments as illustrated in Fig. \ref{fig:accuracy_single_layer} and Fig. \ref{fig:accuracy_single_layer_sq}. Furthermore, we can use the CPU and GPU accuracy plot as the baseline while comparing the accuracy drops in Movidius-NCS experiment.  

\subsubsection{GPU}

Caffe-GPU runtime framework is used with an Intel-Xeon-CPU and an Nvidia-GeForce-GTX-1080Ti for profiling the performance of single layer pruning in our experiment. Fig. \ref{fig:latancy_conv0_gpu}, \ref{fig:latancy_conv1_gpu}, \ref{fig:latancy_conv2_gpu} show the forward inference time after pruning the layers \textit{Conv0}, \textit{Conv1} \textit{Conv2} of SSD-MobileNet, while Fig. \ref{fig:latancy_sqconv1_gpu}, \ref{fig:latancy_sqfire2exp1_gpu}, \ref{fig:latancy_sqfire2exp3_gpu} show the forward inference time after pruning the layers \textit{Conv1}, \textit{Fire2/Expand1x1}, \textit{Fire2/Expand3x3} of SSD-SqueezeNet, respectively. In these figures, we observe only random fluctuations bounded between a 1ms-2ms time difference. This is due to the massively parallel hardware capability of the GPU. Even though, the number of computations and network size is reduced, we do not experience a significant reduction of latency in GPU latency results. The accuracy results are same as the CPU variant of the experiment as mentioned above.

According to the single layer pruning results, we can identify that Movidius-NCS is susceptible to workload imbalance as shown by the periodic bottoms in latency graphs and periodic tops in accuracy graphs. Accordingly, optimum cluster size is identified as 8 for both networks, which is used in the whole model pruning step. Even though, the CPU and GPU experiments do not show the periodic pattern in performance graphs, performance values are evaluated for cluster pruning methodology using the CPU and GPU in the next subsection to differentiate the results with Movidius-NCS.

\subsection{Whole Model Pruning}

After observing the results of single layer pruning and identifying the optimum cluster size, we prune the whole network model irrespective of a selected layer using the filter pruning methodology and cluster pruning methodology. To make the implementation easier, we select the layers from \textit{Conv1} to \textit{Conv9} in SSD-MobileNet and fire modules from \textit{Fire2} to \textit{Fire8} in SSD-SqueezeNet to be pruned. The filter pruning methodology is implemented by ranking all the filters inside the layers according to the importance using the minimum weight criteria. Then, we prune filters unevenly across layers, where least important filters are pruned first. To implement the cluster pruning methodology, we use the selected layers in SSD-MobileNet and SSD-SqueezeNet to follow the Algorithm \ref{cluster_pruning} using the cluster size as 8. In both methods, once we have pruned 8 filters from the network, we measure the total network inference time and the accuracy for both datasets without an intermediate fine-tuning step initially. Furthermore, we fine-tune the models pre-trained on Pascal-VOC using 2000 updates with a learning rate, which is half the base learning rate after pruning every 8 filters. For the models pre-trained on Head-Counting dataset, we use 1000 updates with a learning rate, which is half the base learning rate. Then again we measure the accuracy in both methodologies.

\begin{figure*}
	\centering%
	\subfloat[][NCS forward inference time after prunning]{%
		\includegraphics[width=0.32\textwidth]{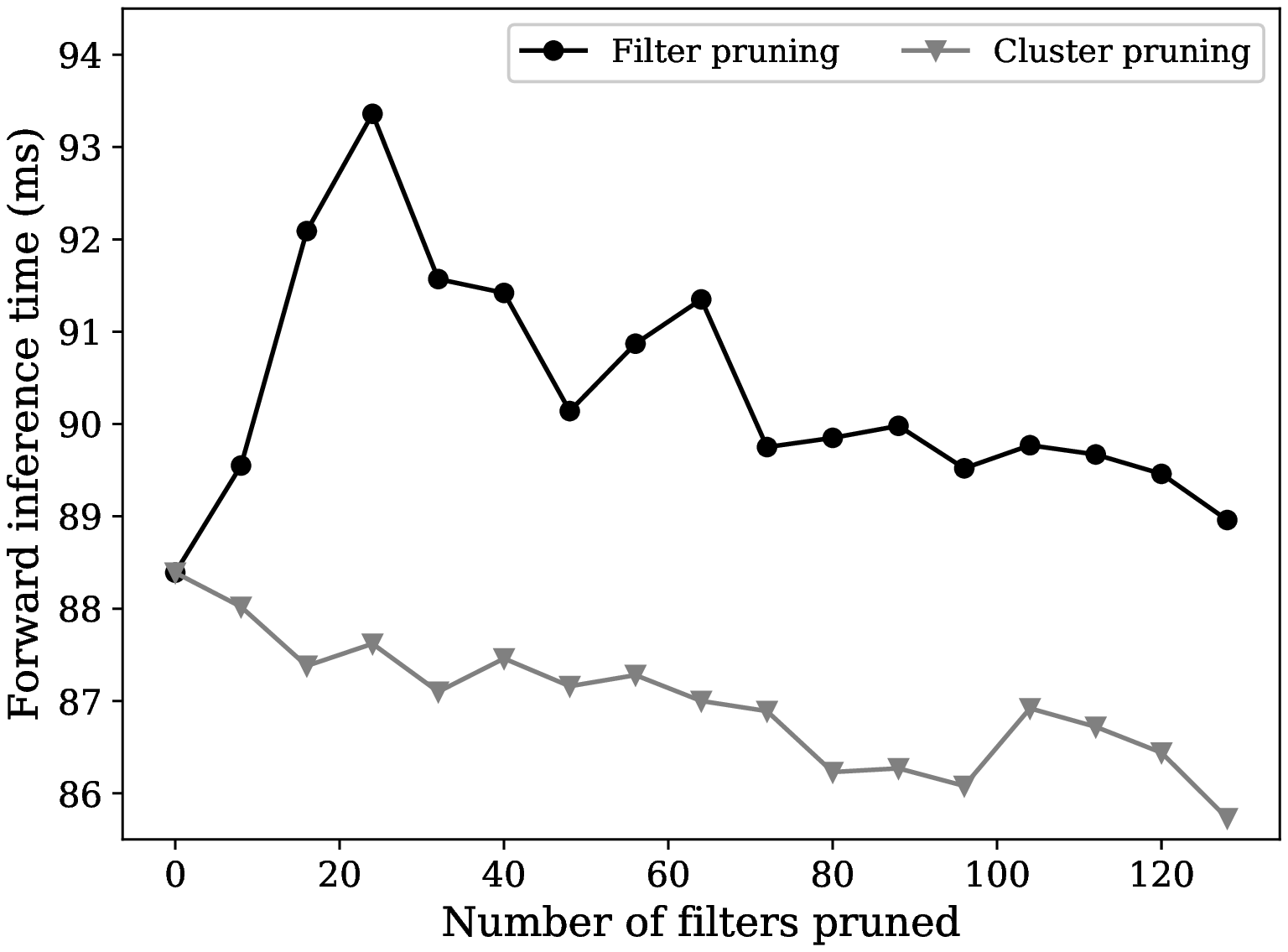}
		\label{fig:ncs_time}%
	}%
	\subfloat[][CPU forward inference time after prunning]{%
		\includegraphics[width=0.32\textwidth]{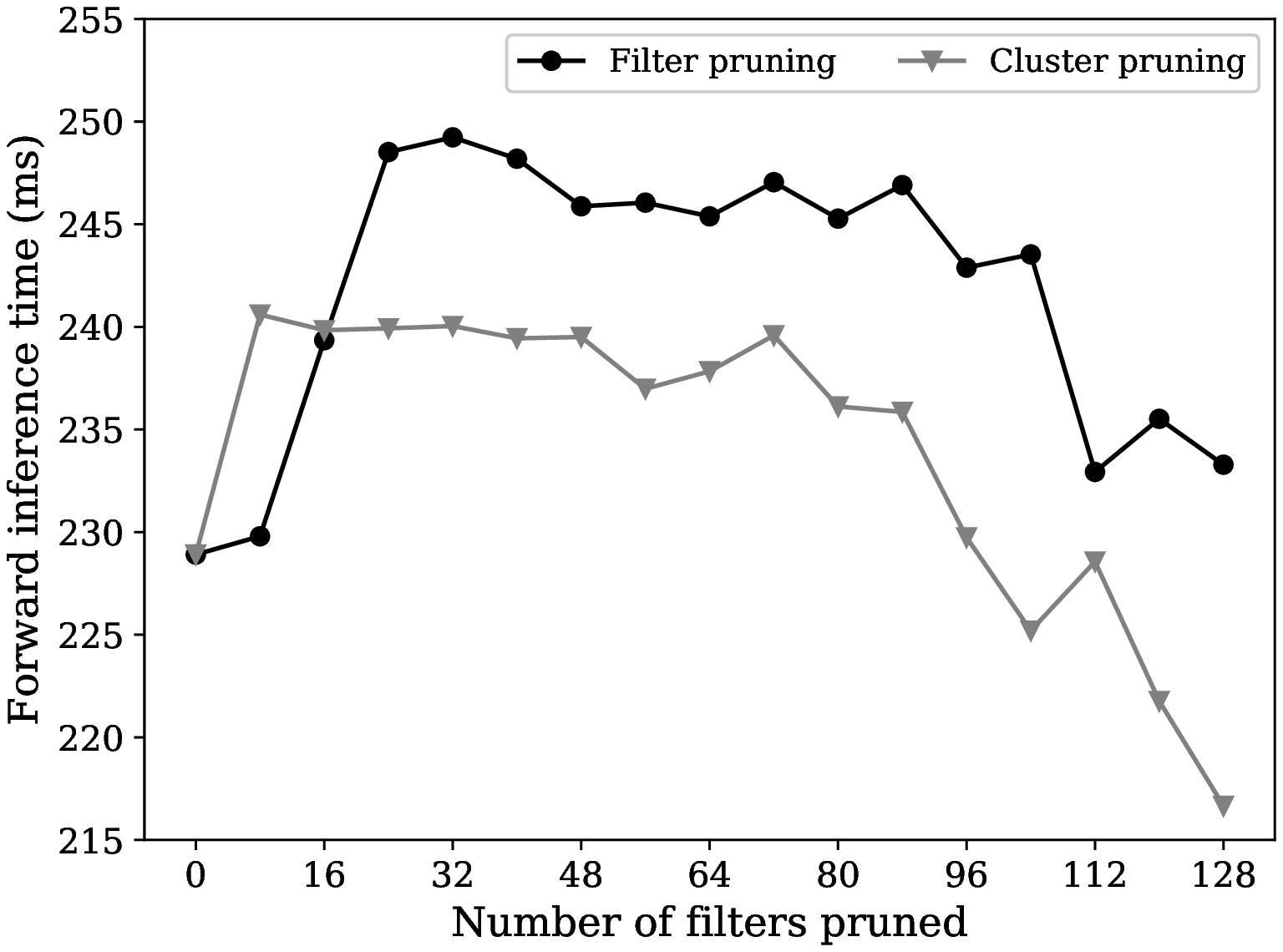}
		\label{fig:cpu_time}%
	}%
	\subfloat[][GPU forward inference time after prunning]{%
		\includegraphics[width=0.32\textwidth]{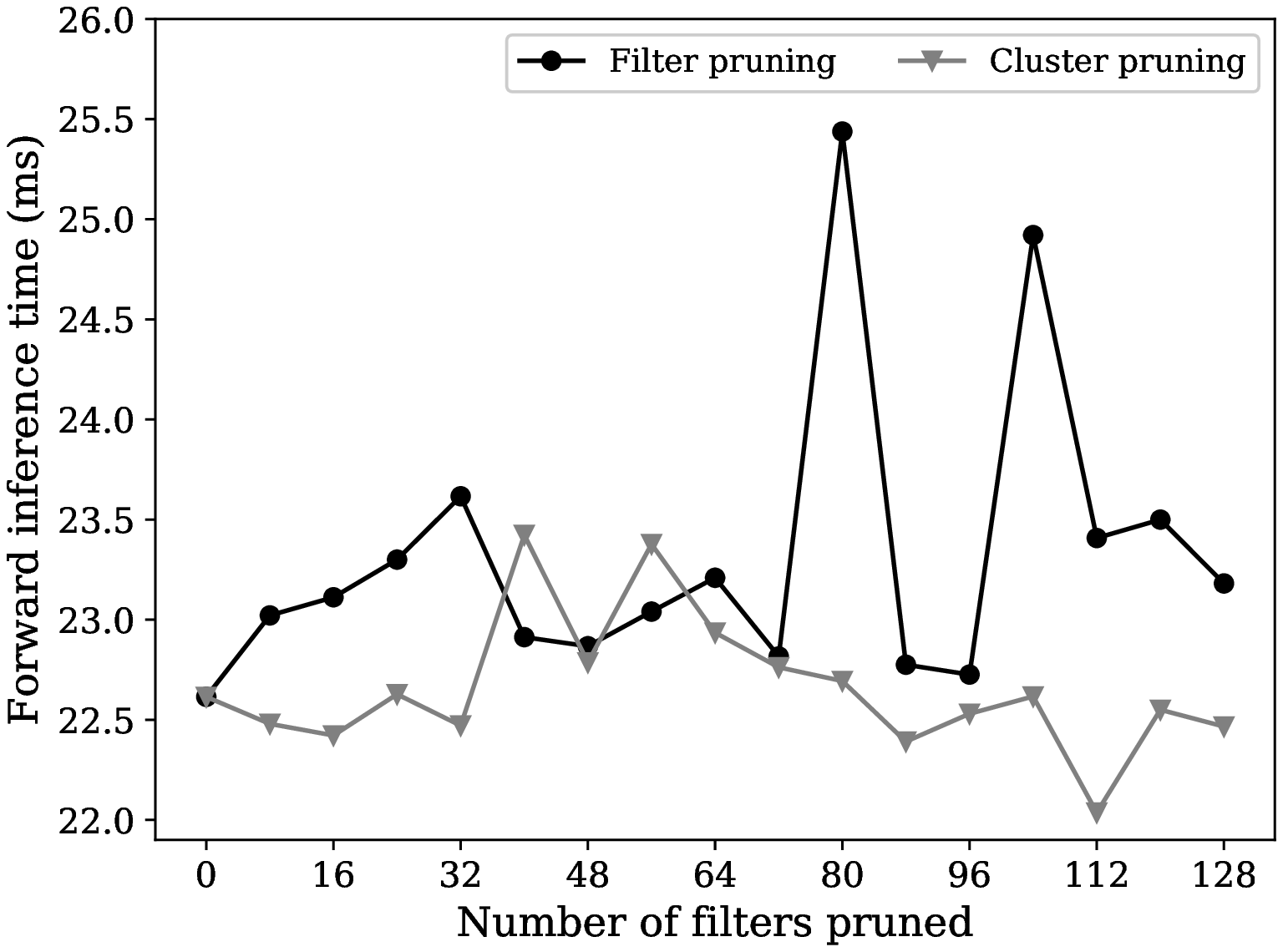}
		\label{fig:gpu_time}%
	}%
	\caption{Inference latency : Filter pruning vs Cluster pruning (SSD-MobileNet)}%
	\label{fig:model_comparison}%
\end{figure*}

\begin{figure*}
	\centering%
	\subfloat[][NCS forward inference time after prunning]{%
		\includegraphics[width=0.32\textwidth]{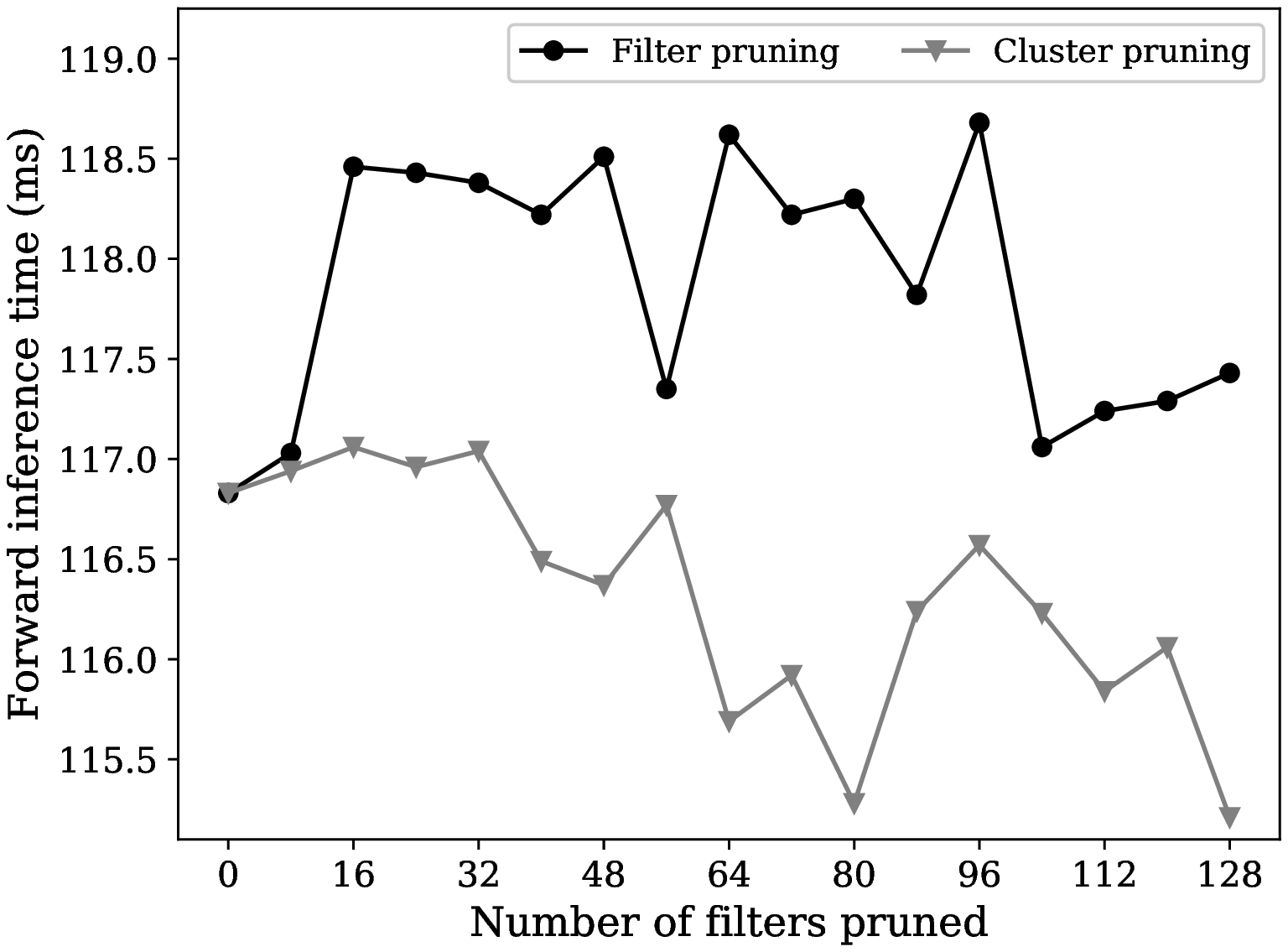}
		\label{fig:ncs_time_sq}%
	}%
	\subfloat[][CPU forward inference time after prunning]{%
		\includegraphics[width=0.32\textwidth]{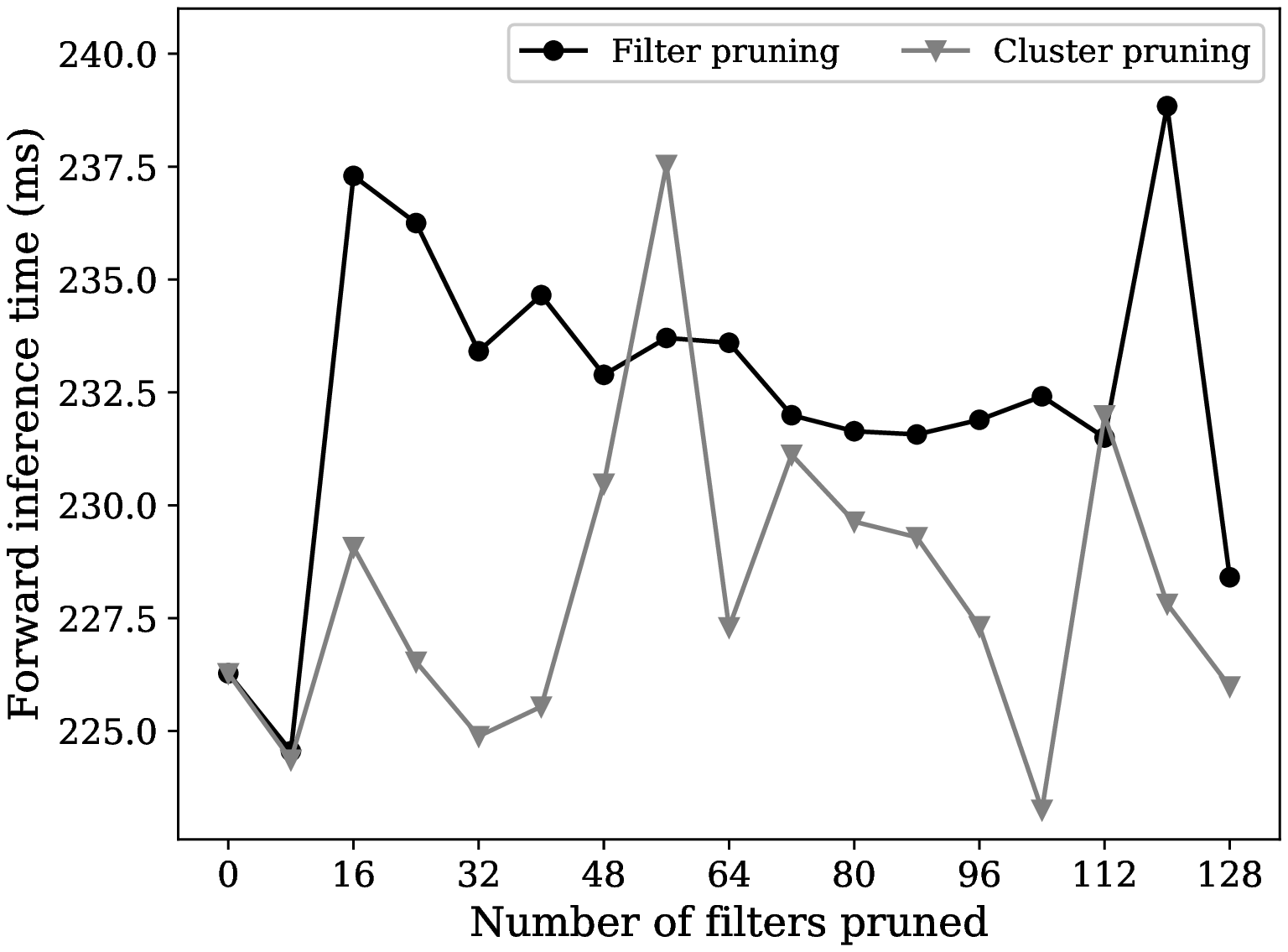}
		\label{fig:cpu_time_sq}%
	}%
	\subfloat[][GPU forward inference time after prunning]{%
		\includegraphics[width=0.32\textwidth]{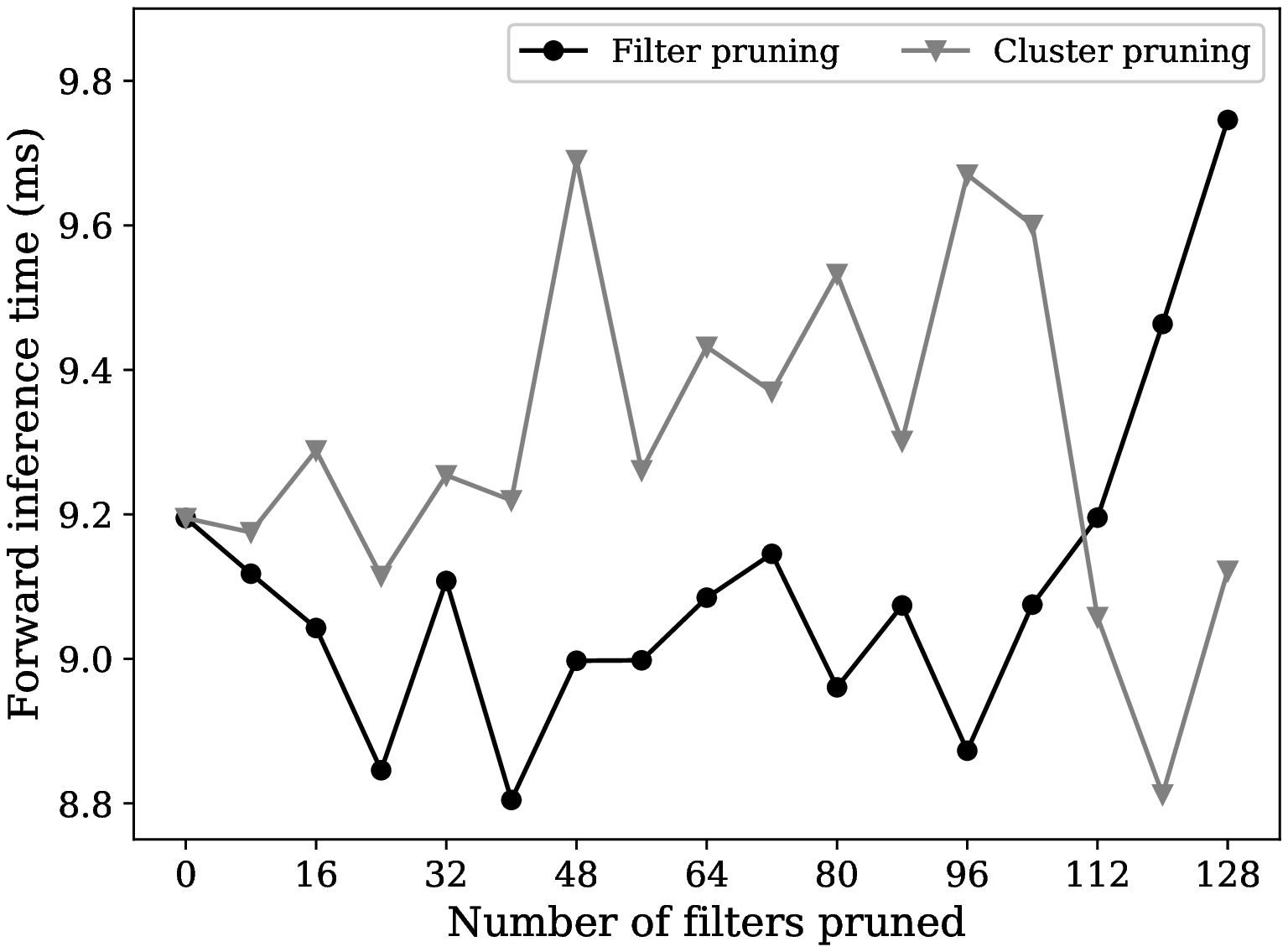}
		\label{fig:gpu_time_sq}%
	}%
	\caption{Inference latency : Filter pruning vs Cluster pruning (SSD-SqueezeNet)}%
	\label{fig:model_comparison_sqnet}%
\end{figure*}

The Fig. \ref{fig:model_comparison} and Fig. \ref{fig:model_comparison_sqnet} indicate the network forward inference time comparison between the filter pruning methodology and proposed cluster pruning methodology using the three hardware architectures NCS, CPU, and GPU for SSD-MobileNet and SSD-SqueezeNet, respectively. Average percentage of the latency drops for SSD-MobileNet from filter pruning method to cluster pruning method are 3.93\%, 3.38\%, and 2.92\% for NCS, CPU, and GPU, respectively. For SSD-SqueezeNet, the average percentage of the latency drops are 1.40\%, 1.93\%, and -2.40\%, respectively. Most of the time, cluster pruning method outperforms the filter pruning method in all three hardware architectures. As demonstrated in Fig. \ref{fig:ncs_time}  and Fig. \ref{fig:ncs_time_sq}, all the time Movidius-NCS has distinct latency drops since it supports the cluster pruning methodology as we identified in the single layer pruning experiment earlier.

\begin{figure*}
	\centering
	\subfloat[][NCS accuracy after pruning]{
		\includegraphics[width=0.40\textwidth]{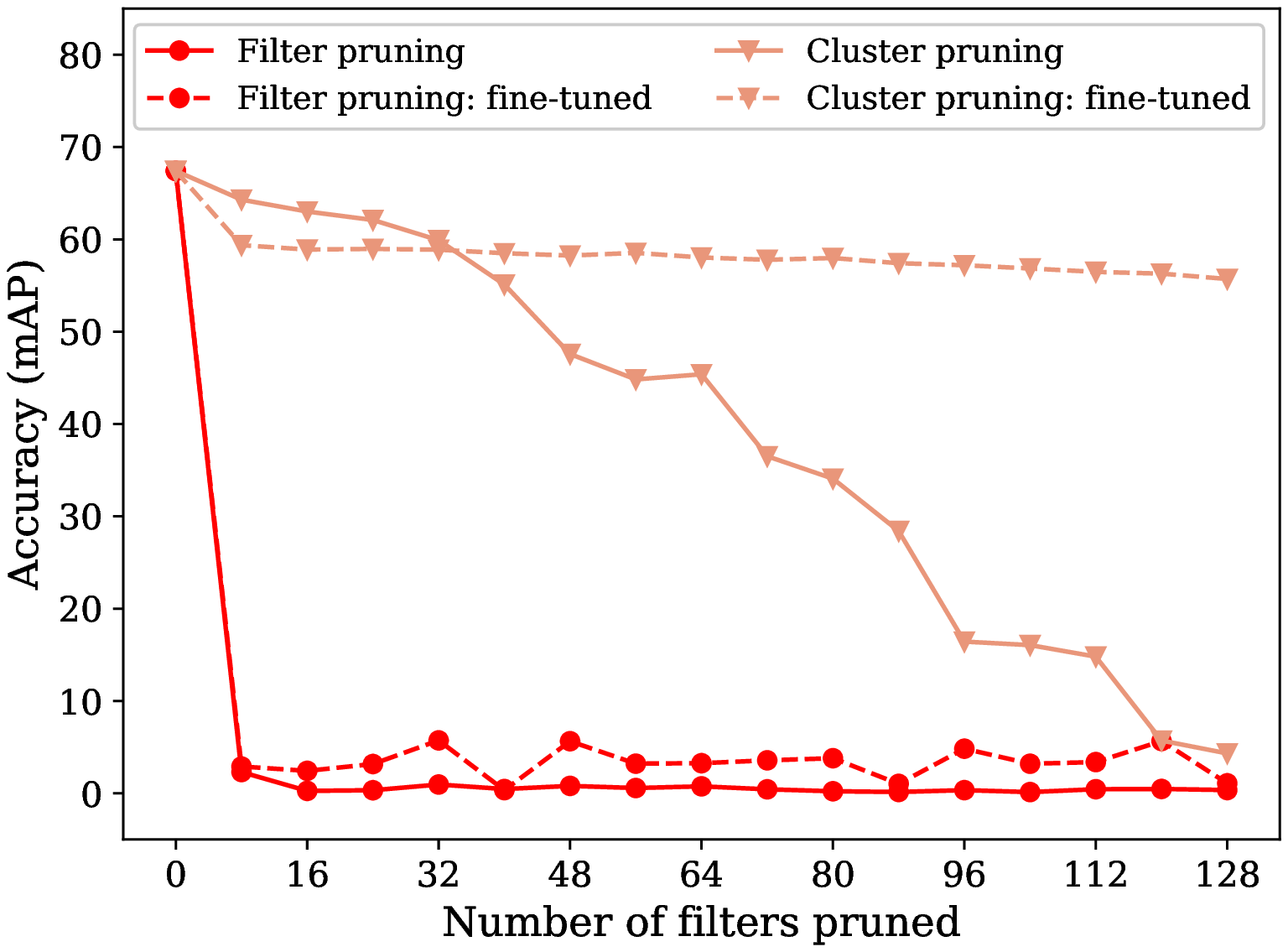}%
		\label{fig:NCS_accuracy_after_prunning}
	}%
	\subfloat[][CPU/GPU accuracy after pruning]{
		\includegraphics[width=0.40\textwidth]{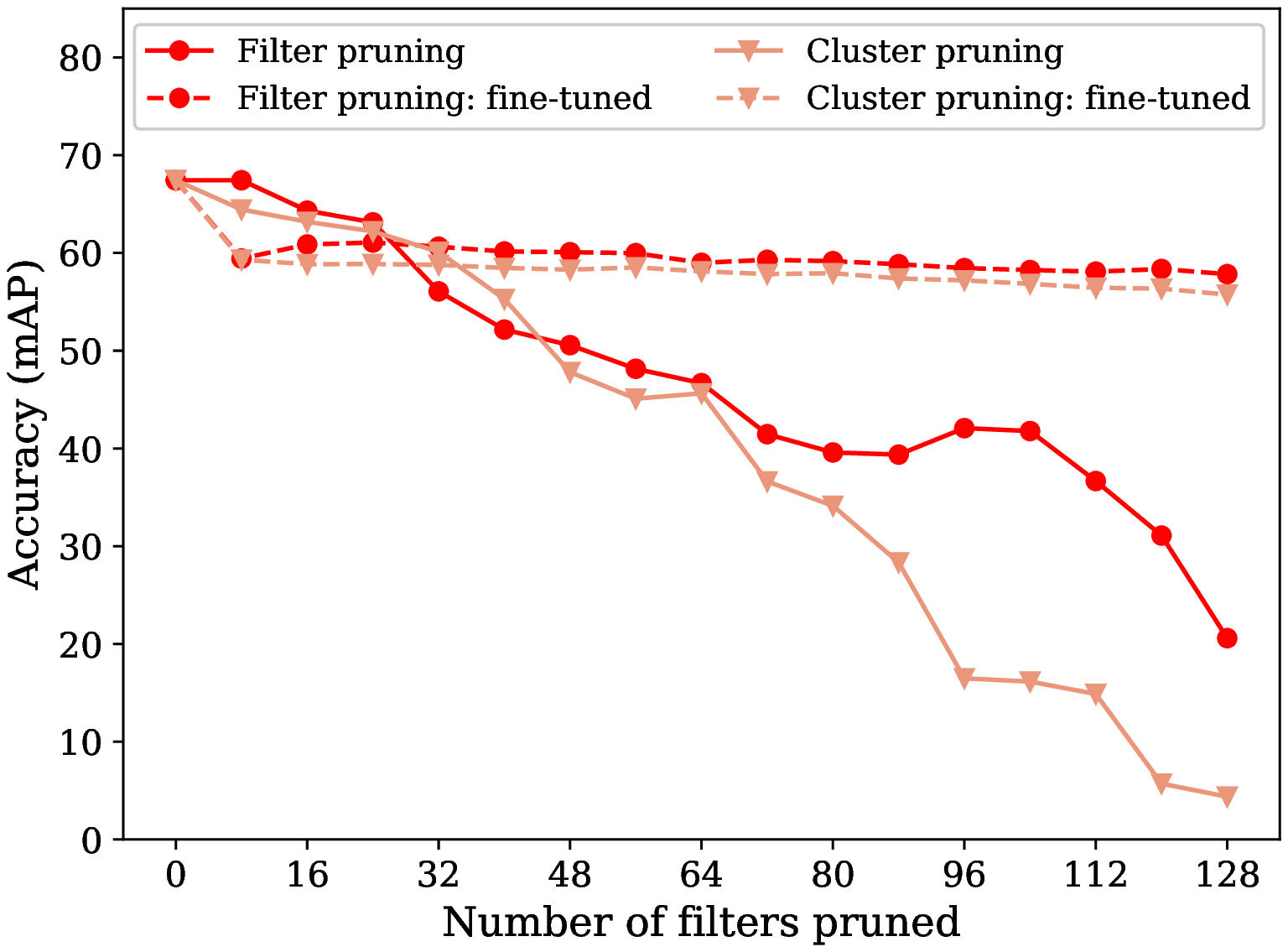}%
		\label{fig:Caffe_GPU_accuracy_after_prunning}\hfill
	}
	\caption{Accuracy for Pascal-VOC dataset : Filter pruning vs Cluster pruning (SSD-MobileNet)}
	\label{fig:Accuracy_Pascal_VOC}%
\end{figure*}

\begin{figure*}
	\centering
	\subfloat[][NCS accuracy after pruning]{
		\includegraphics[width=0.40\textwidth]{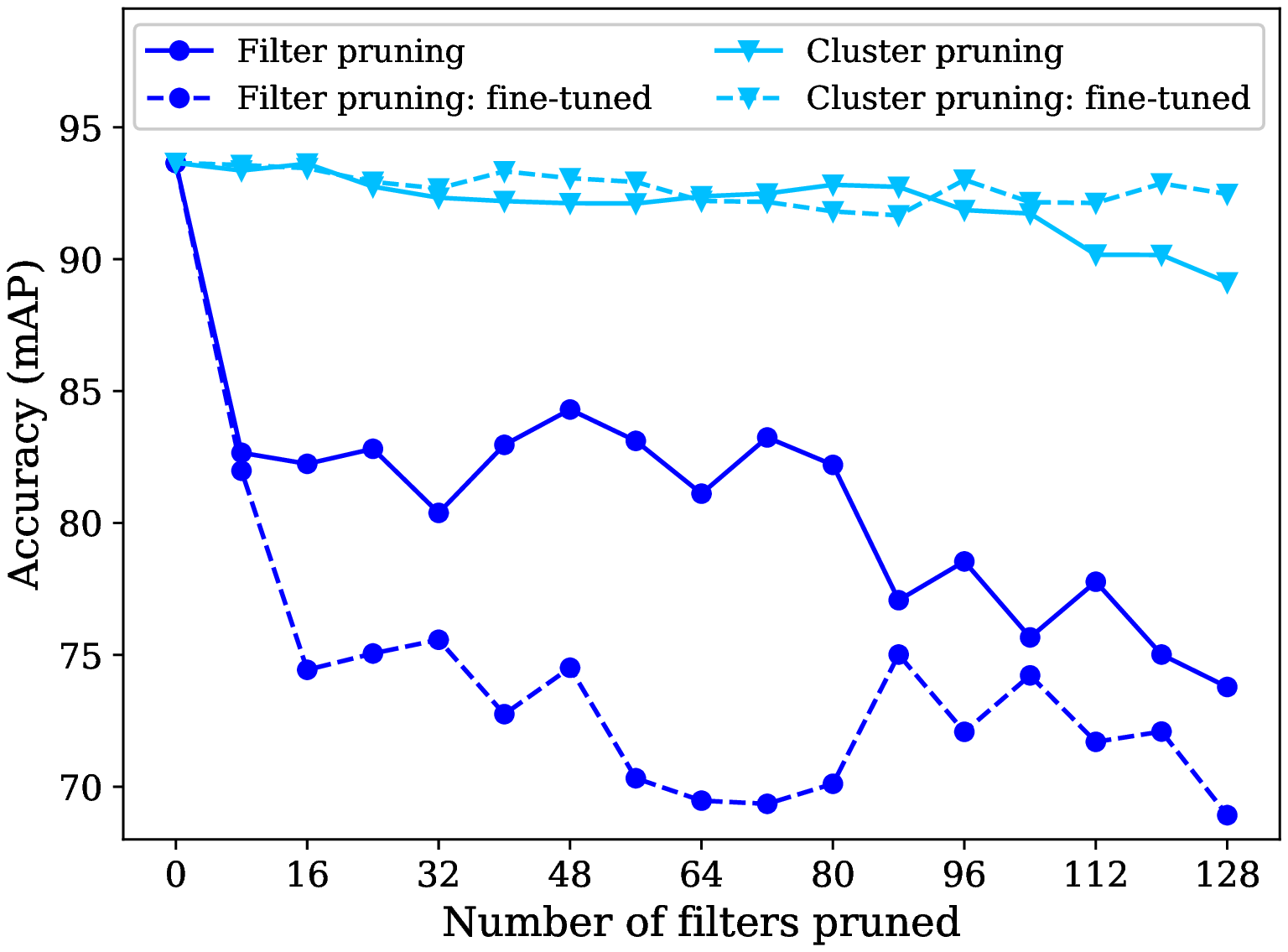}%
		\label{fig:NCS_accuracy_after_prunning_headcount}
	}%
	\subfloat[][CPU/GPU accuracy after pruning]{
		\includegraphics[width=0.40\textwidth]{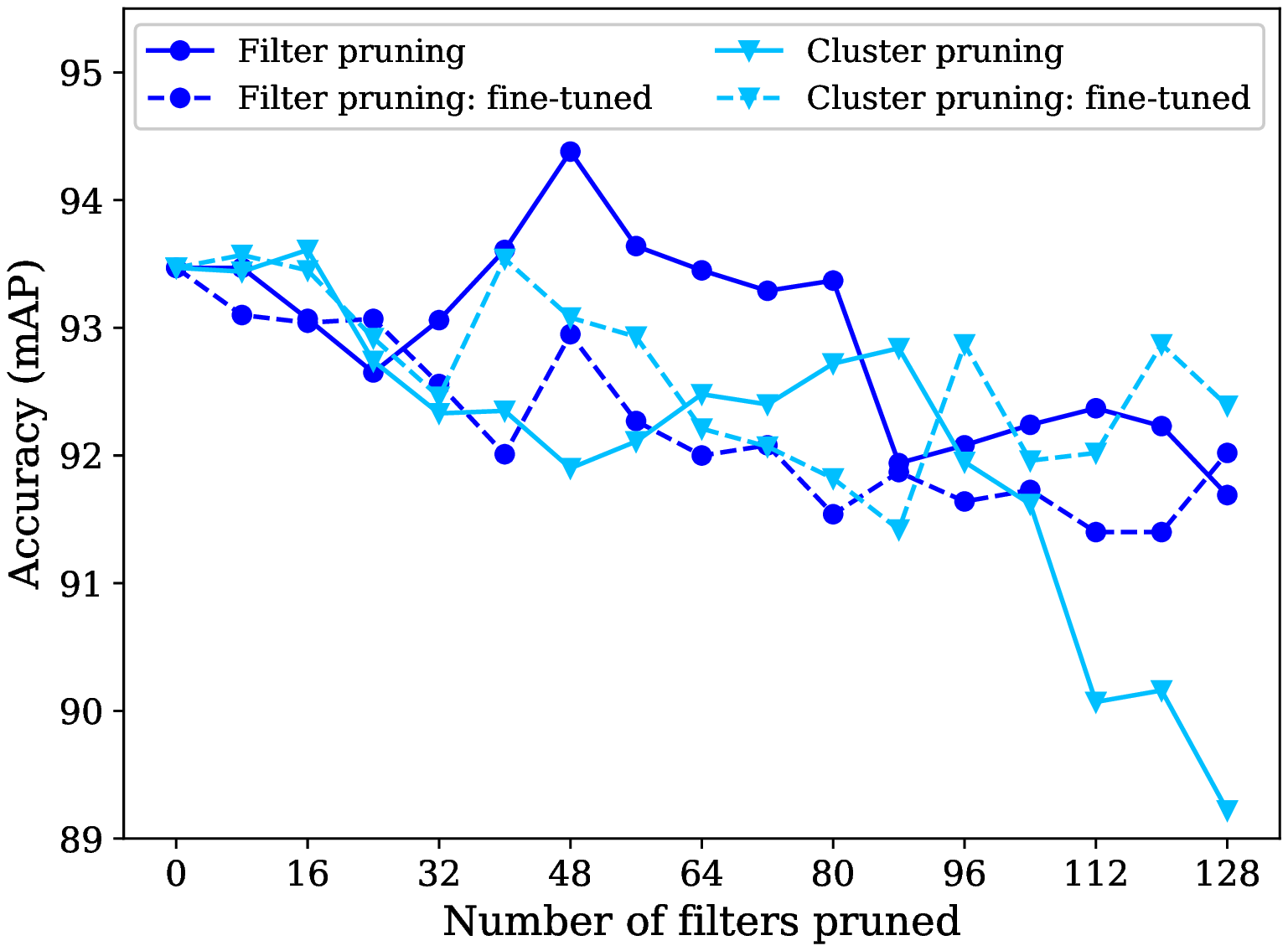}%
		\label{fig:Caffe_GPU_accuracy_after_prunning_headcount}\hfill
	}
	\caption{Accuracy for Head-Counting dataset : Filter pruning vs Cluster pruning (SSD-MobileNet)}
	\label{fig:Accuracy_HeadCount}%
\end{figure*}

\begin{figure*}
	\centering
	\subfloat[][NCS accuracy after pruning]{
		\includegraphics[width=0.40\textwidth]{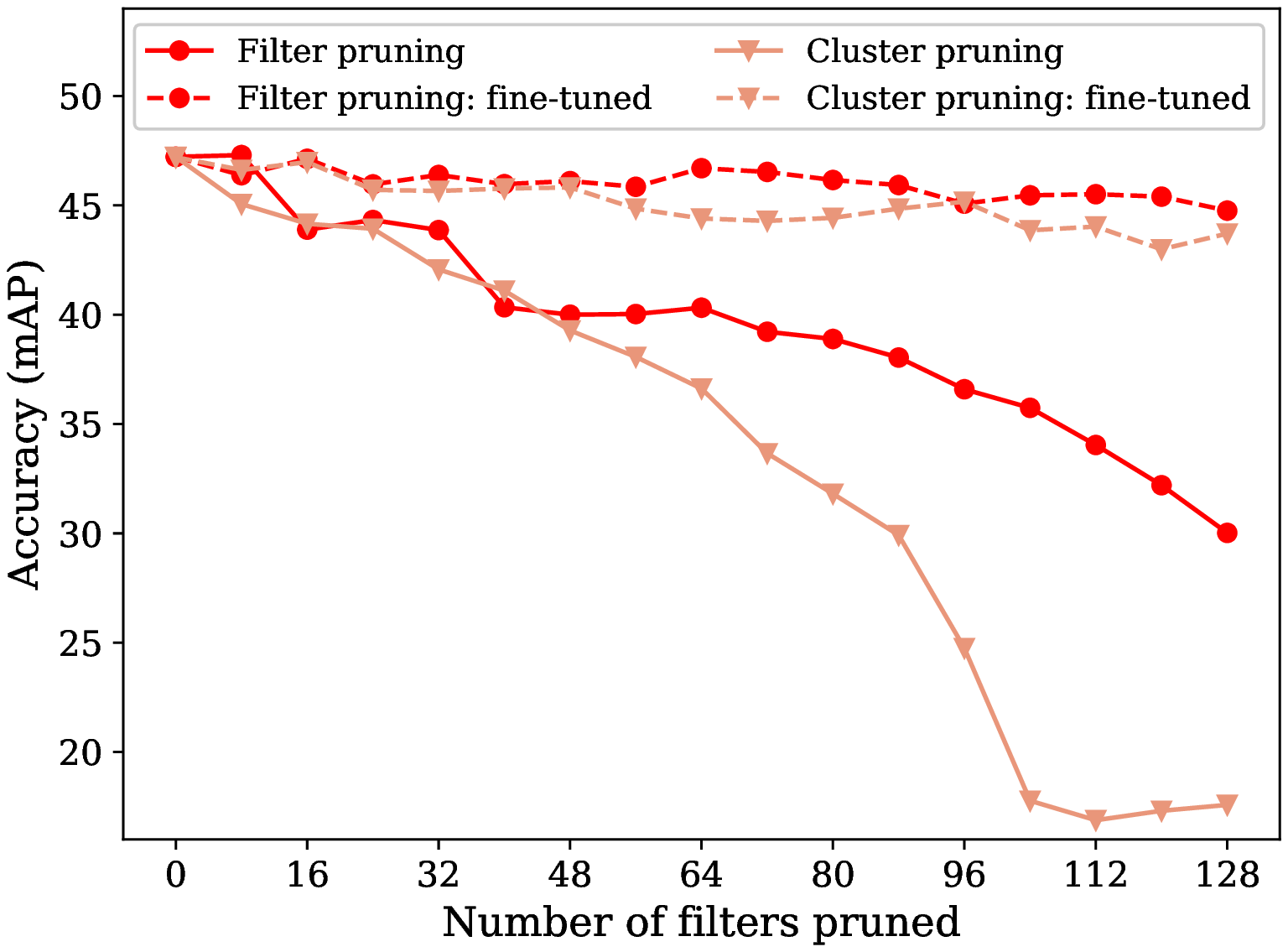}%
		\label{fig:NCS_accuracy_after_prunning_sq}
	}%
	\subfloat[][CPU/GPU accuracy after pruning]{
		\includegraphics[width=0.40\textwidth]{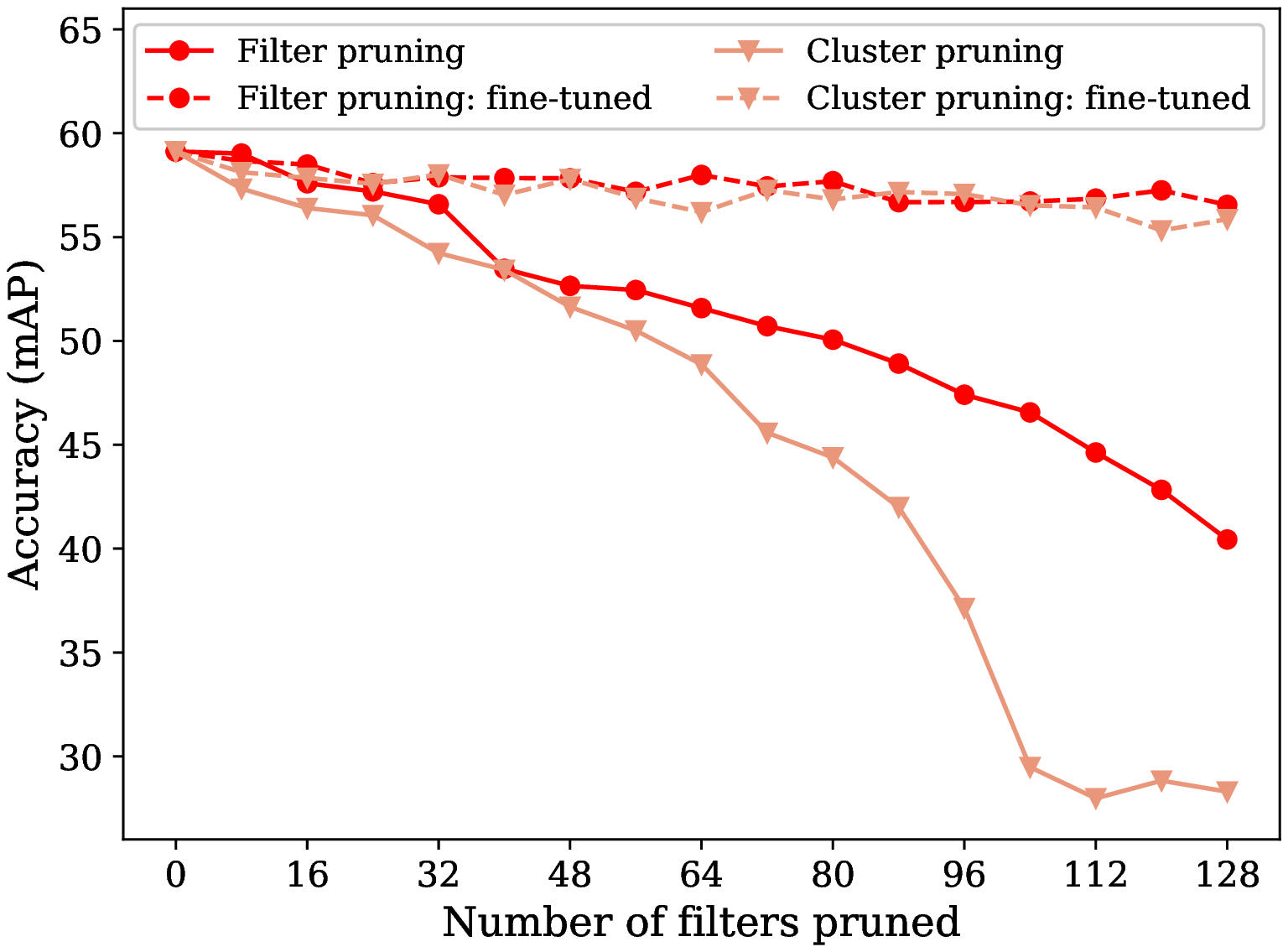}%
		\label{fig:Caffe_GPU_accuracy_after_prunning_sq}\hfill
	}
	\caption{Accuracy for Pascal-VOC dataset : Filter pruning vs Cluster pruning (SSD-SqueezeNet)}
	\label{fig:Accuracy_Pascal_VOC_sq}%
\end{figure*}

\begin{figure*}
	\centering
	\subfloat[][NCS accuracy after pruning]{
		\includegraphics[width=0.40\textwidth]{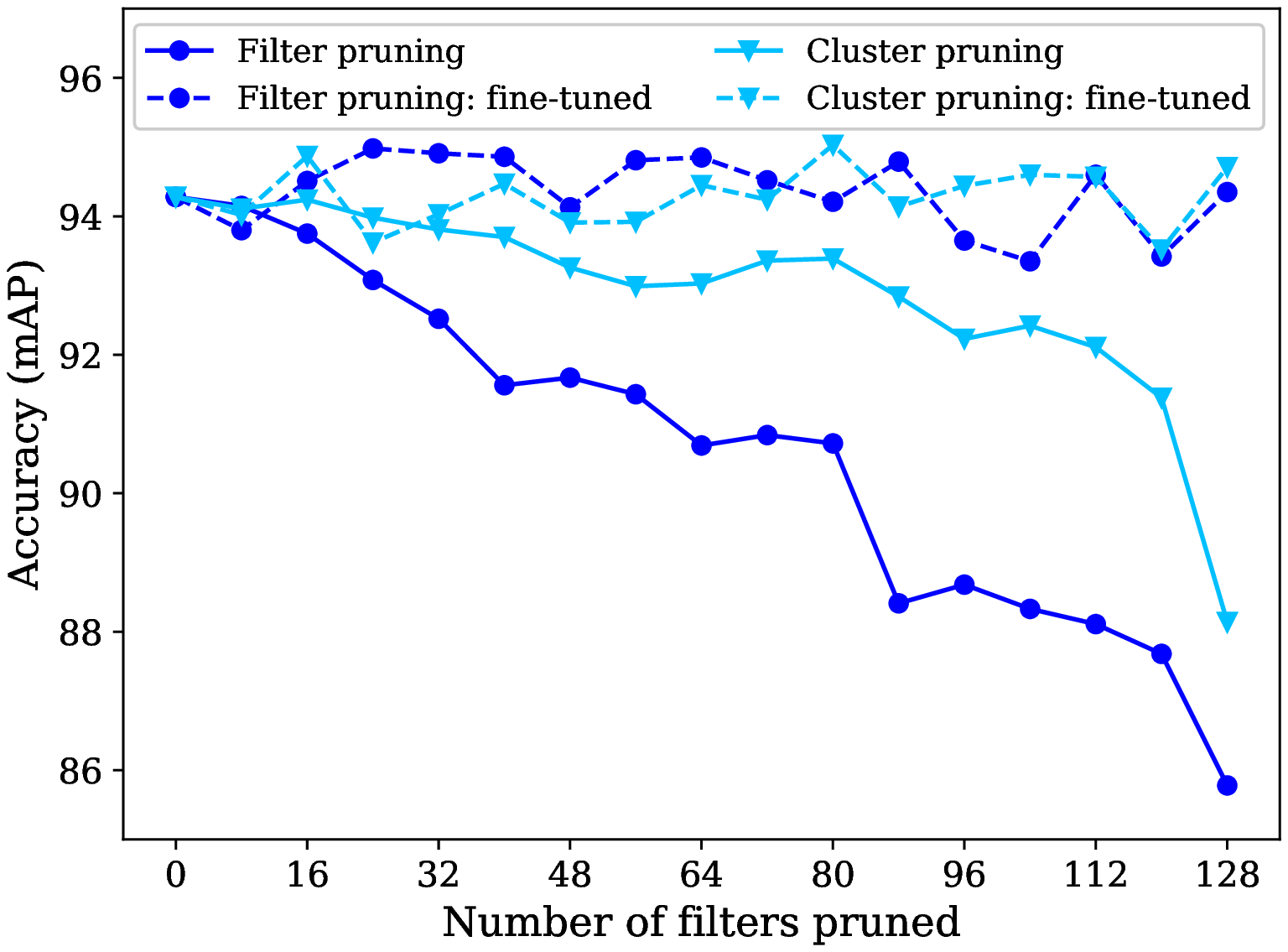}%
		\label{fig:NCS_accuracy_after_prunning_headcount_sq}
	}%
	\subfloat[][CPU/GPU accuracy after pruning]{
		\includegraphics[width=0.40\textwidth]{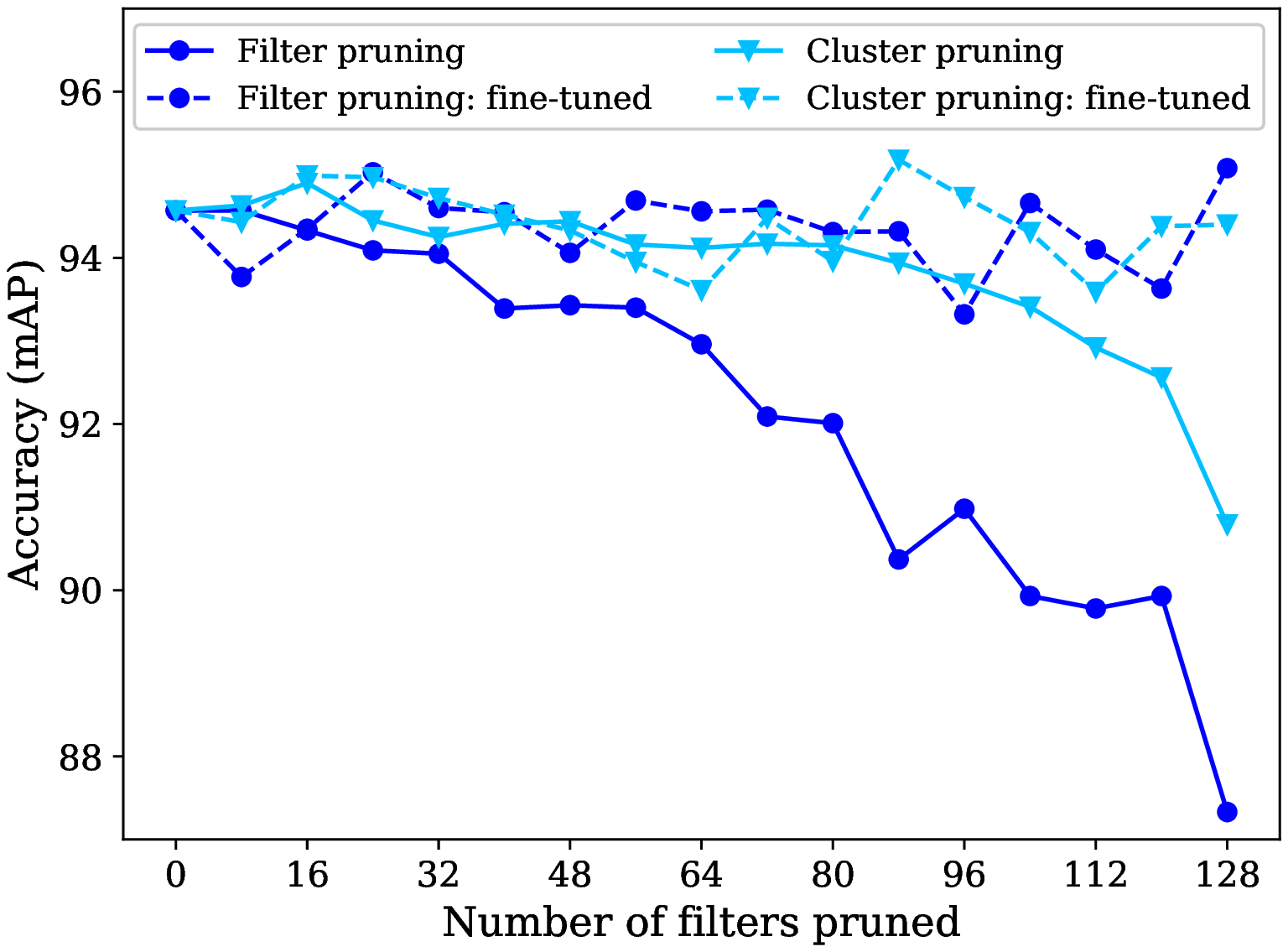}%
		\label{fig:Caffe_GPU_accuracy_after_prunning_headcount_sq}\hfill
	}
	\caption{Accuracy for Head-Counting dataset : Filter pruning vs Cluster pruning (SSD-SqueezeNet)}
	\label{fig:Accuracy_HeadCount_sq}%
\end{figure*}

As illustrated in the Fig. \ref{fig:Caffe_GPU_accuracy_after_prunning} and Fig. \ref{fig:Caffe_GPU_accuracy_after_prunning_headcount}, CPU/ GPU do not show a considerable difference of the drop of accuracy  in early stages when we compare the filter pruning and cluster pruning methodologies. But when the number of filters pruned are increasing, the accuracy drop becomes larger in the cluster pruning methodology. This is due to removal of the least significant filter with minimum weight, one by one considering the whole network in filter pruning method. But in cluster pruning method, filter group with the minimum average weigh is removed from a single layer. In this scenario, there can be filters with lesser weights in other layers than inside the filter group in the current layer. When we fine-tune the models after removing 8 filters in both methodologies, we can achieve almost the same accuracy for both methodologies with an accuracy loss not less than 2\% from the initial accuracy. We can clearly identify in Fig. \ref{fig:NCS_accuracy_after_prunning} and Fig. \ref{fig:NCS_accuracy_after_prunning_headcount}, the accuracy does not drop drastically in the proposed cluster pruning methodology and it outperforms the filter pruning method starting from the first pruning iteration. As we observed in single layer pruning experiment, the accuracy drop for the Movidius-NCS is very high if the network is not pruned in multiples of optimal cluster size. That is the reason behind the accuracy preservation in the cluster pruning method. Even though we fine-tune networks after pruning 8 filters, the filter pruning method can't achieve accuracy preserved by the cluster pruning methodology. Moreover, as shown in Fig. \ref{fig:Accuracy_HeadCount}, the filter pruning method shows an over-fitting scenario when we fine-tune the models pre-trained using the Head-Counting dataset. We also assume this might be due to some hyper parameter mis-specifications in the fine-tuning process.

For the SSD-SqueezeNet, there is no considerable difference between the accuracy results of Movidius-NCS and CPU/ GPU as shown in Fig. \ref{fig:Accuracy_Pascal_VOC_sq} and Fig. \ref{fig:Accuracy_HeadCount_sq}. This is due to no distinct accuracy changes we observed in single layer pruning experiment for both platforms while pruning the SSD-SqueezeNet. Cluster pruning method underperforms than the filter pruning methodology for Pascal-VOC dataset, while it outperforms in the Head-Counting dataset when the pruning is not followed by a fine-tuning step. For both dataset and for both network architectures, we can achieve almost the same accuracy using both methodologies with a fine-tuning step intermediately, where we lose not more than 1\% accuracy from the initial accuracy value. The Table \ref{network_shape} shows the dimensions of layers in SSD-MobileNet before pruning and after pruning, which was pre-trained on Head-Counting dataset. It shows how filter pruning methodology prunes filters unevenly, while cluster pruning method removes filters as clusters of 8 in a structured way.

According to the results of the whole model pruning, it has been proven that the inference latency of a detection network can be minimized using the proposed cluster pruning methodology, which outperforms the widely used filter pruning methodology. For some edge-AI devices, the accuracy drop we experience when the filters are pruned not considering the hardware response, can be mitigated using the proposed cluster pruning methodology. Furthermore, we can meet the same level of accuracy preservation of the filter pruning methodology by an intermediate fine-tuning step for the proposed cluster pruning methodology. Hence, our method can be applied to real-time vision applications to gain the performance requirement at our hand.

\begin{table*}
	\centering
	\caption{Dimensions of the layers after pruning filters in SSD-MobileNet pre-trained on Head-Counting dataset}
	\begin{tabular}{c|c|c|c|c|c}
		Convolution Layer & Original Dimention & Filter Pruning   & \# Filters Pruned & Cluster Pruning  & \# Filters Pruned \\ \hline
		conv1/dw   & (32, 1, 3, 3)      & (32, 1, 3, 3)    & 0              & (32, 1, 3, 3)    & 0              \\
		conv1      & (64, 32, 1, 1)     & (63, 32, 1, 1)   & 1              & (64, 32, 1, 1)   & 0              \\
		conv2/dw   & (64, 1, 3, 3)      & (63, 1, 3, 3)    & 1              & (64, 1, 3, 3)    & 0              \\
		conv2      & (128, 64, 1, 1)    & (124, 63, 1, 1)  & 4              & (128, 64, 1, 1)  & 0              \\
		conv3/dw   & (128, 1, 3, 3)     & (124, 1, 3, 3)   & 4              & (128, 1, 3, 3)   & 0              \\
		conv3      & (128, 128, 1, 1)   & (127, 124, 1, 1) & 1              & (128, 128, 1, 1) & 0              \\
		conv4/dw   & (128, 1, 3, 3)     & (127, 1, 3, 3)   & 1              & (128, 1, 3, 3)   & 0              \\
		conv4      & (256, 128, 1, 1)   & (256, 127, 1, 1) & 0              & (256, 128, 1, 1) & 0              \\
		conv5/dw   & (256, 1, 3, 3)     & (256, 1, 3, 3)   & 0              & (256, 1, 3, 3)   & 0              \\
		conv5      & (256, 256, 1, 1)   & (253, 256, 1, 1) & 3              & (256, 256, 1, 1) & 0              \\
		conv6/dw   & (256, 1, 3, 3)     & (253, 1, 3, 3)   & 3              & (256, 1, 3, 3)   & 0              \\
		conv6      & (512, 256, 1, 1)   & (510, 253, 1, 1) & 2              & (512, 256, 1, 1) & 0              \\
		conv7/dw   & (512, 1, 3, 3)     & (510, 1, 3, 3)   & 2              & (512, 1, 3, 3)   & 0              \\
		conv7      & (512, 512, 1, 1)   & (456, 510, 1, 1) & 56             & (416, 512, 1, 1) & 96             \\
		conv8/dw   & (512, 1, 3, 3)     & (456, 1, 3, 3)   & 56             & (416, 1, 3, 3)   & 96             \\
		conv8      & (512, 384, 1, 1)   & (491, 456, 1, 1) & 21             & (496, 416, 1, 1) & 16             \\
		conv9/dw   & (512, 1, 3, 3)     & (491, 1, 3, 3)   & 21             & (496, 1, 3, 3)   & 16             \\
		conv9      & (512, 512, 1, 1)   & (472, 491, 1, 1) & 40             & (496, 496, 1, 1) & 16             \\
		conv10/dw  & (512, 1, 3, 3)     & (472, 1, 3, 3)   & 40             & (496, 1, 3, 3)   & 16             \\
		conv10     & (512, 512, 1, 1)   & (512, 472, 1, 1) & 0              & (512, 496, 1, 1) & 0             
	\end{tabular}
	\label{network_shape}
\end{table*}

\subsection{\textbf{Edge-AI Application}}

A novel real-time people head counting system is presented in this section. Using a single overhead mounted camera, the system counts the number of people going in and out of an observed room. Counting is performed by analysing two consecutive frames of the video feed using object detection, tracking, and counting methodologies. Then, the number of people stay inside the room is used as a controlling parameter for air conditioner controllers in a Smart-Building system, which is beyond the scope of this paper. The proposed edge-AI hardware setup consists of a Raspberry Pi 3 development board, a camera module, and a Movidius-NCS. 

\begin{figure*}
	\centering
	\includegraphics[scale = 0.8]{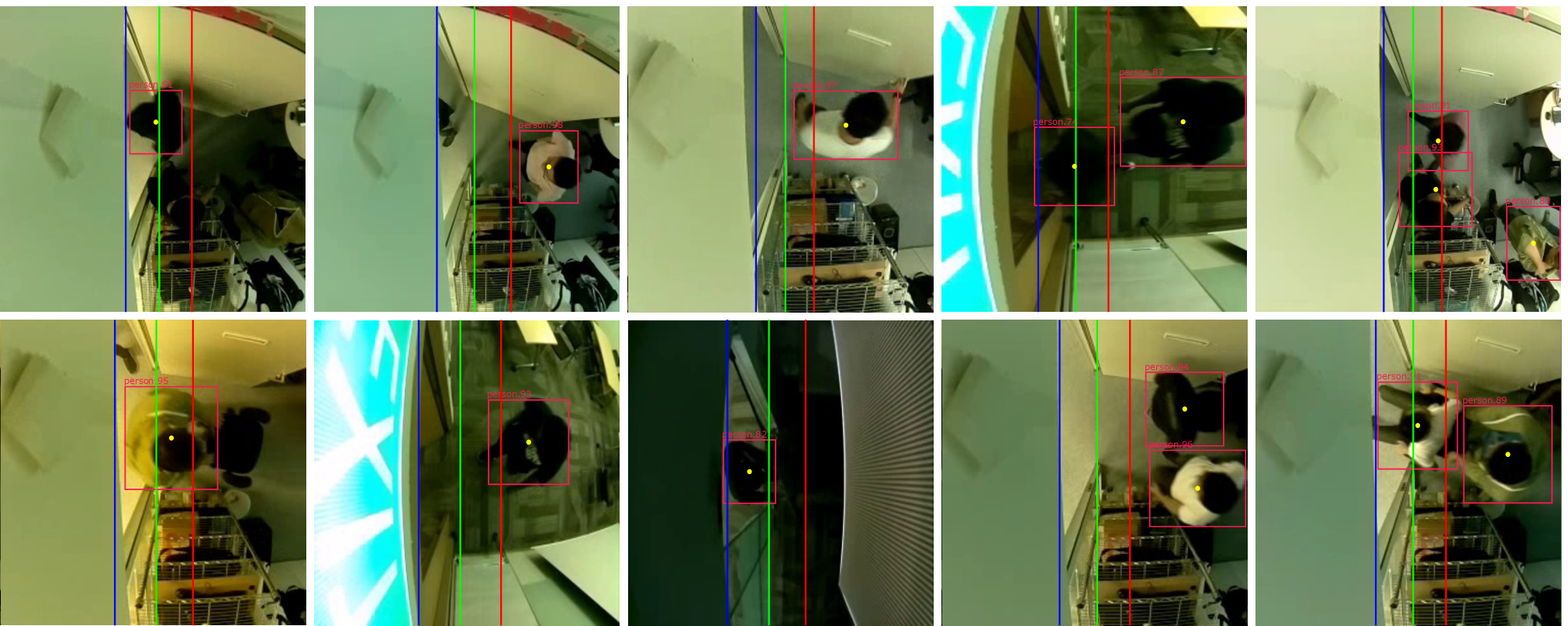}
	\caption{Detection of people entering rooms}
	\label{Detection_Image}
\end{figure*}

First, the SSD-MobileNet detection model, which is pre-trained on Pascal-VOC dataset is fine-tuned using the Head-Counting dataset and deployed in the edge-AI hardware setup mentioned. Real-time video frames from the camera are preprocessed and used as the input to the SSD-MobileNet. Movidius-NCS does the real-time inference by detecting the bounding boxes for the \textit{person} objects in the frame. The detected objects and their details in the current frame are saved in a data structure of the running program. Then, these saved object are compared with the object in the subsequent video frame using OpenCV \cite{bradski2000opencv} histogram comparison method to attain the object tracking capability in real-time. Once the objects are tracked, we determine the starting and end points of the object using centroids of the bounding boxes. As shown in each image of the Fig. \ref{Detection_Image}, we select two regions of interest using three virtual counting lines, where blue and green lines bounds the outside region, while green and red lines bound the inside region. If the starting and end points move from outside region to inside region, we count one person has entered the room and count one went outside vice versa. We have tested the performance of the system, achieving a correct people counting rate of 95\%. Additionally, to benchmark the throughput gain achievable by cluster pruning methodology, we use following hardware setups.

\begin{table}
	\centering
	\caption{Inference time for detection in milliseconds}
	\resizebox{0.45\textwidth}{!}{
		\begin{tabular}{c c c c c c }
			\toprule
			\multirow{2}{*}{Pruning Method} & \multicolumn{5}{c}{Computer Architecture} 	\\ \cline{2-6} 
			& Pi		& Pi+NCS	& CPU		& GPU		& TX2 		\\ \midrule
			Without Pruning		& 4787.18	& 84.92		& 215.22	& 22.33		& 104.23 	\\
			Filter Pruning		& 4756.14	& 86.63		& 215.61	& 21.89		& 97.54		\\
			Cluster Pruning		& 4461.64	& 82.37		& 195.85	& 21.74		& 94.03		\\ \midrule
			Gain from Filter Pruning	& 0.65\%	& -2.01\%	& -0.18\%	&-1.97\%	& 6.42\%	\\ 
			Gain from Cluster Pruning	& 6.80\%	& 3.00\%	& 9.00\%	& 2.64\%	& 9.78\%	\\ \bottomrule
	\end{tabular}}\\
	{\footnotesize$Performance \ Gain =\left( \frac{Without \ Pruning \ - \ After \ Pruning}{Without \ Pruning} \right)  \times 100\% $}
	\label{fw_time}
\end{table}

\begin{itemize}[noitemsep]
	\item Raspberry-Pi 3
	\item Raspberry-Pi 3 and Intel Movidius-NCS
	\item 2.10GHz Intel-Xeon CPU 
	\item 2.10GHz Intel-Xeon CPU and Nvidia GTX-1080Ti
	\item Nvidia Jetson-TX2
\end{itemize}

To test the performance of the application using above mentioned hardware setups, we used SSD-MobileNet model without pruning at first. Then, we pruned that model using filter pruning and cluster pruning methods separately. Approximately 1.28\% filters equal to 1.25\% parameters from the whole network has been pruned in both methodologies. The first approach of measuring the performance is identifying the total forward inference time through the neural network for each hardware setups in milliseconds. As illustrated in the Table \ref{fw_time}, it is clear that the performance gain from cluster pruning method has outperformed the filter pruning method. Furthermore, there is a performance degradation, which is represented as a minus value in the filter pruning method, using the proposed edge-AI hardware setup consisting of the Raspberry Pi and Movidius-NCS. It is overcome using the cluster pruning method as shown in positive percentage value in Table \ref{fw_time}. The next approach is to measure performance in frames per second (FPS) for the edge-AI application. We recorded a video of people entering and leaving a room using a overhead mounted camera. Then, this video is used instead of the real-time video feed and measured the FPS values using each hardware setups. The results shown in Table \ref{fps_tbl} indicate, performance gain in cluster pruning method outperform the filter pruning method in all hardware setups. From the results shown, it can be concluded that the performance of the edge-AI application is successfully uplifted using the proposed cluster pruning methodology.

\begin{table}
	\centering
	\caption{FPS values for the edge-AI application}
	\resizebox{0.45\textwidth}{!}{
		\begin{tabular}{c c c c c c}
			\toprule
			\multirow{2}{*}{Pruning Method}	& \multicolumn{5}{c}{Computer Architecture} 	\\ \cline{2-6} 
								& Pi    	& Pi+NCS 	& CPU		& GPU    	& TX2 		\\ \midrule
			Without Pruning     & 0.186 	& 6.346 	& 4.467 	& 42.389 	& 10.335	\\
			Filter Pruning      & 0.192 	& 6.331 	& 4.940 	& 48.450 	& 10.659	\\
			Cluster Pruning     & 0.204 	& 6.427 	& 5.031 	& 49.361 	& 11.004	\\ \midrule
			Gain from Filter Pruning	& 3.23\%	& -0.23\%	& 10.59\%	& 14.30\%	& 3.13\%	\\ 
			Gain from Cluster Pruning	& 9.68\%	& 1.28\%	& 12.63\%	& 16.45\%	& 6.47\%	\\ \bottomrule
	\end{tabular}}
	{\footnotesize$Performance \ Gain =\left( \frac{After \ Pruning \ - \ Without \ Pruning}{Without \ Pruning} \right)  \times 100\%  $}
	\label{fps_tbl}
\end{table}

\section{Conclusion and future works}

The solution proposed above clearly tackles the problem of steep increment of latency and sudden loss of accuracy when pruning filters in mobile neural networks deployed in edge-AI devices. The proposed cluster pruning methodology outperforms the conventional filter pruning methodology in both latency and accuracy perspectives and consistent across all the tested computing architectures. The proposed single layer pruning method can be used as a performance profiling methodology for neural networks using FPGA and ASIC AI computing architectures. Moreover, edge-AI applications can be optimized using the proposed cluster pruning methodology for resource efficient inference. 

We see a future direction of performing an ablation study to evaluate the best criteria for ranking filters according to their importance in the network. Therefore, we can extend our cluster pruning methodology with criteria such as Average Percentage of Zeros, Talor Criteria, and Thinet greedy algorithm etc. In addition, cluster pruning can be combined with novel training time pruning methods, such as Network Slimming, by introducing a group scaling factor for better hardware awareness. On the other hand, automatic pruning methods such as AMC and NetAdapt can be extended by pruning filters in clusters using the optimum cluster size mentioned in our work to reduce the exhaustive learning time and network searching time. Furthermore, this experiment can be extended to other popular neural network architectures such as AlexNet, VGG16, ResNet, ShuffleNet, TinyYolo and FastRCNN using other popular datasets, ImageNet, SVHN, CIFAR, etc.

\ifCLASSOPTIONcaptionsoff
  \newpage
\fi




\balance

\FloatBarrier
{\small
	\bibliographystyle{IEEEtran}
	\bibliography{paper_bib}

\begin{thebibliography}{10}
\providecommand{\url}[1]{#1}
\csname url@samestyle\endcsname
\providecommand{\newblock}{\relax}
\providecommand{\bibinfo}[2]{#2}
\providecommand{\BIBentrySTDinterwordspacing}{\spaceskip=0pt\relax}
\providecommand{\BIBentryALTinterwordstretchfactor}{4}
\providecommand{\BIBentryALTinterwordspacing}{\spaceskip=\fontdimen2\font plus
\BIBentryALTinterwordstretchfactor\fontdimen3\font minus
  \fontdimen4\font\relax}
\providecommand{\BIBforeignlanguage}[2]{{%
\expandafter\ifx\csname l@#1\endcsname\relax
\typeout{** WARNING: IEEEtran.bst: No hyphenation pattern has been}%
\typeout{** loaded for the language `#1'. Using the pattern for}%
\typeout{** the default language instead.}%
\else
\language=\csname l@#1\endcsname
\fi
#2}}
\providecommand{\BIBdecl}{\relax}
\BIBdecl

\bibitem{krizhevsky2012imagenet}
A.~Krizhevsky, I.~Sutskever, and G.~E. Hinton, ``Imagenet classification with
  deep convolutional neural networks,'' in \emph{Advances in neural information
  processing systems}, 2012, pp. 1097--1105.

\bibitem{szegedy2015going}
C.~Szegedy, W.~Liu, Y.~Jia, P.~Sermanet, S.~Reed, D.~Anguelov, D.~Erhan,
  V.~Vanhoucke, and A.~Rabinovich, ``Going deeper with convolutions,'' in
  \emph{Proceedings of the IEEE conference on computer vision and pattern
  recognition}, 2015, pp. 1--9.

\bibitem{simonyan2014very}
K.~Simonyan and A.~Zisserman, ``Very deep convolutional networks for
  large-scale image recognition,'' \emph{arXiv preprint arXiv:1409.1556}, 2014.

\bibitem{he2016deep}
K.~He, X.~Zhang, S.~Ren, and J.~Sun, ``Deep residual learning for image
  recognition,'' in \emph{Proceedings of the IEEE conference on computer vision
  and pattern recognition}, 2016, pp. 770--778.

\bibitem{russakovsky2015imagenet}
O.~Russakovsky, J.~Deng, H.~Su, J.~Krause, S.~Satheesh, S.~Ma, Z.~Huang,
  A.~Karpathy, A.~Khosla, M.~Bernstein \emph{et~al.}, ``Imagenet large scale
  visual recognition challenge,'' \emph{International journal of computer
  vision}, vol. 115, no.~3, pp. 211--252, 2015.

\bibitem{iandola2016squeezenet}
F.~N. Iandola, S.~Han, M.~W. Moskewicz, K.~Ashraf, W.~J. Dally, and K.~Keutzer,
  ``Squeezenet: Alexnet-level accuracy with 50x fewer parameters and< 0.5 mb
  model size,'' \emph{arXiv preprint arXiv:1602.07360}, 2016.

\bibitem{chollet2017xception}
F.~Chollet, ``Xception: Deep learning with depthwise separable convolutions,''
  in \emph{Proceedings of the IEEE conference on computer vision and pattern
  recognition}, 2017, pp. 1251--1258.

\bibitem{howard2017mobilenets}
A.~G. Howard, M.~Zhu, B.~Chen, D.~Kalenichenko, W.~Wang, T.~Weyand,
  M.~Andreetto, and H.~Adam, ``Mobilenets: Efficient convolutional neural
  networks for mobile vision applications,'' \emph{arXiv preprint
  arXiv:1704.04861}, 2017.

\bibitem{sandler2018mobilenetv2}
M.~Sandler, A.~Howard, M.~Zhu, A.~Zhmoginov, and L.-C. Chen, ``Mobilenetv2:
  Inverted residuals and linear bottlenecks,'' in \emph{Proceedings of the IEEE
  Conference on Computer Vision and Pattern Recognition}, 2018, pp. 4510--4520.

\bibitem{guo2018network}
J.~Guo, Y.~Li, W.~Lin, Y.~Chen, and J.~Li, ``Network decoupling: From regular
  to depthwise separable convolutions,'' \emph{arXiv preprint
  arXiv:1808.05517}, 2018.

\bibitem{zhang2018shufflenet}
X.~Zhang, X.~Zhou, M.~Lin, and J.~Sun, ``Shufflenet: An extremely efficient
  convolutional neural network for mobile devices,'' in \emph{Proceedings of
  the IEEE Conference on Computer Vision and Pattern Recognition}, 2018, pp.
  6848--6856.

\bibitem{ma2018shufflenet}
N.~Ma, X.~Zhang, H.-T. Zheng, and J.~Sun, ``Shufflenet v2: Practical guidelines
  for efficient cnn architecture design,'' in \emph{Proceedings of the European
  Conference on Computer Vision (ECCV)}, 2018, pp. 116--131.

\bibitem{huang2018condensenet}
G.~Huang, S.~Liu, L.~Van~der Maaten, and K.~Q. Weinberger, ``Condensenet: An
  efficient densenet using learned group convolutions,'' in \emph{Proceedings
  of the IEEE Conference on Computer Vision and Pattern Recognition}, 2018, pp.
  2752--2761.

\bibitem{redmon2016you}
J.~Redmon, S.~Divvala, R.~Girshick, and A.~Farhadi, ``You only look once:
  Unified, real-time object detection,'' in \emph{Proceedings of the IEEE
  conference on computer vision and pattern recognition}, 2016, pp. 779--788.

\bibitem{liu2016ssd}
W.~Liu, D.~Anguelov, D.~Erhan, C.~Szegedy, S.~Reed, C.-Y. Fu, and A.~C. Berg,
  ``Ssd: Single shot multibox detector,'' in \emph{European conference on
  computer vision}.\hskip 1em plus 0.5em minus 0.4em\relax Springer, 2016, pp.
  21--37.

\bibitem{wu2017squeezedet}
B.~Wu, F.~Iandola, P.~H. Jin, and K.~Keutzer, ``Squeezedet: Unified, small, low
  power fully convolutional neural networks for real-time object detection for
  autonomous driving,'' in \emph{Proceedings of the IEEE Conference on Computer
  Vision and Pattern Recognition Workshops}, 2017, pp. 129--137.

\bibitem{shen2017dsod}
Z.~Shen, Z.~Liu, J.~Li, Y.-G. Jiang, Y.~Chen, and X.~Xue, ``Dsod: Learning
  deeply supervised object detectors from scratch,'' in \emph{Proceedings of
  the IEEE International Conference on Computer Vision}, 2017, pp. 1919--1927.

\bibitem{li2018tiny}
Y.~Li, J.~Li, W.~Lin, and J.~Li, ``Tiny-dsod: Lightweight object detection for
  resource-restricted usages,'' \emph{arXiv preprint arXiv:1807.11013}, 2018.

\bibitem{lecun1990optimal}
Y.~LeCun, J.~S. Denker, and S.~A. Solla, ``Optimal brain damage,'' in
  \emph{Advances in neural information processing systems}, 1990, pp. 598--605.

\bibitem{hassibi1993second}
B.~Hassibi and D.~G. Stork, ``Second order derivatives for network pruning:
  Optimal brain surgeon,'' in \emph{Advances in neural information processing
  systems}, 1993, pp. 164--171.

\bibitem{yu2012exploiting}
D.~Yu, F.~Seide, G.~Li, and L.~Deng, ``Exploiting sparseness in deep neural
  networks for large vocabulary speech recognition,'' in \emph{2012 IEEE
  International conference on acoustics, speech and signal processing
  (ICASSP)}.\hskip 1em plus 0.5em minus 0.4em\relax IEEE, 2012, pp. 4409--4412.

\bibitem{han2015learning}
S.~Han, J.~Pool, J.~Tran, and W.~Dally, ``Learning both weights and connections
  for efficient neural network,'' in \emph{Advances in neural information
  processing systems}, 2015, pp. 1135--1143.

\bibitem{han2015deep}
S.~Han, H.~Mao, and W.~J. Dally, ``Deep compression: Compressing deep neural
  networks with pruning, trained quantization and huffman coding,'' \emph{arXiv
  preprint arXiv:1510.00149}, 2015.

\bibitem{hu2016network}
H.~Hu, R.~Peng, Y.-W. Tai, and C.-K. Tang, ``Network trimming: A data-driven
  neuron pruning approach towards efficient deep architectures,'' \emph{arXiv
  preprint arXiv:1607.03250}, 2016.

\bibitem{molchanov2016pruning}
P.~Molchanov, S.~Tyree, T.~Karras, T.~Aila, and J.~Kautz, ``Pruning
  convolutional neural networks for resource efficient inference,'' \emph{arXiv
  preprint arXiv:1611.06440}, 2016.

\bibitem{xu2018deep}
Y.~Xu, Y.~Wang, A.~Zhou, W.~Lin, and H.~Xiong, ``Deep neural network
  compression with single and multiple level quantization,'' in
  \emph{Thirty-Second AAAI Conference on Artificial Intelligence}, 2018.

\bibitem{li2016pruning}
H.~Li, A.~Kadav, I.~Durdanovic, H.~Samet, and H.~P. Graf, ``Pruning filters for
  efficient convnets,'' \emph{arXiv preprint arXiv:1608.08710}, 2016.

\bibitem{anwar2016compact}
S.~Anwar and W.~Sung, ``Compact deep convolutional neural networks with coarse
  pruning,'' \emph{arXiv preprint arXiv:1610.09639}, 2016.

\bibitem{he2017channel}
Y.~He, X.~Zhang, and J.~Sun, ``Channel pruning for accelerating very deep
  neural networks,'' in \emph{Proceedings of the IEEE International Conference
  on Computer Vision}, 2017, pp. 1389--1397.

\bibitem{luo2018thinet}
J.-H. Luo, H.~Zhang, H.-Y. Zhou, C.-W. Xie, J.~Wu, and W.~Lin, ``Thinet:
  pruning cnn filters for a thinner net,'' \emph{IEEE transactions on pattern
  analysis and machine intelligence}, 2018.

\bibitem{liu2017learning}
Z.~Liu, J.~Li, Z.~Shen, G.~Huang, S.~Yan, and C.~Zhang, ``Learning efficient
  convolutional networks through network slimming,'' in \emph{Proceedings of
  the IEEE International Conference on Computer Vision}, 2017, pp. 2736--2744.

\bibitem{corp}
\BIBentryALTinterwordspacing
N.~Corporation, ``Cuda c best practices guide.'' [Online]. Available:
  \url{https://docs.nvidia.com/cuda/cuda-c-best-practices-guide/index.html}
\BIBentrySTDinterwordspacing

\bibitem{han2016eie}
S.~Han, X.~Liu, H.~Mao, J.~Pu, A.~Pedram, M.~A. Horowitz, and W.~J. Dally,
  ``Eie: efficient inference engine on compressed deep neural network,'' in
  \emph{2016 ACM/IEEE 43rd Annual International Symposium on Computer
  Architecture (ISCA)}.\hskip 1em plus 0.5em minus 0.4em\relax IEEE, 2016, pp.
  243--254.

\bibitem{han2017ese}
S.~Han, J.~Kang, H.~Mao, Y.~Hu, X.~Li, Y.~Li, D.~Xie, H.~Luo, S.~Yao, Y.~Wang
  \emph{et~al.}, ``Ese: Efficient speech recognition engine with sparse lstm on
  fpga,'' in \emph{Proceedings of the 2017 ACM/SIGDA International Symposium on
  Field-Programmable Gate Arrays}.\hskip 1em plus 0.5em minus 0.4em\relax ACM,
  2017, pp. 75--84.

\bibitem{dai2018bigdl}
J.~Dai, Y.~Wang, X.~Qiu, D.~Ding, Y.~Zhang, Y.~Wang, X.~Jia, C.~Zhang, Y.~Wan,
  Z.~Li \emph{et~al.}, ``Bigdl: A distributed deep learning framework for big
  data,'' \emph{arXiv preprint arXiv:1804.05839}, 2018.

\bibitem{hadjis2015caffe}
S.~Hadjis, F.~Abuzaid, C.~Zhang, and C.~R{\'e}, ``Caffe con troll: Shallow
  ideas to speed up deep learning,'' in \emph{Proceedings of the Fourth
  Workshop on Data analytics in the Cloud}.\hskip 1em plus 0.5em minus
  0.4em\relax ACM, 2015, p.~2.

\bibitem{nvidia_cuda}
\BIBentryALTinterwordspacing
N.~Corporation, ``Cuda c programming guide.'' [Online]. Available:
  \url{https://docs.nvidia.com/cuda/cuda-c-programming-guide/index.html}
\BIBentrySTDinterwordspacing

\bibitem{lau2019survey}
B.~P.~L. Lau, S.~H. Marakkalage, Y.~Zhou, N.~U. Hassan, C.~Yuen, M.~Zhang, and
  U.-X. Tan, ``A survey of data fusion in smart city applications,''
  \emph{Information Fusion}, vol.~52, pp. 357--374, 2019.

\bibitem{marakkalage2018understanding}
S.~H. Marakkalage, S.~Sarica, B.~P.~L. Lau, S.~K. Viswanath,
  T.~Balasubramaniam, C.~Yuen, B.~Yuen, J.~Luo, and R.~Nayak, ``Understanding
  the lifestyle of older population: Mobile crowdsensing approach,'' \emph{IEEE
  Transactions on Computational Social Systems}, vol.~6, no.~1, pp. 82--95,
  2018.

\bibitem{liu2020cooperative}
R.~Liu, C.~Yuen, T.-N. Do, M.~Zhang, Y.~L. Guan, and U.-X. Tan, ``Cooperative
  positioning for emergency responders using self imu and peer-to-peer radios
  measurements,'' \emph{Information Fusion}, vol.~56, pp. 93--102, 2020.

\bibitem{parashar2017scnn}
A.~Parashar, M.~Rhu, A.~Mukkara, A.~Puglielli, R.~Venkatesan, B.~Khailany,
  J.~Emer, S.~W. Keckler, and W.~J. Dally, ``Scnn: An accelerator for
  compressed-sparse convolutional neural networks,'' in \emph{2017 ACM/IEEE
  44th Annual International Symposium on Computer Architecture (ISCA)}.\hskip
  1em plus 0.5em minus 0.4em\relax IEEE, 2017, pp. 27--40.

\bibitem{piyasena2019reducing}
D.~Piyasena, R.~Wickramasinghe, D.~Paul, S.-K. Lam, and M.~Wu, ``Reducing
  dynamic power in streaming cnn hardware accelerators by exploiting
  computational redundancies,'' in \emph{2019 29th International Conference on
  Field Programmable Logic and Applications (FPL)}.\hskip 1em plus 0.5em minus
  0.4em\relax IEEE, 2019, pp. 354--359.

\bibitem{piyasena2019lowering}
------, ``Lowering dynamic power of a stream-based cnn hardware accelerator,''
  in \emph{2019 IEEE 21st International Workshop on Multimedia Signal
  Processing (MMSP)}.\hskip 1em plus 0.5em minus 0.4em\relax IEEE, 2019, pp.
  1--6.

\bibitem{zoph2018learning}
B.~Zoph, V.~Vasudevan, J.~Shlens, and Q.~V. Le, ``Learning transferable
  architectures for scalable image recognition,'' in \emph{Proceedings of the
  IEEE conference on computer vision and pattern recognition}, 2018, pp.
  8697--8710.

\bibitem{cai2017reinforcement}
H.~Cai, T.~Chen, W.~Zhang, Y.~Yu, and J.~Wang, ``Reinforcement learning for
  architecture search by network transformation,'' \emph{arXiv preprint
  arXiv:1707.04873}, 2017.

\bibitem{ashok2017n2n}
A.~Ashok, N.~Rhinehart, F.~Beainy, and K.~M. Kitani, ``N2n learning: Network to
  network compression via policy gradient reinforcement learning,'' \emph{arXiv
  preprint arXiv:1709.06030}, 2017.

\bibitem{he2018amc}
Y.~He, J.~Lin, Z.~Liu, H.~Wang, L.-J. Li, and S.~Han, ``Amc: Automl for model
  compression and acceleration on mobile devices,'' in \emph{Proceedings of the
  European Conference on Computer Vision (ECCV)}, 2018, pp. 784--800.

\bibitem{yang2018netadapt}
T.-J. Yang, A.~Howard, B.~Chen, X.~Zhang, A.~Go, M.~Sandler, V.~Sze, and
  H.~Adam, ``Netadapt: Platform-aware neural network adaptation for mobile
  applications,'' in \emph{Proceedings of the European Conference on Computer
  Vision (ECCV)}, 2018, pp. 285--300.

\bibitem{sze2017efficient}
V.~Sze, Y.-H. Chen, T.-J. Yang, and J.~S. Emer, ``Efficient processing of deep
  neural networks: A tutorial and survey,'' \emph{Proceedings of the IEEE},
  vol. 105, no.~12, pp. 2295--2329, 2017.

\bibitem{movidius}
\BIBentryALTinterwordspacing
``Enabling machine intelligence at high performance and low power.'' [Online].
  Available: \url{https://www.movidius.com/technology}
\BIBentrySTDinterwordspacing

\bibitem{liu2018rethinking}
Z.~Liu, M.~Sun, T.~Zhou, G.~Huang, and T.~Darrell, ``Rethinking the value of
  network pruning,'' \emph{arXiv preprint arXiv:1810.05270}, 2018.

\bibitem{chuanqi305_2018}
\BIBentryALTinterwordspacing
``chuanqi305/mobilenet-ssd.'' [Online]. Available:
  \url{https://github.com/chuanqi305/MobileNet-SSD}
\BIBentrySTDinterwordspacing

\bibitem{chuanqi305}
\BIBentryALTinterwordspacing
``chuanqi305/squeezenet-ssd.'' [Online]. Available:
  \url{https://github.com/chuanqi305/SqueezeNet-SSD}
\BIBentrySTDinterwordspacing

\bibitem{everingham2010pascal}
M.~Everingham, L.~Van~Gool, C.~K. Williams, J.~Winn, and A.~Zisserman, ``The
  pascal visual object classes (voc) challenge,'' \emph{International journal
  of computer vision}, vol.~88, no.~2, pp. 303--338, 2010.

\bibitem{bradski2000opencv}
G.~Bradski, ``The opencv library,'' \emph{Dr. Dobb's Journal of Software
  Tools}, 2000.

\end{thebibliography}
}


\end{document}